\documentclass{article}

\usepackage{microtype}
\usepackage{graphicx}
\usepackage{subfigure}
\usepackage{booktabs} 

\usepackage{hyperref}

\usepackage[accepted]{icml2023}

\usepackage{amsmath}
\usepackage{amssymb}
\usepackage{mathtools}
\usepackage{amsthm}

\usepackage[capitalize,noabbrev]{cleveref}

\theoremstyle{plain}

\usepackage[textsize=tiny]{todonotes}

\usepackage{amsmath}
\usepackage{amssymb}
\usepackage{mathtools}
\usepackage{abbreviation} 
\usepackage{booktabs} 
\usepackage{algorithm} 
\usepackage{algorithmic} 
\usepackage{natbib} 
\usepackage{todonotes} 
\usepackage{optidef} 
\usepackage{subfigure} 

\usepackage[utf8]{inputenc} 
\usepackage{nomencl}

\usepackage{hologo} 
\usepackage{booktabs} 
\usepackage{tablefootnote} 

\usepackage[toc,page,header]{appendix}
\usepackage{minitoc}

\icmltitlerunning{Combinatorial Neural Bandits}

\doparttoc 
\faketableofcontents 

\begin{document}

\twocolumn[
\icmltitle{Combinatorial Neural Bandits}



\icmlsetsymbol{equal}{*}

\begin{icmlauthorlist}
\icmlauthor{Taehyun Hwang}{equal,snu}
\icmlauthor{Kyuwook Chai}{equal,snu}
\icmlauthor{Min-hwan Oh}{snu}

\end{icmlauthorlist}

\icmlaffiliation{snu}{Graduate School of Data Science, Seoul National University, Seoul, Republic of Korea}

\icmlcorrespondingauthor{Min-hwan Oh}{minoh@snu.ac.kr}

\icmlkeywords{Machine Learning, ICML}

\vskip 0.3in
]



\printAffiliationsAndNotice{\icmlEqualContribution} 

\begin{abstract}
We consider a contextual combinatorial bandit problem where in each round a learning agent selects a subset of arms and receives feedback on the selected arms according to their scores. The score of an arm is an unknown function of the arm's feature. Approximating this unknown score function with deep neural networks, we propose algorithms: Combinatorial Neural UCB ($\texttt{CN-UCB}$) and Combinatorial Neural Thompson Sampling ($\texttt{CN-TS}$). We prove that $\texttt{CN-UCB}$ achieves $\tilde{\mathcal{O}}(\tilde{d} \sqrt{T})$ or $\tilde{\mathcal{O}}(\sqrt{\tilde{d} T K})$ regret, where $\tilde{d}$ is the effective dimension of a neural tangent kernel matrix, $K$ is the size of a subset of arms, and $T$ is the time horizon. For $\texttt{CN-TS}$, we adapt an optimistic sampling technique to ensure the optimism of the sampled combinatorial action, achieving a worst-case (frequentist) regret of $\tilde{\mathcal{O}}(\tilde{d} \sqrt{TK})$.
To the best of our knowledge, these are the first combinatorial neural bandit algorithms with regret performance guarantees. In particular, $\texttt{CN-TS}$ is the first Thompson sampling algorithm with the worst-case regret guarantees for the general contextual combinatorial bandit problem. The numerical experiments demonstrate the superior performances of our proposed algorithms.
\end{abstract}

\section{Introduction}\label{sec:intro}

\begin{table*}[t!]
\begin{minipage}{\textwidth}
    \centering
    \caption{Comparison with the related work. For the neural bandit algorithms with single arm selection~\cite{zhou2020neural, zhang2021neural}, the reward function is not defined for a super arm (or the reward function can be viewed the same as the feedback for a single arm). All of the feedback models assume the boundedness of feedback.
    $\tilde{\mathcal{O}}$ is a big-$\mathcal{O}$ notation up to logarithmic factors.}
    \begin{tabular}{lccccc}
         \toprule
          & Combinatorial & Feedback & Reward  & Regret\\
         \midrule
         \texttt{C$^2$UCB}~\cite{qin2014contextual} & Yes & Linear & Lipschitz & $\tilde{\Ocal}(d \sqrt{T})$\\
         \texttt{CombLinUCB}~\cite{wen2015efficient} & Yes & Linear & Sum of feedback & $\tilde{\Ocal}(K\sqrt{d T \min \{ \log N, d\}})$ \\ 
         \texttt{CombLinTS}~\cite{wen2015efficient} & Yes & Linear & Sum of feedback & $\tilde{\Ocal}(d K\sqrt{T})$\textsuperscript{$\dag$} \\
         \texttt{CC-MAB}~\cite{chen2018contextual} & Yes & Lipschitz & Sub-modular & $\tilde{\Ocal}(2^d T^{\frac{d+4}{d+6}})$\\
         \texttt{ACC-UCB}~\cite{nika2020contextual} & Yes & Lipschitz & Sub-modular & $\tilde{\Ocal}(T^{\frac{\bar{d}+1}{\bar{d} + 2}})$\textsuperscript{$\ddag$} \\
         \texttt{Neural-UCB}~\cite{zhou2020neural} & No & General & - & $\tilde{\Ocal}(\tilde{d} \sqrt{T})$ \\
         \texttt{Neural-TS}~\cite{zhang2021neural} & No & General & - & $\tilde{\Ocal}(\tilde{d} \sqrt{T})$ \\
         \texttt{CN-UCB} (\textbf{this work}) & Yes & General & Lipschitz & $\tilde{\Ocal}(\tilde{d} \sqrt{T})$ or $\tilde{\Ocal}(\sqrt{\tilde{d} T K})$ \\
         \texttt{CN-TS} (\textbf{this work}) & Yes & General & Lipschitz & $\tilde{\Ocal}(\tilde{d} \sqrt{T K})$ \\
         \bottomrule
         \multicolumn{5}{l}{\textsuperscript{$\dag$}\footnotesize{Bayesian regret, which is a weaker notion of regret than the worst-case regret.}} \\
         \multicolumn{5}{l}{\textsuperscript{$\ddag$}\footnotesize{$\bar{d}$ represents the approximate optimality dimension related to context space.}}
    \end{tabular}
    \label{tab:comparison table}    
\end{minipage}
\end{table*}

We consider a general class of contextual semi-bandits with combinatorial actions, where in each round the learning agent is given a set of arms, chooses a subset of arms, and receives feedback on each of the chosen arms along with the reward based on the combinatorial actions. The goal of the agent is to maximize cumulative rewards through these repeated interactions.
The feedback is given as a function of the feature vectors (contexts) of the chosen arms. However, the functional form of the feedback model is unknown to the agent. 
Therefore, the agent needs to carefully balance exploration and exploitation in order to simultaneously learn the feedback model and optimize cumulative rewards. 

Many real-world applications are naturally combinatorial action selection problems. For example, in most online recommender systems, such as streaming services and online retail, recommended items are typically presented as a set or a list. 
Real-time vehicle routing can be formulated as the shortest-path problem under uncertainty which is a classic combinatorial problem. 
Network routing is also another example of a combinatorial optimization problem. 
Often, in these applications, the response model is not fully known a priori (e.g., user preferences in recommender systems, arrival time in vehicle routing) 
but can only be queried by sequential interactions. 
Therefore, these applications can be formulated as a combinatorial bandit problem.

Despite the generality and wide applicability of the combinatorial bandit problem in practice, the combinatorial action space
poses a greater challenge in balancing exploration and exploitation. 
To overcome such a challenge, parametric models such as the (generalized) linear model are often assumed for the feedback model \cite{qin2014contextual, wen2015efficient, kveton2015cascading, zong2016cascading, li2016contextual, li2019online, oh2019thompson}. These works typically extend the techniques in the (generalized) linear contextual bandits \cite{abe1999associative, auer2002using, filippi2010parametric, rusmevichientong2010linearly, abbasi2011improved,chu2011contextual, li2017provably} to utilize contextual information  
and the structure of the feedback/reward model to avoid the naive exploration in combinatorial action space. 
However, the representation power of the (generalized) linear model can be limited in many real-world applications. When the model assumptions are violated, often the performances of the algorithms that exploit the structure of a model can severely deteriorate.

Beyond the parametric assumption for the feedback model, discretization-based techniques~\cite{chen2018contextual, nika2020contextual} have been proposed to capture the non-linearity of the base arm under the Lipschitz condition on the feedback model. 
These techniques split the context space and compute an upper confidence bound of rewards for each context partition.
The performances of the algorithms strongly depend on the policy of how to partition the context space. 
However, splitting the context space is computationally expensive.
As the reward function becomes more complex, so does the splitting procedure.
Thus, it is challenging to apply these methods to high-dimensional contextual bandits.
In addition, the Lipschitz assumption on the feedback model (not on the reward function) does not hold when contexts close in the context space yield significantly different outcomes, i.e., when context space cannot be partitioned with respect to the outcome.

Deep neural networks have shown remarkable empirical performances in various learning tasks \cite{lecun2015deep, goodfellow2016deep, silver2016mastering}. 
Incorporating the superior representation power and recent advances in generalization theory of deep neural networks \cite{jacot2018ntk,cao2019generalization} into contextual bandits, 
an {\it upper confidence bound} (UCB) algorithm as an extension of the linear contextual bandit has been proposed~\cite{zhou2020neural}.
Extending the UCB approach, \citet{zhang2021neural} proposed a neural network-based  {\it Thompson Sampling} (TS) algorithm~\cite{thompson1933likelihood}.
However, these algorithms are proposed only for single-action selection.
How these algorithms generalize to the combinatorial action selection has remained open.

In this paper, we study provably efficient contextual combinatorial bandit algorithms without any modeling assumptions on the feedback model (with mild assumptions on the reward function which takes the feedback as an input). The extension to the combinatorial actions and providing provable performance guarantees requires more involved analysis and novel algorithmic modifications, particularly for the TS algorithm.
To briefly illustrate this challenge, even under the simple linear feedback model, a worst-case regret bound has not been known for a TS algorithm with various classes of combinatorial actions. This is due to the difficulty of ensuring the optimism of randomly sampled combinatorial actions (see Section~\ref{sec:ts_challenges}). Addressing such challenges, we adapt an optimistic sampling technique to our proposed TS algorithm, which allows us to achieve a sublinear regret.

Our main contributions are as follows:

\begin{itemize}
    \item We propose  algorithms for a general class of contextual combinatorial bandits: \textit{Combinatorial Neural UCB} ($\CNUCB$) and \textit{Combinatorial Neural Thompson Sampling} ($\CNTS$). To the best of our knowledge, these are the first neural-network based combinatorial bandit algorithms with regret guarantees.
    \item We establish that $\CNUCB$ is statistically efficient achieving $\OTilde(\tilde{d} \sqrt{T})$ or $\tilde{\Ocal}(\sqrt{\tilde{d} T K}) $ regret, where $\tilde{d}$ is the effective dimension of a neural tangent kernel matrix, $K$ is the size of a subset of arms, and $\TotalRound$ is the time horizon. This result matches the corresponding regret bounds of linear contextual bandits. 
    \item The highlight of our contributions is that $\CNTS$ is the first TS algorithm with the worst-case regret guarantees of $\OTilde( \tilde{d} \sqrt{T K})$ for a general class of contextual combinatorial bandits. 
    To our best knowledge, even under a simpler, linear feedback model, the existing TS algorithms with various combinatorial actions (including semi-bandit) do not have the worst-case regret guarantees. 
    This is due to the difficulty of ensuring the optimism of sampled combinatorial actions. We overcome this challenge by adapting optimistic sampling of the estimated reward while directly sampling in the reward space.
    \item The numerical evaluations demonstrate the superior performances of our proposed algorithms. We observe that the performances of the benchmark methods deteriorate significantly when the modeling assumptions are violated. In contrast, our proposed methods exhibit consistent competitive performances. 
\end{itemize}


\section{Problem setting}

\subsection{Notations}
 For a vector $\mathbf{x} \in \mathbb{R}^d$, we denote its $\ell_2$-norm by $\| \mathbf{x} \|_2 $ and its transpose by $\mathbf{x}^\top$. 
 The weighted $\ell_2$-norm associated with a positive definite matrix $\mathbf{A}$ is defined by $\| \mathbf{x} \|_\mathbf{A} := \sqrt{\mathbf{x}^\top \mathbf{A} \mathbf{x}}$. The trace of a matrix $\mathbf{A}$ is $tr(\mathbf{A})$. We define $[N]$ for a positive integer $N$ to be a set containing positive integers up to $N$, i.e., $\{1, 2, \ldots, N \}$. 

\subsection{Contextual Combinatorial Bandit}
In this work, we consider a contextual combinatorial bandit, where $\TotalRound$ is the total number of rounds, and $\TotalAction$ is the number of arms. 
At round $t \in [\TotalRound]$, 
a learning agent observes the set of context vectors for all arms $\{ \mathbf{x}_{t, i} \in \mathbb{R}^d \mid i \in [\TotalAction] \}$ 
and chooses a set of arms $S_t \subset [N]$ with size constraint $|S_t| = K$. 
$S_t$ is called a {\it super arm}. 
We introduce the notion of candidate super arm set $\mathcal{S} \subset 2^{[\TotalAction]}$ defined as the set of all possible subsets of arms with size $\SelectAction$, i.e., $\mathcal{S} := \{ S \subset [\TotalAction] \mid |S| = \SelectAction \}$.

\subsubsection{Score Function for Feedback} \label{subsec: subscore function}
Once a super arm $S_t \in \mathcal{S}$ is chosen, 
the agent then observes the scores of the chosen arms $\{ \Score_{t,i} \}_{i \in S_t}$ 
and receives a reward $R(S_t, \ScoreVector_t)$ as a function of the scores $\ScoreVector_t := [ \Score_{t,i} ]_{i=1}^N$ (which we discuss in the next section). 
This type of feedback is also known as {\it semi-bandit} feedback~\citep{audibert2014regret}. 
Note that in combinatorial bandits,
feedback and reward are not necessarily the same
as is the case in non-combinatorial bandits.
For each $t \in [T]$ and $i \in [\TotalAction]$, score $\Score_{t, i}$ is assumed to be generated as follows:
\begin{equation}\label{eq_true score}
    \Score_{t, i} = h(\mathbf{x}_{t,i}) + \Noise_{t,i}
\end{equation}
where $h$ is an {\it unknown} function satisfying $0 \le h(\mathbf{x}) \le 1$ for any  $\mathbf{x}$, and $\Noise_{t,i}$ is a $\SubGaussian$-sub-Gaussian noise satisfying $\mathbb{E}[\Noise_{t,i}| \mathcal{F}_t] = 0$ where $\mathcal{F}_t$ is the history up to round $t$. 

To learn the score function $h$ in Eq.\eqref{eq_true score}, we use a fully connected neural network~\cite{zhou2020neural, zhang2021neural}
with depth $\NetworkDepth \ge 2$, defined recursively:
\begin{equation}\label{DNN f}
\begin{split}
& f_1 = \mathbf{W}_1 \mathbf{x}\\
& f_\ell = \mathbf{W}_\ell \Activation(f_{\ell-1}), \quad 2 \le \ell \le \NetworkDepth, \\
& f(\mathbf{x} ; \NNParam) = \sqrt{\NetworkWidth} f_L
\end{split}
\end{equation}
where $\NNParam := [ \text{vec}(\mathbf{W}_1)^\top\!, ...,\text{vec}(\mathbf{W}_L)^\top ]^\top\!\!\in \mathbb{R}^p$ is the parameter of the neural network with $p=dm + m^2(\NetworkDepth-2) + m$, $\Activation(x):= \text{max}\{x, 0\}$ is the ReLU activation function, and $m$ is the width of each hidden layer. We denote the gradient of the neural network by $\mathbf{g}(\mathbf{x};\NNParam) := \nabla_{\NNParam} f(\mathbf{x};\NNParam) \in \mathbb{R}^p$.

\subsubsection{Reward Function \& Regret}
$R(S, \ScoreVector)$ is a deterministic reward function that measures the quality of the super arm $S$ based on the scores $\ScoreVector$. 
For example, the reward of a super arm $S_t$ can be the sum of the scores of arms in $S_t$, i.e., $R(S_t, \ScoreVector_t) = \sum_{i \in S_t} \Score_{t, i}$.
For our analysis, the reward function can be any function (linear or non-linear) which satisfies the following mild assumptions standard in the combinatorial bandit literature~\cite{qin2014contextual, li2016contextual}.

\begin{assumption}[Monotonicity]\label{assum:monotone}
$R(S, \ScoreVector)$ is monotone non-decreasing with respect to the score vector $\ScoreVector = [\Score_i]_{i=1}^N$, which means, for any $S$, if $\Score_i \le \Score'_i$ for all $i \in [\TotalAction]$, we have $R(S, \ScoreVector) \le R(S, \ScoreVector') \,$.
\end{assumption}

\begin{assumption}[Lipschitz continuity]\label{assum:lips}
$R(S, \ScoreVector)$ is Lipschitz continuous with respect to the score vector $\ScoreVector$ restricted on the arms in $S$, which means, there exists a constant $C_0 > 0 $ such that for any $\ScoreVector$ and $\ScoreVector'$, we have $ | R(S, \ScoreVector) - R(S, \ScoreVector') | \le C_0 \sqrt{\sum_{i \in S} (\Score_i - \Score'_i)^2 } \,$. 
\end{assumption}

\begin{remark} \label{remark_extension of reward}
Reward function satisfying Assumptions~\ref{assum:monotone} and ~\ref{assum:lips} encompasses a wide range of combinatorial feedback models including semi-bandit,
document-based or position based ranking models, and cascading models with little change to the learning algorithm. See Appendix~\ref{appendix_combinaotiral feedbacks} for more detailed discussions.
\end{remark}

Note that we do not require the agent to have direct knowledge on the explicit form of the reward function $R(S, \ScoreVector)$.
For the sake of clear exposition, we assume that the agent has access to an exact optimization oracle $\Oracle_{\mathcal{S}}(\ScoreVector)$ which takes a score vector $\ScoreVector$ as an input and returns the solution of the maximization problem $\argmax_{S \in \mathcal{S}} R(S, \ScoreVector)$. 

\begin{remark} \label{remark_exact oracle}
One can trivially extend the exact optimization oracle to an $\alpha$-approximation oracle without altering the learning algorithm or regret analysis. For problems such as semi-bandit algorithms choosing top-$\SelectAction$ arms, exact optimization can be done by simply sorting base scores. Even for more challenging assortment optimization, there are many polynomial-time (approximate) optimization methods available~\cite{rusmevichientong2010dynamic, james2014assortment}. 
For this reason, we present the regret analysis without $\alpha$-approximation assumption.
Extension of our regret analysis to an $\alpha$-approximation oracle is given in Appendix~\ref{appendix_Alpha Oracle}.
\end{remark}

The goal of the agent is to minimize the following (worst-case) cumulative expected regret:
\begin{equation}\label{eq_expected cumulative regret}
    \Regret(\TotalRound) = \sum_{t=1}^{\TotalRound} \mathbb{E} \left[ R(S_t^*, \ScoreVector_t^*) - R(S_t, \ScoreVector_t^*) \right]
\end{equation}
where $\ScoreVector^*_t := [ h(\mathbf{x}_{t, i}) ]_{i=1}^{\TotalAction}$ is the expected score which is unknown, and $S_t^* := \argmax_{S \in \mathcal{S}} R (S, \ScoreVector_t^*)$ is the offline {\it optimal} super arm at round $t$ under the expected score. 


\section{Combinatorial Neural UCB ($\CNUCB$)} \label{sec:CN-UCB}

\subsection{$\CNUCB$ Algorithm} \label{subsec:CN-UCB}

\begin{algorithm*}[t!]
   \caption{Combinatorial Neural UCB ($\CNUCB$)}
   \label{alg:CN-UCB}
\begin{algorithmic}
    \STATE {\bfseries Input:} Number of rounds $\TotalRound$, regularization parameter $\RegularizationParam$, norm parameter $\NormParam$, step size $\StepSize$, network width $\NetworkWidth$, number of gradient descent steps $\GDSteps$, network depth $\NetworkDepth$.
    \STATE {\bfseries Initialization: } Randomly initialize $\NNParam_0$ as described in Section~\ref{subsec:CN-UCB} and $\UCBGramMatrix_0 = \RegularizationParam \mathbf{I}$
    \FOR{$t = 1, ..., T$}
    \STATE Observe $\{ \mathbf{x}_{t,i} \}_{i \in [\TotalAction]}$
    \STATE Compute $\hat{\Score}_{t,i} = f(\mathbf{x}_{t,i}; \NNParam_{t-1})$ and 
    $u_{t,i} = \hat{\Score}_{t,i} + \gamma_{t-1} \left\| \Gradient(\mathbf{x}_{t,i}; \NNParam_{t-1}) / \sqrt{\NetworkWidth} \right\|_{\UCBGramMatrix_{t-1}^{-1}}$ for $i \in [\TotalAction]$
    \STATE Let $S_t = \Oracle_{\mathcal{S}} (\mathbf{u}_t + \mathbf{\UCBErrorTerm}_t)$ 
    \STATE Play super arm $S_t$ and observe $\{ \Score_{t, i} \}_{i \in S_t}$
    \STATE Update $\UCBGramMatrix_{t} = \UCBGramMatrix_{t-1} + \sum_{i \in S_t} \Gradient(\Context_{t,i}; \NNParam_{t-1})  \Gradient(\Context_{t,i}; \NNParam_{t-1})^\top / \NetworkWidth$
    \STATE Update $\NNParam_{t}$ to minimize the loss in Eq.\eqref{eq_L2-loss} using gradient descent with $\StepSize$ for $\GDSteps$ times
    \STATE Compute $\gamma_t$ and $e_{t+1}$ described in lemma~\ref{lemma_u'_modified UCB} 
  \ENDFOR
\end{algorithmic}
\end{algorithm*}

In this section, we present our first algorithm, Combinatorial Neural UCB ($\CNUCB$). $\CNUCB$ is a neural network-based UCB algorithm that operates using the {\it optimism in the face of uncertainty} (OFU) principle \cite{lai1985asymptotically} for combinatorial actions. 

In our proposed method, the neural network used for feedback model approximation is initialized by randomly generating each entry of $\NNParam_0 = [ \text{vec}(\mathbf{W}_1)^\top\!, ...,\text{vec}(\mathbf{W}_L)^\top ]^\top\!$, where for each $\ell \in [\NetworkDepth-1]$, $\mathbf{W}_\ell = (\mathbf{W}, \mathbf{0}; \mathbf{0}, \mathbf{W})$ with
each entry of $\mathbf{W}$ generated independently from $\mathcal{N}(0, 4/\NetworkWidth)$ 
and $\mathbf{W}_{\NetworkDepth} = (\mathbf{w}^\top, -\mathbf{w}^\top)$ with each entry of $\mathbf{w}$ generated independently from $\mathcal{N}(0, 2/\NetworkWidth)$. 
At each round $t \in [T]$, the algorithm observes the contexts for all arms, $\{ \mathbf{x}_{t, i} \}_{i \in [\TotalAction]}$ and computes an upper confidence bound $u_{t, i}$ of the expected score for each arm $i$, 
based on $\mathbf{x}_{t,i}, \NNParam_{t-1}$, and the exploration parameter $\gamma_{t-1}$.
Then, the sum of upper confidence bound score vector $\mathbf{u}_t := \left[ u_{t,i}\right]_{i=1}^{N}$ and the offset term vector $\mathbf{\UCBErrorTerm}_t := [\UCBErrorTerm_t, \cdots, \UCBErrorTerm_t]$,
(specified in Lemma~\ref{lemma_u'_modified UCB}),
is passed to the optimization oracle $\Oracle_{\mathcal{S}}$ as input. Then, the agent plays  $S_t = \Oracle_{\mathcal{S}}(\mathbf{u}_t + \mathbf{\UCBErrorTerm}_t)$ and receives the corresponding scores $\{ \Score_{t,i} \}_{i \in S_t}$ as feedback along with the reward associated with super arm $S_t$. 
Then the algorithm updates $\NNParam_{t}$ by minimizing the following loss function in Eq.\eqref{eq_L2-loss} using gradient descent with step size $\StepSize$ for $\GDSteps$ times.
\begin{equation}
\label{eq_L2-loss}
    \mathcal{L}(\NNParam) = \frac{1}{2}\sum_{k=1}^{n} \left( f(\mathbf{x}^k;\NNParam) - \Score^k \right)^2 + \frac{\NetworkWidth \RegularizationParam}{2} \|\NNParam - \NNParam_0 \|_2^2
\end{equation}
Here, the loss is minimized using $\ell_2$-regularization. Hyperparameter $\lambda$ controls the level of regularization, where the regularization centers at the randomly initialized neural network parameter~$\NNParam_0$.
The $\CNUCB$ algorithm is summarized in Algorithm~\ref{alg:CN-UCB}.

\subsection{Regret of $\CNUCB$}
\label{subsec:CN-UCB-Reg}
For brevity, we denote $\{ \mathbf{x}^k \}_{k=1}^{\TotalRound \TotalAction}$ be the collection of all contexts $\{ \mathbf{x}_{1,1}, \ldots, \mathbf{x}_{\TotalRound, \TotalAction}\}$.

\begin{definition} \cite{jacot2018ntk, cao2019generalization} \label{def_NTK matrix}
Define
    \begin{align*}
        & \tilde{\mathbf{H}}_{i,j}^{(1)} = \boldsymbol{\Sigma}_{i,j}^{(1)} = \langle \mathbf{x}^i, \mathbf{x}^j \rangle, \mathbf{A}_{i,j}^{(\ell)}= \begin{pmatrix} \boldsymbol{\Sigma}_{i,i}^{(\ell)} & \boldsymbol{\Sigma}_{i,j}^{(\ell)} \\ \boldsymbol{\Sigma}_{j,i}^{(\ell)} & \boldsymbol{\Sigma}_{j,j}^{(\ell)} \end{pmatrix},
        \\
        & \boldsymbol{\Sigma}_{i, j}^{(\ell +1)} = 2 \mathbb{E}_{(y,z) \sim \mathcal{N}(\mathbf{0}, \mathbf{A}_{i,j}^{(\ell)})} \left[ \Activation(y) \Activation(z) \right], 
        \\        
        & \tilde{\mathbf{H}}_{i,j}^{(\ell +1)} = 2 \tilde{\mathbf{H}}_{i,j}^{(\ell)} \mathbb{E}_{(y,z) \sim \mathcal{N}(\mathbf{0}, \mathbf{A}_{i,j}^{(\ell)})} \left[ \Activation'(y) \Activation'(z)\right]
         + \boldsymbol{\Sigma}_{i, j}^{(\ell +1)}.
    \end{align*}
Then, $\mathbf{H} = (\tilde{\mathbf{H}}^{(\NetworkDepth)} + \boldsymbol{\Sigma}^{(\NetworkDepth)})/2$ is called the neural tangent kernel (NTK) matrix on the context set $\{ \Context^k\}_{k=1}^{\TotalRound \TotalAction}$.
\end{definition}
The NTK matrix $\mathbf{H}$ on the contexts $\{ \mathbf{x}^k \}_{k=1}^{\TotalRound \TotalAction}$ is defined recursively from the input layer to the output layer of the network \cite{zhou2020neural,  zhang2021neural}. 
Then,
we define the effective dimension of the NTK matrix $\mathbf{H}$.
\begin{definition} \label{def_effective dimension}
The effective dimension $\EffectiveDim$ of the NTK matrix $\mathbf{H}$ with regularization parameter $\RegularizationParam$ is defined as
\begin{equation} \label{eq_effective dimenstion}
    \EffectiveDim = \frac{ \log \det ( \mathbf{I} + \mathbf{H}/\RegularizationParam )}{\log (1+\TotalRound \TotalAction / \RegularizationParam )}.    
\end{equation}
\end{definition}
The effective dimension can be thought of as the actual dimension of contexts in the Reproducing Kernel Hilbert Space spanned by the NTK. 
For further detailed information, we refer the reader to \citet{jacot2018ntk}. We proceed under the following assumption regarding  contexts:%
\begin{assumption}\label{assum:context}
 For any $k \in [\TotalRound \TotalAction]$, $\| \mathbf{x}^k \|_2 = 1$ and 
$[\mathbf{x}^k]_j = [\mathbf{x}^k]_{j+\frac{d}{2}} $ for $1 \le j \le \frac{d}{2}$.
Furthermore, for some $\lambda_0 > 0 $, $\mathbf{H} \succeq \RegularizationParamNTK \mathbf{I}$ .
\end{assumption}
This is a mild assumption commonly used in the neural contextual bandits \cite{zhou2020neural, zhang2021neural}. $\|\mathbf{x}\|_2 = 1$ is only imposed for simplicity of exposition. For the condition on the entries of $\mathbf{x}$, we can always re-construct a new context $\mathbf{x}' = [\mathbf{x}^\top, \mathbf{x}^\top]^\top/\sqrt{2}$. A positive definite NTK matrix is a standard assumption in the NTK literature \cite{du2018gradient, arora2019fine}, also used in the aforementioned neural contextual bandit literature. The following theorem provides the regret bound of Algorithm~\ref{alg:CN-UCB}.
\begin{theorem} \label{thm_CN-UCB}
Suppose Assumptions~\ref{assum:monotone}-\ref{assum:context} hold.
Let $\mathbf{h} = \left[ h(\mathbf{x}^{k}) \right]_{k=1}^{\TotalRound \TotalAction} \in \mathbb{R}^{\TotalRound \TotalAction}$. If we run $\CNUCB$ with
    \begin{align*}
        &\NetworkWidth \ge \text{poly}(\TotalRound, \NetworkDepth, \TotalAction, \RegularizationParam^{-1}, \RegularizationParamNTK^{-1}, \log\TotalRound) \,,
        \\
        & \StepSize = \bar{C}_1 (\TotalRound \SelectAction \NetworkWidth \NetworkDepth + \NetworkWidth \RegularizationParam)^{-1}, \, \RegularizationParam \ge \bar{C}_2 \NetworkDepth \SelectAction,
        \\
        & \GDSteps = 2 \log \left( \sqrt{\RegularizationParam/\TotalRound \SelectAction}/ (\RegularizationParam + \bar{C}_3 \TotalRound \SelectAction \NetworkDepth ) \right) \TotalRound \SelectAction \NetworkDepth/(\bar{C}_1 \RegularizationParam)
    \end{align*}
for some positive constants $\bar{C}_1, \bar{C}_2, \bar{C}_3$ with $\bar{C}_2 \ge \sqrt{ \max_{t, i} \| \Gradient (\Context_{t,i}; \NNParam_{t-1}) / \sqrt{\NetworkWidth} \|_2^2/\NetworkDepth}$ and $\NormParam \ge \sqrt{ 2 \mathbf{h}^\top \NTKMatrix^{-1} \mathbf{h}}$,
then the cumulative expected regret of $\CNUCB$\, over horizon~$T$ is upper-bounded by
    \begin{equation*}
        \Rcal(T) = \tilde{\Ocal}\left( \sqrt{ \tilde{d} T \max\{ \tilde{d}, K\} }\right) \, .
    \end{equation*}
\end{theorem}
\noindent {\bf Discussion of Theorem~\ref{thm_CN-UCB}.} 
Theorem~\ref{thm_CN-UCB} establishes that the cumulative regret of \texttt{CN-UCB} is $\OTilde(\tilde{d}\sqrt{T})$ or $\OTilde(\sqrt{ \EffectiveDim \TotalRound \SelectAction})$, whichever is higher. 
This result matches the state-of-the-art regret bounds for the contextual combinatorial bandits with the linear feedback model \cite{li2016contextual, zong2016cascading, li2018online}.
Note that the existence of $\bar{C}_2$ in Theorem~\ref{thm_CN-UCB} follows from Lemma~B.6 in~\citet{zhou2020neural} and Lemma~B.3 in \citet{cao2019generalization}. While the regret analysis for Theorem~\ref{thm_CN-UCB} has its own merit, the technical lemmas for Theorem~\ref{thm_CN-UCB} also provide the building block for the more challenging analysis of the TS algorithm which is presented in Section~\ref{sec:CN-TS}.

\subsection{Proof Sketch of Theorem~\ref{thm_CN-UCB}}
In this section, we provide a proof sketch of the regret upper bound in Theorem~\ref{thm_CN-UCB} and the key lemmas whose proofs are deferred to Appendix~\ref{appx: regret bound for CN-UCB}. 

Recall that we do not make any parametric assumption on the score function, but a neural network is used to approximate the unknown score function. Hence, we need to carefully control the approximation error. To achieve this, we use an over-parametrized neural network, for which the following condition on the neural network width is required.
\begin{condition} \label{cond_network width}
The network width $\NetworkWidth$ satisfies
    \[\begin{aligned}
        &\begin{aligned}
            & \NetworkWidth \ge C \max \{
            \NetworkDepth^{-\frac{3}{2}} \SelectAction^{-\frac{1}{2}} \RegularizationParam^{\frac{1}{2}} \left( \log(\TotalRound \TotalAction \NetworkDepth^2 / \ConfidenceParam) \right)^{\frac{3}{2}},
            \\
            & \phantom{{}={}}\TotalRound^6 \TotalAction^6 \NetworkDepth^6 \log(\TotalRound^2 \TotalAction^2 \NetworkDepth / \ConfidenceParam) \max \{ \RegularizationParam_0^{-4}, 1 \} \} \, ,
        \end{aligned}
        \\
        &\begin{aligned}
            & \NetworkWidth \left( \log \NetworkWidth \right)^{-3}
            \ge C \TotalRound^4 \SelectAction^4 \NetworkDepth^{21} \RegularizationParam^{-4} (1+\sqrt{\TotalRound/\RegularizationParam})^6
            \\
            & \phantom{{}={}} + C \TotalRound \SelectAction \NetworkDepth^{12} \RegularizationParam^{-1}
            + C \TotalRound^4 \SelectAction^4 \NetworkDepth^{18} \RegularizationParam^{-10} (\RegularizationParam + \TotalRound \NetworkDepth)^6 \, ,
        \end{aligned}
    \end{aligned}
    \]
where $C$ is a positive absolute constant.
\end{condition}
Unlike the analysis of the (generalized) linear UCB algorithms \cite{abbasi2011improved, li2017provably}, we do not have guarantees on 
the upper confidence bound $u_{t,i}$ being higher than the expected score $v_{t,i}^* =h(\Context_{t,i})$ due to the approximation error.
Therefore, we consider adding the offset term to the the upper confidence bound to ensure optimism.
The following lemma shows that the upper confidence bounds $u_{t,i}$ do not deviate far from the expected score $h(\Context_{t,i})$ and 
specifies the value of the offset term.
\begin{lemma} \label{lemma_u'_modified UCB}
For any $\delta \in (0,1)$, suppose the width of the neural network $m$ satisfies Condition~\ref{cond_network width}.
Let $\gamma_t$ be a positive scaling factor defined as
    \begin{equation*} \label{eq:gamma t}
        \begin{split}
            \gamma_{t} &=
             \Gamma_{1,t} \bigg( \SubGaussian \sqrt{ \log \frac{ \det \UCBGramMatrix_t}{\det \RegularizationParam \mathbf{I}} + \Gamma_{2,t} -2 \log \ConfidenceParam} + \sqrt{\RegularizationParam} \NormParam \bigg)
            \\
            &\quad + (\RegularizationParam + C_1 t \SelectAction \NetworkDepth) \left( (1-\StepSize \NetworkWidth \RegularizationParam)^{\frac{\GDSteps}{2}} \sqrt{t \SelectAction/\RegularizationParam} + \Gamma_{3,t} \right) \, , 
        \end{split}
    \end{equation*}
where
    \begin{align*}
        & \Gamma_{1,t} = \sqrt{1+C_{\Gamma,1} t^{\frac{7}{6}} K^{\frac{7}{6}} \NetworkDepth^4 \RegularizationParam^{-\frac{7}{6}} \NetworkWidth^{-\frac{1}{6}} \sqrt{\log \NetworkWidth}} \, ,
        \\
        & \Gamma_{2,t} = C_{\Gamma, 2} t^{\frac{5}{3}} K^{\frac{5}{3}} \NetworkDepth^4 \RegularizationParam^{-\frac{1}{6}} \NetworkWidth^{-\frac{1}{6}} \sqrt{\log \NetworkWidth} \, ,
        \\
        & \Gamma_{3,t} = C_{\Gamma, 3} t^{\frac{7}{6}} \SelectAction^{\frac{7}{6}} \NetworkDepth^{\frac{7}{2}} \RegularizationParam^{-\frac{7}{6}} \NetworkWidth^{-\frac{1}{6}} \sqrt{\log \NetworkWidth} (1+\sqrt{t \SelectAction / \RegularizationParam}) \, ,
    \end{align*}
for some constants $C_1, C_{\Gamma, 1}, C_{\Gamma,2}, C_{\Gamma,3} > 0$.
If $\StepSize \le C_2( \TotalRound \SelectAction \NetworkWidth \NetworkDepth + \NetworkWidth \RegularizationParam)^{-1}$ for some $C_2 > 0$,
then for any $t \in [\TotalRound]$ and $i \in [\TotalAction]$, with probability at least $1-\ConfidenceParam$ we have
    \begin{equation*}
         |u_{t, i} - h(\Context_{t,i})|  \le 2 \gamma_{t-1} \left\| \Gradient(\Context_{t,i}; \NNParam_{t-1}) /  \sqrt{\NetworkWidth} \right\|_{\UCBGramMatrix_{t-1}^{-1}} + \UCBErrorTerm_t \, ,
    \end{equation*}
where $\UCBErrorTerm_t $ is defined for some absolute constants $C_3, C_4  > 0 $ as follows.
    \begin{align*}
        \UCBErrorTerm_t 
        &:= C_3 \gamma_{t-1}  t^{\frac{1}{6}} \SelectAction^{\frac{1}{6}} \NetworkDepth^{\frac{7}{2}} \RegularizationParam^{-\frac{2}{3}}   \NetworkWidth^{-\frac{1}{6}}\! \sqrt{ \log \NetworkWidth } 
        \\
        & \phantom{{}={}} + C_4 t^{\frac{2}{3}} \SelectAction^{\frac{2}{3}} \RegularizationParam^{-\frac{2}{3}} \NetworkWidth^{-\frac{1}{6}}\! \sqrt{ \log \NetworkWidth } \,.
    \end{align*}
\end{lemma}
The next corollary shows that the surrogate upper confidence bound $u_{t,i} + \UCBErrorTerm_t$ is higher than true mean score $h(\Context_{t,i})$ with high probability.
\begin{corollary}\label{cor:ucb-optimism}
With probability at least $1-\ConfidenceParam$ 
    \begin{equation*}
        u_{t,i} + \UCBErrorTerm_{t}  \ge h(\Context_{t,i}) \,.
    \end{equation*}
\end{corollary}
The point of Corollary~\ref{cor:ucb-optimism} is that in~\citet{zhou2020neural}, to bound the instantaneous regret, it is enough for the agent to choose only one optimistic action (see Lemma 5.3 in~\citet{zhou2020neural}), while in our case, the agent has to choose the optimistic super arm in order to bound the instantaneous regret (See Eq.~\eqref{eq:instantaneous regret} in Proof of Theorem 1). However, in order to ensure the optimism of the chosen super arm, it is necessary to guarantee the optimism of all individual arms in the chosen super arm, which is represented in Corollary~\ref{cor:ucb-optimism}.

The following technical lemma bounds the sum of weighted norms which is similar to Lemma~4.2 in \citet{qin2014contextual} and Lemma~5.4 in \citet{zhou2020neural}.
\begin{lemma}
\label{lemma_bound of sum of gradient norm}
For any $\ConfidenceParam \in (0,1)$ suppose the width of the neural network $\NetworkWidth$ satisfies Condition~\ref{cond_network width}. If $\StepSize \le C_1( \TotalRound \SelectAction \NetworkWidth \NetworkDepth + \NetworkWidth \RegularizationParam)^{-1}$, and $\RegularizationParam \ge C_2 \NetworkDepth \SelectAction$, for some positive constant $C_1, C_2$ with $C_2 \ge \sqrt{ \max_{t, i} \| \Gradient(\Context_{t,i}; \NNParam_{t-1}) / \sqrt{\NetworkWidth} \|_2^2/\NetworkDepth}$,
then with probability at least $1-\ConfidenceParam$, for some $C_3>0$,
    \begin{align*}
        & \sum_{t=1}^\TotalRound \sum_{i \in S_t} \left\| \Gradient(\Context_{t,i}; \NNParam_{t-1}) /\sqrt{\NetworkWidth} \right\|_{\UCBGramMatrix_{t-1}^{-1}}^2 
        \\
        & \le 2 \EffectiveDim \log (1\!+\! \TotalRound \TotalAction/\RegularizationParam) + 2 + C_3  \TotalRound^{\frac{5}{3}} \SelectAction^{\frac{3}{2}} \NetworkDepth^4  \RegularizationParam\!^{-\frac{1}{6}} \NetworkWidth\!^{-\frac{1}{6}}\!\sqrt{ \log \NetworkWidth} \, .
    \end{align*}
\end{lemma}
Combining these results, we can derive the regret bound in Theorem~\ref{thm_CN-UCB}. First, using the Lipschitz continuity of the reward function,
we bound the instantaneous regret with the sum of scores for each individual arm within the super arm. 
By Lemma~\ref{lemma_u'_modified UCB}, the upper confidence bound of the over-parametrized neural network concentrates well around the true score function.
By adding an arm-independent offset term, we can ensure the optimism of the surrogate upper confidence bound. 
Then, we apply Lemma~\ref{lemma_bound of sum of gradient norm} to derive the desired cumulative regret bound.


\section{Combinatorial Neural TS ($\CNTS$)} \label{sec:CN-TS}

\subsection{Challenges in Worst-Case Regret Analysis for Combinatorial Actions}\label{sec:ts_challenges}

The challenges in the worst-case (non-Bayesian) regret analysis for TS algorithms with combinatorial actions lie in the difficulty of ensuring optimism of a sampled combinatorial action. 
The key analytical element to drive a sublinear regret for any TS algorithm, either combinatorial or non-combinatorial, is to show that a sampled action is optimistic with sufficient frequency \cite{agrawal2013thompson, abeille2017linear}. With combinatorial actions, however, ensuring optimism becomes more challenging than single-action selection. 
In particular, if the structure of the reward and feedback model is not known, one can only resort to hoping that all of the sampled base arms in the chosen super arm $S_t$ are optimistic, i.e., the scores of all sampled base arms are higher than their expected scores. The probability of such an event can be exponentially small in the size of the super arm $K$.

For example, let the probability that the sampled score of the $i$-th arm is higher than the corresponding expected score be at least $\tilde{p}$, i.e., $\mathbb{P}(\tilde{v}_i > h(\mathbf{x}_i)) \ge \tilde{p}$.
If the sampled score of every arm is optimistic, by the monotonicity property of the reward function, the reward induced by the sampled scores would be larger than the reward induced by the expected score, i.e., $R(S, \tilde{\mathbf{v}}) \ge R(S, \mathbf{v}^*)$.
However, the probability of the event that all the $K$ sampled scores are higher than their corresponding expected scores would be in the order of $\tilde{p}^K$.  
Hence, the probability of such an event can be exponentially small in the size of the super arm~$K$.

Note that in the UCB exploration, one can ensure high-probability optimism even with combinatorial actions in a straightforward manner since action selection is deterministic. 
However, in TS with combinatorial actions, suitable random exploration with provable efficiency is much more challenging to guarantee.
This challenge is further exacerbated by the complex analysis based on neural networks that we consider in this work.

\subsection{$\CNTS$ Algorithm}
\begin{algorithm*}[t!]
  \caption{Combinatorial Neural Thompson Sampling ($\CNTS$)}
  \label{alg:CN-TS}
\begin{algorithmic}
    \STATE {\bfseries Input:} Number of rounds $\TotalRound$, regularization parameter $\RegularizationParam$, exploration variance $\ExplorationVariance$, step size $\StepSize$, network width $\NetworkWidth$, number of gradient descent steps $\GDSteps$, network depth $\NetworkDepth$, sample size $\TotalSampleNum$.
    \STATE {\bfseries Initialization: } Randomly initialize $\NNParam_0$  as described in Section~\ref{subsec:CN-UCB} and $\TSGramMatrix_0 = \RegularizationParam \mathbf{I}$
    \FOR{$t = 1, ..., T$}
    \STATE Observe $\{ \mathbf{x}_{t,i} \}_{i \in [\TotalAction]}$
    \STATE Compute $\sigma_{t, i}^2 = \RegularizationParam \Gradient(\Context_{t,i}; \NNParam_{t-1})^\top \TSGramMatrix_{t-1}^{-1} \Gradient(\Context_{t,i}; \NNParam_{t-1}) 
    / \NetworkWidth$ for each $i \in [\TotalAction]$
    \\
    \STATE Sample $\{ \tilde{\Score}_{t,i}^{(j)} \}_{j=1}^{\TotalSampleNum}$ independently from $\mathcal{N}( f(\Context_{t,i}; \NNParam_{t-1}), \ExplorationVariance^2 \sigma_{t,i}^2)$ for each $i \in [\TotalAction]$
    \STATE Compute $\tilde{\Score}_{t, i} = \max_{j} \tilde{\Score}_{t,i}^{(j)}$ for each $i \in [\TotalAction]$
    \STATE Let $S_t = \Oracle_{\mathcal{S}} (\tilde{\ScoreVector}_t + \boldsymbol{\TSErrorTerm})$ 
    \STATE Play super arm $S_t$ and observe $\{ \Score_{t, i} \}_{i \in S_t}$
    
    \STATE Update $\TSGramMatrix_{t} = \TSGramMatrix_{t-1} + \sum_{i \in S_t} \Gradient(\Context_{t,i} ; \NNParam_{t-1}) \Gradient (\Context_{t,i}; \NNParam_{t-1})^\top /\NetworkWidth$
    \STATE Update $\NNParam_{t}$ to minimize the loss~\eqref{eq_L2-loss} using gradient descent with $\StepSize$ for $\GDSteps$ times
  \ENDFOR
\end{algorithmic}
\end{algorithm*}

To address the challenge of TS exploration with combinatorial actions described above,
we present $\CNTS$, a neural network-based TS algorithm. We make two modifications from conventional TS for parametric bandits.
First, instead of maintaining an actual Bayesian posterior as in the canonical TS algorithms, 
$\CNTS$ is a generic randomized algorithm that 
samples rewards from a Gaussian distribution. 
The algorithm directly samples  estimated rewards from a Gaussian distribution, rather than sampling network parameters -- this modification is adapted from \citet{zhang2021neural}. 

Second, in order to ensure sufficient \textit{optimistic} sampling in combinatorial action space, we draw multiple $\TotalSampleNum$ independent score samples for each arm instead of drawing a single sample. 
Leveraging these multiple samples, we compute the most optimistic (the highest estimated) score for each arm. We demonstrate that implementing this modification effectively ensures the required optimism of samples, formalized in  Lemma~\ref{lemma_Optimistic sampling}.
The algorithm is summarized in Algorithm~\ref{alg:CN-TS}.
%

\subsection{Regret of $\CNTS$}
Under the same assumptions introduced in the analysis of $\CNUCB$, we present the worst-case regret bound for $\CNTS$ in Theorem~\ref{thm_CN-TS}. 
%
\begin{theorem} \label{thm_CN-TS}
Suppose Assumptions~\ref{assum:monotone}-\ref{assum:context} hold and $\NetworkWidth$ satisfies Condition~\ref{cond_network width}.
If we run \texttt{CN-TS} with 
\begin{align*}
    & \StepSize = \bar{C}_1 (\TotalRound \SelectAction \NetworkWidth \NetworkDepth + \NetworkWidth \RegularizationParam)\, , \RegularizationParam = \max\{1 + 1/\TotalRound, \bar{C}_2 \NetworkDepth \SelectAction\},
    \\
    & \GDSteps = 2 \log \left( \sqrt{\RegularizationParam/\TotalRound\SelectAction\NetworkDepth} / (4 \bar{C}_3 \TotalRound) \right) \TotalRound \SelectAction \NetworkDepth/(\bar{C}_1 \RegularizationParam)
    \\
    & \ExplorationVariance = \NormParam + \SubGaussian \sqrt{ \EffectiveDim \log (1+ \TotalRound \TotalAction / \RegularizationParam) + 2 + 2 \log\TotalRound} \, ,
    \\
    & B = \max \{1/(22 e \sqrt{\pi}), \sqrt{2 \mathbf{h}^\top \mathbf{H} \mathbf{h}} \} \, ,
    \\
    & \TotalSampleNum = \lceil 1 - \log \SelectAction/\log(1-\OptimisticProbability) \rceil \, 
\end{align*}
for some positive constants $\bar{C}_1 > 0, \bar{C}_3 >0$, and $\bar{C}_2 \ge \sqrt{ \max_{t, i} \| \Gradient (\Context_{t,i}; \NNParam_{t-1}) / \sqrt{\NetworkWidth} \|_2^2/\NetworkDepth}$,
then the cumulative expected regret of \texttt{CN-TS} over horizon~$T$ is upper-bounded by
    \begin{equation*}
        \Rcal (T) = \tilde{\Ocal}( \tilde{d} \sqrt{T K}) \, .
    \end{equation*}
\end{theorem}
\noindent {\bf Discussion of Theorem~\ref{thm_CN-TS}.} Theorem~\ref{thm_CN-TS} establishes that the cumulative regret of $\CNTS$ is $\OTilde(\EffectiveDim\sqrt{\TotalRound  \SelectAction})$.
To the best of our knowledge, this is the first TS algorithm with the worst-case regret guarantees for general combinatorial action settings. 
This is crucial since various combinatorial bandit problems were prohibitive for TS methods due to the difficulty of ensuring the optimism of randomly selected super-action as discussed in Section~\ref{sec:ts_challenges}. 
Our result also encompasses the linear feedback model setting, for which, to our best knowledge, a worst-case regret bound has not been proven for TS with combinatorial actions in general.
\begin{remark} \label{remark_limitation of NTK}
Both $\CNUCB$ and $\CNTS$ depend on the condition of network size $\NetworkWidth$. 
However, our experiments show superior performances of the proposed algorithms even when they are implemented with much smaller $\NetworkWidth$ (see Section~\ref{sec:numerical exp}). 
The large value of $\NetworkWidth$ is 
sufficient
for regret analysis, due to the current state of the NTK theory. 
The same phenomenon is also present in the single action selection version of the neural bandits~\cite{zhang2021neural,zhou2020neural}.
\end{remark}
\begin{remark}
For a clear exposition of main ideas, the knowledge of $T$ is assumed for both $\CNUCB$ and $\CNTS$. 
This knowledge was also assumed in the previous neural bandit literature~\cite{zhang2021neural,zhou2020neural}. We can replace this requirement of knowledge on $T$ by using a doubling technique. We provide modified algorithms that do not depend on such knowledge of $T$ in Appendix~\ref{sec:unknown_T}.
\end{remark}
\begin{remark}
The proposed optimistic sampling technique can be applied to the regret analysis for TS algorithms with combinatorial actions other than neural bandit settings.
Regarding the cost of the optimistic sampling, this salient feature of the algorithm is controlled by the number of multiple samples $M$.
A notable feature is that while this technique provides provably sufficient optimism, the proposed optimistic sampling technique comes at a minimal cost of $\log M$. 
That is, even if we over-sample by the factor of 2, the additional cost in the regret bound only increases by the additive logarithmic factor, i.e., $\log 2 M = \log M + \log 2$.
Also, given that a theoretically suggested value of $M$ as shown in Theorem~\ref{thm_CN-TS} is only $\Omega(\log K)$, the regret caused by the optimistic sampling is of $\Ocal(\log \log K)$.
\end{remark}

\subsection{Proof Sketch of Theorem~\ref{thm_CN-TS}}
For any $t \in [T]$, we define events $\EventSampling_t$ and $\EventEstimating_t$ similar to the prior literature on TS \cite{agrawal2013thompson, zhang2021neural} defined as follows.
\begin{align*}
    & \EventSampling_t := \{ \omega \in \Filtration_{t+1} \mid \forall i, \left| \tilde{\Score}_{t, i} - f(\mathbf{x}_{t,i};\NNParam_{t-1}) \right| \le \SigmaEventConstant_t \ExplorationVariance \sigma_{t, i} \}
    \\
    & \EventEstimating_t := \{ \omega \in \Filtration_{t+1} \mid \forall i, \left| f(\mathbf{x}_{t,i}; \NNParam_{t-1}) - h(\mathbf{x}_{t,i}) \right| \le \ExplorationVariance \sigma_{t, i} + \TSErrorTerm \}
\end{align*}
where for some constants $\{C_{\TSErrorTerm, k} \}_{k=1}^{4}$, $\TSErrorTerm$ is defined as
\begin{align*}
    \TSErrorTerm &:= 
     C_{\TSErrorTerm, 1} \TotalRound^{\frac{2}{3}} \SelectAction^{\frac{2}{3}}  \NetworkDepth^{3} \RegularizationParam^{-\frac{2}{3}} \NetworkWidth^{-\frac{1}{6}} \sqrt{\log \NetworkWidth}
    \\
    &\quad + C_{\TSErrorTerm, 2} (1-\StepSize \NetworkWidth \RegularizationParam)^{\GDSteps/2} \sqrt{\TotalRound \SelectAction \NetworkDepth / \RegularizationParam} \nonumber
    \\
    &\quad + C_{\TSErrorTerm, 3} \TotalRound^{\frac{7}{6}} \SelectAction^{\frac{7}{6}} \NetworkDepth^{4} \RegularizationParam^{-\frac{7}{6}} \NetworkWidth^{-\frac{1}{6}} \sqrt{\log\NetworkWidth}(1+\sqrt{\TotalRound \SelectAction/\RegularizationParam}) \nonumber
    \\
    &\quad + C_{\TSErrorTerm, 4} \TotalRound^{\frac{7}{6}} \SelectAction^{\frac{7}{6}} \RegularizationParam^{-\frac{2}{3}} \NetworkDepth^{\frac{9}{2}} \NetworkWidth^{-\frac{1}{6}} \sqrt{\log \NetworkWidth}
    \\
    &\quad \cdot 
    \left( \NormParam + \SubGaussian \sqrt{ \EffectiveDim \log (1+ \TotalRound \TotalAction / \RegularizationParam) + 2 - 2\log\ConfidenceParam} \right) \, .
\end{align*}
Under event $\EventSampling_t$, the difference between the optimistic sampled score and the estimated score can be controlled by the score's approximate posterior variance.
Under the event $\EventEstimating_t$, the estimated score based on the neural network does not deviate far from the expected score up to the approximate error term. 
Note that both events $\EventEstimating_t, \EventSampling_t$ happen with high probability. 
The remaining part is a guarantee on the probability of optimism for randomly sampled actions.
Lemma~\ref{lemma_Optimistic sampling} shows that the proposed optimistic sampling ensures a constant probability of optimism. 
\begin{lemma} \label{lemma_Optimistic sampling}
Suppose we take optimistic samples of size $\TotalSampleNum = \lceil 1 - \frac{\log \SelectAction}{\log(1-\OptimisticProbability)} \rceil$ where $\OptimisticProbability:= 1/(4e\sqrt{\pi})$. Then we have
    \begin{equation*}
        \mathbb{P} \bigg( R (S_t, \tilde{\ScoreVector}_t + \boldsymbol{\TSErrorTerm}) > R (S_t^*, \ScoreVector_t^*) | \mathcal{F}_t, \EventEstimating_t \bigg) \ge \OptimisticProbability
    \end{equation*}
where 
$\boldsymbol{\TSErrorTerm} = [\TSErrorTerm, \ldots, \TSErrorTerm] \in \mathbb{R}^{\TotalAction}$.
\end{lemma}

Lemma~\ref{lemma_Optimistic sampling} implies that even in the worst case, our randomized action selection still provides optimistic rewards at least with constant frequency. Hence, the regret pertaining to random sampling can be upper-bounded based on this frequent-enough optimism.
The complete proof is deferred to Appendix~\ref{appx:sec_regret for CN-TS}. 


\section{Numerical Experiments} \label{sec:numerical exp}

In this section, we perform numerical evaluations on $\CNUCB$ and $\CNTS$.  
For each round in $\CNTS$, we draw $\TotalSampleNum=10$ samples for each arm.  
We also present the performances of $\CNTSOne$, which is a special case of $\CNTS$ drawing only one sample per arm.
We perform synthetic experiments and measure the cumulative regret of each algorithm.
In Experiment 1, we compare our algorithms with contextual combinatorial bandits based on a linear assumption: \texttt{CombLinUCB} and \texttt{CombLinTS}~\cite{wen2015efficient}.  
\begin{figure*}[t!]
\begin{center}
\centerline{\includegraphics[scale=0.4]{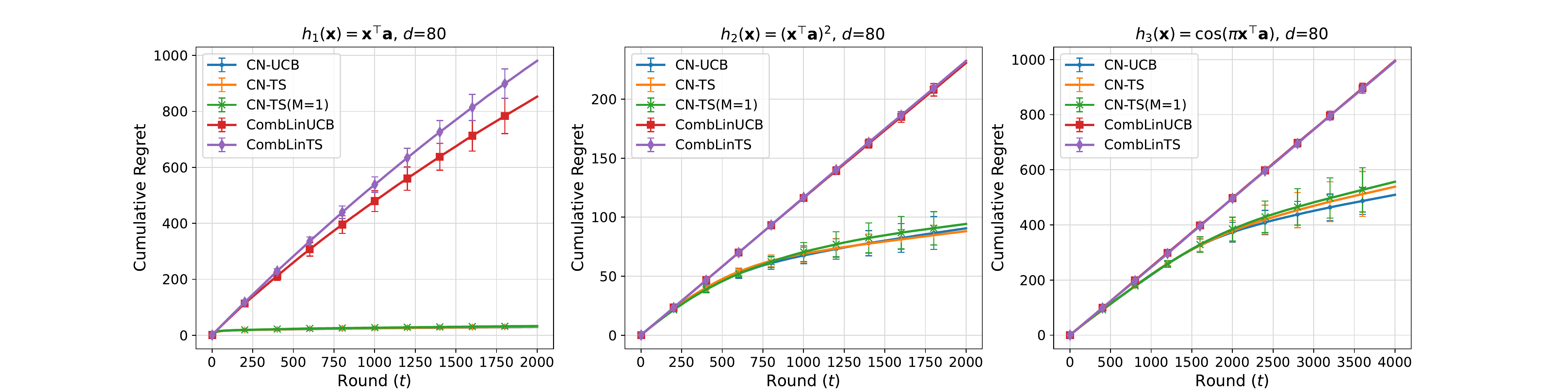}}
\caption{Cumulative regret of $\CNUCB$ and $\CNTS$ compared with algorithms based on linear models.}
\label{Fig1}
\end{center}
\end{figure*}

\begin{figure*}[ht]
\vskip -0.2in
\begin{center}
\centerline{\includegraphics[scale=0.4]{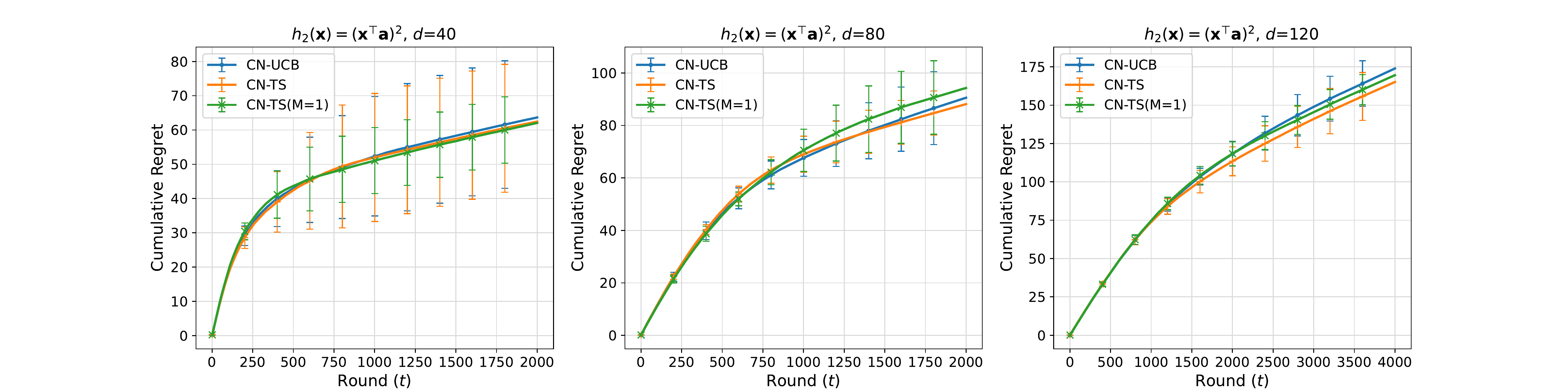}}
\caption{Experiment results of $\CNUCB$, $\CNTS$, and $\CNTSOne$ as context dimension $d$ increases.}
\label{Fig2}
\end{center}
\vskip -0.2in
\end{figure*}

In Experiment 2, we demonstrate the empirical performances of our algorithms as the context dimension $d$ increases.
The contexts given to the agent in each round are randomly generated from a unit ball.
The dimension of each context is $d=80$ for Experiment 1, and $d={40, 80, 120}$ for Experiment 2.
For each round, the agent of each algorithm chooses $\SelectAction=4$ arms among $\TotalAction=20$.

Similar to the experiments in \citet{zhou2020neural}, we assume three unknown score functions
\begin{align*}
        & h_{1}(\mathbf{x}_{t,i}) = \mathbf{x}_{t,i}^{\top}\mathbf{a}\, ,        
        \\
        & h_{2}(\mathbf{x}_{t,i}) = (\mathbf{x}_{t,i}^{\top}\mathbf{a})^{2}\, ,
        \\
        & h_{3}(\mathbf{x}_{t,i}) = \cos(\pi \mathbf{x}_{t,i}^{\top}\mathbf{a}) \, ,
\end{align*}
where $\mathbf{a}$ has the same dimension of the context and is randomly generated from a unit ball and remains fixed during the horizon.
We suppose a top-$\SelectAction$ problem and use the sum of scores $R(S_t, \ScoreVector_t) = \sum_{i \in S_t} \Score_{t, i}$ as the reward function.  
However, as mentioned in Remark~\ref{remark_extension of reward}, the reward function can be any function that satisfies Assumptions~\ref{assum:monotone} and \ref{assum:lips}.  
For example, $R(S_t, \ScoreVector_t)$ can be the quality of positions of a position-based click model \cite{BanditAlgo} or the expected revenue given by a multinomial logit (MNL) choice model~\cite{oh2019thompson} although the regret bound under the MNL choice model is not provided under the current theoretical result.

We use regularization parameter $\lambda = 1$ for all methods, 
confidence bound coefficient $\alpha = 1$ for \texttt{CombLinUCB} and $\gamma = 1$ for $\CNUCB$, and exploration variance $\nu = 1$ for $\CNTS$, $\CNTSOne$ and \texttt{CombLinTS}.
To estimate the score of each arm, we design a neural network with depth $\NetworkDepth=2$ and hidden layer width $\NetworkWidth=100$. 
The number of parameters is $p=\NetworkWidth d + \NetworkWidth= 8100$ for Experiment 1, and $p={4100, 8100, 12100}$ for Experiment 2.
The activation function is the rectified linear unit (ReLU).
We use the loss function in~Eq.\eqref{eq_L2-loss} and
use stochastic gradient descent with a batch of $100$ super arms. 
We train the neural network every $10$ rounds.
The training epoch is $100$, and the learning rate is $0.01$.

\noindent \textbf{Experiment 1.}
We evaluate the cumulative regret of the algorithms for each score function $h$. For score functions $h_{1}(\mathbf{x})$ and $h_{2}(\mathbf{x})$, we set the number of rounds, $T$, to 2000, while $T$ is set to 4000 for $h_{3}(\mathbf{x})$. We then present the average results, derived from 20 independent runs for each score function instance.
The results are depicted in Figure~\ref{Fig1}. 
Our proposed algorithms show significant improvements over those based on linear models. 
In contrast to linear baselines, the cumulative regrets for $\CNUCB$ and $\CNTS$ demonstrate a sub-linear trend, even when the score function is quadratic or non-linear. 
These findings suggest that our algorithms can be readily applied to a diverse range of complex reward functions.

\noindent \textbf{Experiment 2.}
We present the results of our proposed algorithms for context dimensions $d={40,80,120}$. 
To highlight the advantage of optimistic sampling, we show a comparison between $\CNTS$ and $\CNTSOne$. 
For these experiments, we utilize the quadratic score function $h_{2}(\mathbf{x})$. 
The number of rounds, $T$, is set to 2000 for $d=40, 80$ and 4000 for $d=120$. 
Similar to Experiment 1, the results represent averages derived from 20 independent runs.
Figure~\ref{Fig2} demonstrates the proficient performance of our algorithms, even as the feature dimension increases. 
The empirical results suggest a scalability of our algorithms in $d$ that is no greater than linear. Furthermore, when $d$ is large, $\CNTS$ exhibits a marginally lower cumulative regret compared to $\CNTSOne$. 
This observation substantiates our assertion that $\CNTS$ ensures a constant probability of optimism by drawing multiple $\TotalSampleNum$ samples.


\section{Conclusion}

In this paper, we study a general class of a contextual combinatorial bandit problem, where the model of the score function is unknown. 
Approximating the score function with deep neural networks, we propose two algorithms: $\CNUCB$ and $\CNTS$.  
We prove that $\texttt{CN-UCB}$ achieves $\tilde{\mathcal{O}}(\tilde{d} \sqrt{T})$ or $\tilde{\mathcal{O}}(\sqrt{\tilde{d} T K})$ regret.
For $\texttt{CN-TS}$, we adapt an optimistic sampling technique to ensure the optimism of the sampled combinatorial action, establish a worst-case (frequentist) regret of $\tilde{\mathcal{O}}(\tilde{d} \sqrt{TK})$.
To our knowledge, these are the first combinatorial neural bandit algorithms with sub-linear regret guarantees.
In particular, $\CNTS$ is the first general contextual combinatorial Thompson sampling algorithm  with the  worst-case regret guarantees.
Compared to the benchmark methods, our proposed methods exhibit consistent competitive performances, hence achieving both provable efficiency and practicality.


\section*{Acknowledgements}
This work was supported by the New Faculty Startup Fund from Seoul National University and the National Research Foundation of Korea(NRF) grant funded by the Korea government(MSIT) (No. 2022R1C1C1006859, No. 2022R1A4A103057912, No.~2021M3E5D2A01024795).


\bibliography{bib_arXiv}
\bibliographystyle{icml2023}

\newpage
\onecolumn
\makenomenclature

\mbox{}
\nomenclature[B]{$S_t$}{The super arm played at time $t$}
\nomenclature[B]{$S^*_t$}{The offline optimal super arm at time $t$}
\nomenclature[B]{$\Score_{t,i}^* = h(\Context_{t,i})$}{Expected score of arm $i$ at time $t$}
\nomenclature[B]{$R(S, \ScoreVector)$}{Reward for the super arm $S$ based on scores $\ScoreVector$}
\nomenclature[B]{$\TotalRound$}{The number of rounds}
\nomenclature[B]{$\TotalAction$}{The number of arms}
\nomenclature[B]{$\SelectAction$}{The size of super arm}
\nomenclature[B]{$\xi_{t,i}$}{Sub-Gaussian noise}
\nomenclature[B]{$\SubGaussian$}{Sub-Gaussian parameter}

\nomenclature[N]{$\NNParam_t$}{The parameter of the neural network at time $t$}
\nomenclature[N]{$\hat{\Score}_{t,i} = f(\Context_{t,i} ; \NNParam_{t-1})$}{Estimated score of arm $i$ at time $t$}
\nomenclature[N]{$\Gradient(\Context_{t,i};\NNParam_{t-1})$}{Gradient of the neural network for arm $i$ at time $t$}
\nomenclature[N]{$\NetworkWidth$}{The hidden layer width of the neural network}
\nomenclature[N]{$\NetworkDepth$}{The number of hidden layer of the neural network}
\nomenclature[N]{$p$}{The number of parameters of the neural network}
\nomenclature[N]{$\RegularizationParam$}{Regularziation parameter}
\nomenclature[N]{$\NTKMatrix$}{The Neural Tangent Kernel matrix}
\nomenclature[N]{$\EffectiveDim$}{The effective dimension}
\nomenclature[N]{$\RegularizationParamNTK$}{NTK matrix parameter}
\nomenclature[N]{$\StepSize$}{Step size for gradient descent}
\nomenclature[N]{$\GDSteps$}{The number of gradient descent steps}
\nomenclature[N]{$\NormParam$}{$ \ge \max \{ \sqrt{ 2\mathbf{h}^\top \NTKMatrix^{-1} \mathbf{h}} , 1/(22e\sqrt{\pi}) \} $}

\nomenclature[U]{$u_{t,i}$}{Upper confidence bound of the expected score for arm $i$ at time $t$}
\nomenclature[U]{$\gamma_t$}{Exploration scaling parameter for upper confidence bound}
\nomenclature[U]{$\UCBErrorTerm_t$}{Offset term added to the upper confidence bound at time $t$}
\nomenclature[U]{$\UCBGramMatrix_t$}{$= \RegularizationParam \mathbf{I} + \sum_{k=1}^t \sum_{i \in S_k} \Gradient(\Context_{k,i};\NNParam_{k-1}) \Gradient(\Context_{k,i};\NNParam_{k-1})^\top/\NetworkWidth$}
\nomenclature[U]{$\Gamma_{1,t}$}{$= \sqrt{1+C_{\Gamma,1} t^{\frac{7}{6}} \SelectAction \NetworkDepth^4 \RegularizationParam^{-\frac{7}{6}} \NetworkWidth^{-\frac{1}{6}} \sqrt{\log \NetworkWidth}}$}
\nomenclature[U]{$\Gamma_{2,t}$}{$ = C_{\Gamma, 2} t^{\frac{5}{3}} \SelectAction^{\frac{3}{2}} \NetworkDepth^4 \RegularizationParam^{-\frac{1}{6}} \NetworkWidth^{-\frac{1}{6}} \sqrt{\log \NetworkWidth}$}
\nomenclature[U]{$\Gamma_{3,t}$}{$ = C_{\Gamma, 3} t^{\frac{7}{6}} \SelectAction^{\frac{7}{6}} \NetworkDepth^{\frac{7}{2}} \RegularizationParam^{-\frac{7}{6}} \NetworkWidth^{-\frac{1}{6}} \sqrt{\log \NetworkWidth} (1+\sqrt{t \SelectAction / \RegularizationParam})$}

\nomenclature[T]{$\sigma_{t,i}^2$}{$= \RegularizationParam \Gradient(\Context_{t,i}; \NNParam_{t-1})^\top \TSGramMatrix_{t-1}^{-1} \Gradient(\Context_{t,i}; \NNParam_{t-1}) 
    / \NetworkWidth$}
\nomenclature[T]{$\tilde{\Score}_{t,i}^{(j)}$}{$j$-th sampled score generated by from distribution $\mathcal{N}\left(f(\Context_{t,i};\NNParam_{t-1}), \ExplorationVariance \sigma_{t,i}^2 \right)$}
\nomenclature[T]{$\tilde{\Score}_{t,i}$}{$=\max_{j} \tilde{\Score}_{t,i}^{(j)}$}
\nomenclature[T]{$\TSGramMatrix_t$}{$= \RegularizationParam \mathbf{I} + \sum_{k=1}^t \sum_{i \in S_k} \Gradient(\Context_{k,i};\NNParam_{k-1}) \Gradient(\Context_{k,i};\NNParam_{k-1})^\top/\NetworkWidth$}
\nomenclature[T]{$\ExplorationVariance$}{$= \NormParam + \SubGaussian \sqrt{ \EffectiveDim \log (1+ \TotalRound \TotalAction / \RegularizationParam) + 2 + 2 \log\TotalRound}$}
\nomenclature[T]{$\TotalSampleNum$}{$= \lceil 1 - \log  \SelectAction / \log(1-\OptimisticProbability) \rceil$}
\nomenclature[T]{$\OptimisticProbability$}{$=1/(4e\sqrt{\pi})$}
\nomenclature[T]{$\SigmaEventConstant_t$}{$= \sqrt{4 \log t + 2 \log \SelectAction + 2 \log \TotalSampleNum}$}
\nomenclature[T]{$\EventSampling_t$}{Event $ \forall i \in [N] : \mid \tilde{\Score}_{t, i} - f(\Context_{t,i};\NNParam_{t-1}) \mid \le \SigmaEventConstant_t \ExplorationVariance \sigma_{t, i} $}
\nomenclature[T]{$\EventEstimating_t$}{Event $\forall i \in [N]: \mid f(\Context_{t,i}; \NNParam_{t-1}) - h(\Context_{t,i}) \mid \le \ExplorationVariance \sigma_{t, i} + \TSErrorTerm$}
\nomenclature[T]{$\TSErrorTerm$}{Approximation error of the neural network}

\printnomenclature

\newpage
\appendix
\onecolumn

\addcontentsline{toc}{section}{Appendix} 
\part{Appendix} 
\parttoc 

\section{Regret Bound for $\CNUCB$} \label{appx: regret bound for CN-UCB}
In this section, we present all the necessary technical lemmas and their proof, followed by the proof of Theorem~\ref{thm_CN-UCB}.

\subsection{Proof of Lemma~\ref{lemma_u'_modified UCB}}
We introduce the following technical lemmas which are necessary for proof of Lemma~\ref{lemma_u'_modified UCB}.

\begin{lemma}[Lemma 5.1 in~\citet{zhou2020neural}]  \label{aux lemma:lemma 5.1 in NeuralUCB}
For any $\delta \in (0,1)$, suppose that there exists a positive constant $\bar{C}$ such that 
    \begin{equation*}
        m \ge \bar{C} T^4 N^4 L^6 \lambda_0^{-1} \log(T^2 N^2 L / \delta) \, .
    \end{equation*}
Then, with probability at least $1 - \delta$, there exists a $\thetab^* \in \RR^p$ such that for all $i \in [TN]$,
    \begin{align*}
        & h(\xb^k) = \gb(\xb^k; \thetab_0)^\top (\thetab^* - \thetab_0) \, ,
        \\
        & \sqrt{m} \| \thetab^* - \thetab_0 \|_2 \le \sqrt{\hb^\top \Hb^{-1} \hb} \, ,
    \end{align*}
where $\Hb$ is the NTK matrix defined in Definition~\ref{def_NTK matrix} and $\hb = [h(\xb^k)]_{k=1}^{TN}$.
\end{lemma}

\begin{lemma}[Lemma 4.1 in~\citet{cao2019generalization}] \label{aux lemma:Lemma B.4 in NeuralUCB}
Suppose that there exist $\bar{C}_1, \bar{C}_2 > 0$ such that for any $\ConfidenceParam \in (0,1)$, $\TwoNormCondition$ satisfies
    \begin{equation*}
        \bar{C}_1 \NetworkWidth^{-\frac{3}{2}} \NetworkDepth^{-\frac{3}{2}} \left( \log(\TotalRound \TotalAction \NetworkDepth^2 / \ConfidenceParam) \right)^{\frac{3}{2}}
        \le \TwoNormCondition 
        \le \bar{C}_2 \NetworkDepth^{-6} \left( \log \NetworkWidth \right)^{-\frac{3}{2}} \, .
    \end{equation*}
Then, with probability at least $1-\ConfidenceParam$, for all $\tilde{\NNParam}, \hat{\NNParam}$ satisfying $ \| \tilde{\NNParam} - \NNParam_0 \|_2 \le \TwoNormCondition$, $ \| \hat{\NNParam} - \NNParam_0 \|_2 \le \TwoNormCondition$  and $k \in \left[ \TotalRound \TotalAction \right]$, we have
    \begin{equation*}
        \left| f(\Context^k; \tilde{\NNParam}) - f(\Context^k; \hat{\NNParam}) - \Gradient(\Context^k ; \hat{\NNParam})^\top (\tilde{\NNParam} - \hat{\NNParam}) \right| \le \bar{C}_3 \TwoNormCondition^{\frac{4}{3}} \NetworkDepth^3 \sqrt{\NetworkWidth \log \NetworkWidth},
    \end{equation*}
where $\bar{C}_3 \ge 0$ is an absolute constant.
\end{lemma}

\begin{lemma}[Lemma 5 in~\citet{allenzhu2019convergence}] \label{aux lemma:lemma B.5 in NeuralUCB}
For any $\delta \in (0,1)$, suppose that there exist $\bar{C}_1, \bar{C}_2 > 0$ such that if $\tau$ satisfies
    \begin{equation*}
        \bar{C}_1 m^{-\frac{3}{2}} L^{-\frac{3}{2}} \max \left\{ (\log m)^{-\frac{3}{2}}, \left( \log(TN/\delta) \right)^{\frac{3}{2}} \right\}
        \le \tau \le
        \bar{C}_2 L^{-\frac{9}{2}} (\log m)^{-3} \, .
    \end{equation*}
Then, with probability at least $1 - \delta$, for all $\thetab$ satisfying $\| \thetab - \thetab_0 \|_2 \le \tau$ and $k \in [TN]$ we have
    \begin{equation*}
        \| \gb(\xb^k ; \thetab) - \gb(\xb^k ;\thetab_0) \|_2 \
        \le 
        \bar{C}_3 \sqrt{\log m} \tau^{\frac{1}{3}} L^3 \| \gb(\xb^k; \thetab_0) \|_2 \, ,
    \end{equation*}
where $\bar{C}_3 >0$ is an absolute constant.
\end{lemma}

\begin{lemma}[Lemma B.3 in~\citet{cao2019generalization}] \label{aux lemma:lemma B.6 in NeuralUCB}
Suppose that there exist $\bar{C}_1, \bar{C}_2 >0 $ such that for any $\ConfidenceParam \in (0,1)$, $\TwoNormCondition$ satisfies 
    \begin{equation*}
        \bar{C}_1 \NetworkWidth^{-\frac{3}{2}} \NetworkDepth^{-\frac{3}{2}} \left( \log(\TotalRound \TotalAction \NetworkDepth^2 / \ConfidenceParam) \right)^{\frac{3}{2}}
        \le \TwoNormCondition \le
        \bar{C}_2 \NetworkDepth^{-6} \left( \log \NetworkWidth \right)^{-\frac{3}{2}}.        
    \end{equation*}
Then with probability at least $1-\ConfidenceParam$, for any $\NNParam$ satisfying $\| \NNParam - \NNParam_0 \|_2 \le \TwoNormCondition$ and $k \in [\TotalRound \TotalAction]$ we have
    \begin{equation*}
        \| \Gradient(\Context^k; \NNParam) \|_F \le \bar{C}_3 \sqrt{\NetworkWidth \NetworkDepth}
    \end{equation*}
where $\bar{C}_3 > 0$ is an absolute constant.
\end{lemma}

\begin{proof}[Proof of Lemma~\ref{lemma_u'_modified UCB}]
First of all, note that because $\NetworkWidth$ satisfies Condition~\ref{cond_network width}, the required conditions in Lemma~\ref{aux lemma:lemma 5.1 in NeuralUCB}-\ref{aux lemma:lemma B.6 in NeuralUCB} are satisfied. 
For any $t \in [T], i \in [N]$, by definition of $u_{t,i}$ and $\Score_{t,i}^*$, we have
    \begin{align*}
        \left| u_{t,i} - \Score_{t,i}^* \right| & = 
        \left| f (\Context_{t,i}; \NNParam_{t-1}) + \gamma_{t-1} \left\| \Gradient (\Context_{t,i};\NNParam_{t-1}) / \sqrt{\NetworkWidth} \right\|_{\UCBGramMatrix_t^{-1}} - h(\Context_{t,i}) \right|
        \\
        & = \left| f (\Context_{t,i};\NNParam_{t-1}) + \gamma_{t-1} \left\| \Gradient (\Context_{t,i};\NNParam_{t-1}) / \sqrt{\NetworkWidth} \right\|_{\UCBGramMatrix_{t-1}^{-1}} - \Gradient (\Context_{t,i}; \NNParam_0)^\top (\NNParam^* - \NNParam_0) \right|
        \\
        & \le \underbrace{ \left| f (\Context_{t,i}; \NNParam_{t-1}) - \Gradient (\Context_{t,i}; \NNParam_0)^\top (\NNParam^* - \NNParam_0) \right|}_{I_0} 
        + \underbrace{\gamma_{t-1} \left\| g (\Context_{t,i}; \NNParam_{t-1}) / \sqrt{\NetworkWidth}\right\|_{\UCBGramMatrix_{t-1}^{-1}}}_{I_1}
    \end{align*}
where the second equality holds due to Lemma~\ref{aux lemma:lemma 5.1 in NeuralUCB}, and the inequality follows from the triangle inequality.

For $I_0$, we have 
    \begin{align}
        I_0 & =  \left| f (\Context_{t,i} ; \NNParam_{t-1}) - \Gradient (\Context_{t,i};\NNParam_0)^\top (\NNParam^* - \NNParam_0 + \NNParam_{t-1} - \NNParam_{t-1}) \right| \nonumber
        \\
        & = \left| f (\Context_{t,i};\NNParam_{t-1}) - f (\Context_{t,i};\NNParam_0) - \Gradient (\Context_{t,i}; \NNParam_0)^\top (\NNParam_{t-1} - \NNParam_0) - \Gradient (\Context_{t,i}; \NNParam_0)^\top (\NNParam^* - \NNParam_{t-1}) \right| \nonumber
        \\
        & \le \underbrace{ \left| f (\Context_{t,i}; \NNParam_{t-1}) - f (\Context_{t,i};\NNParam_0) - \Gradient (\Context_{t,i}; \NNParam_0)^\top (\NNParam_{t-1} - \NNParam_0) \right|}_{I_2} 
        + \underbrace{ \left| \Gradient (\Context_{t,i}; \NNParam_0)^\top (\NNParam^* - \NNParam_{t-1}) \right| }_{I_3} \label{eq:surrogate ucb eq 1}
    \end{align}
where the second equality holds due to the initial condition of $f$, i.e., $f(\xb; \NNParam_0) = 0$ for all $\xb$, and the inequality comes from the triangle inequality.

To bound $I_2$, we have
    \begin{align*}
        I_2 & = \left| f (\Context_{t,i};\NNParam_{t-1}) - f (\Context_{t,i}; \NNParam_0) - \Gradient (\Context_{t,i}; \NNParam_0)^\top (\NNParam_{t-1} - \NNParam_0)  \right|
        \\
        & \le C'_3 \TwoNormCondition^{\frac{4}{3}} \NetworkDepth^3 \sqrt{ \NetworkWidth \log \NetworkWidth }
        \\
        & = C_3 t^{\frac{2}{3}} \SelectAction^{\frac{2}{3}} \RegularizationParam^{-\frac{2}{3}} \NetworkWidth^{-\frac{1}{6}} \sqrt{ \log \NetworkWidth}
    \end{align*}
where the first inequality follows from Lemma~\ref{aux lemma:Lemma B.4 in NeuralUCB} for some constant $C'_3>0$, and the second equality is due to setting $\TwoNormCondition$ of Lemma~\ref{aux lemma:Lemma B.4 in NeuralUCB} as $2\sqrt{t \SelectAction /(\NetworkWidth \RegularizationParam)}$ of Lemma~\ref{extended lemma:lemma 5.2 in NeuralUCB}, i.e., $\TwoNormCondition = 2\sqrt{t \SelectAction /(\NetworkWidth \RegularizationParam)}$.

To bound $I_3$, we have
    \begin{align*}
        I_3 & = \left| \Gradient (\Context_{t,i}; \NNParam_0)^\top (\NNParam^* - \NNParam_{t-1}) \right|
        \le \left\| \Gradient (\Context_{t,i}; \NNParam_0) \right\|_{\UCBGramMatrix_{t-1}^{-1}} \left\|\NNParam^* - \NNParam_{t-1} \right\|_{\UCBGramMatrix_{t-1}}
        \le \frac{\gamma_{t-1}}{\sqrt{\NetworkWidth}} \left\| \Gradient (\Context_{t,i}; \NNParam_0) \right\|_{\UCBGramMatrix_{t-1}^{-1}} 
    \end{align*}
where the first inequality holds due to the Cauchy-Schwarz inequality, and the second inequality follows from Lemma~\ref{extended lemma:lemma 5.2 in NeuralUCB}.

Combining the results, we have
    \begin{align*}
        \left| u_{t,i} - \Score_{t,i}^* \right| & \le I_2 + I_3 + I_1
        \\
        & \le C_3 t^{\frac{2}{3}} \SelectAction^{\frac{2}{3}} \RegularizationParam^{-\frac{2}{3}} \NetworkWidth^{-\frac{1}{6}} \sqrt{ \log \NetworkWidth} 
        + \frac{\gamma_{t-1}}{\sqrt{\NetworkWidth}} \left\| \Gradient (\Context_{t,i}; \NNParam_0) \right\|_{\UCBGramMatrix_{t-1}^{-1}} 
        + \frac{\gamma_{t-1}}{\sqrt{\NetworkWidth}} \left\| g (\Context_{t,i};\NNParam_{t-1}) \right\|_{\UCBGramMatrix_{t-1}^{-1}}
        \\
        & = C_3 t^{\frac{2}{3}} \SelectAction^{\frac{2}{3}} \RegularizationParam^{-\frac{2}{3}} \NetworkWidth^{-\frac{1}{6}} \sqrt{ \log \NetworkWidth} 
        + \frac{\gamma_{t-1}}{\sqrt{\NetworkWidth}} \underbrace{ \left(
        \left\| \Gradient (\Context_{t,i}; \NNParam_0) \right\|_{\UCBGramMatrix_{t-1}^{-1}} +
        \left\| \Gradient (\Context_{t,i}; \NNParam_{t-1}) \right\|_{\UCBGramMatrix_{t-1}^{-1}} \right)}_{I_4} \,.
    \end{align*}
Now $I_4$ can be bounded as
    \begin{align*}
        I_4 & = \left\| \Gradient (\Context_{t,i}; \NNParam_0) + \Gradient (\Context_{t,i};\NNParam_{t-1}) - \Gradient (\Context_{t, i};\NNParam_{t-1}) \right\|_{\UCBGramMatrix_{t-1}^{-1}} +
        \left\| \Gradient (\Context_{t,i};\NNParam_{t-1}) \right\|_{\UCBGramMatrix_{t-1}^{-1}}
        \\
        & \le \left\| \Gradient (\Context_{t,i};\NNParam_0) - \Gradient (\Context_{t,i};\NNParam_{t-1}) \right\|_{\UCBGramMatrix_{t-1}^{-1}} 
        + 2 \left\| \Gradient (\Context_{t,i};\NNParam_{t-1}) \right\|_{\UCBGramMatrix_{t-1}^{-1}}
        \\
        & \le \frac{1}{\sqrt{\RegularizationParam}} \left\| \Gradient (\Context_{t,i}; \NNParam_0) - \Gradient (\Context_{t,i}; \NNParam_{t-1}) \right\|_2 
        + 2 \left\| \Gradient (\Context_{t,i}; \NNParam_{t-1}) \right\|_{\UCBGramMatrix_{t-1}^{-1}}
        \\
        & \le \frac{1}{ \sqrt{\RegularizationParam}} C'_2 \sqrt{ \log \NetworkWidth } \left( 2 \sqrt{t \SelectAction /(\NetworkWidth \RegularizationParam)} \right)^{\frac{1}{3}} \NetworkDepth^3 \left\| \Gradient (\Context_{t,i}; \NNParam_0) \right\|_2
        + 2 \left\| \Gradient (\Context_{t,i}; \NNParam_{t-1}) \right\|_{\UCBGramMatrix_{t-1}^{-1}}
        \\
        & \le C_2 t^{\frac{1}{6}} \SelectAction^{\frac{1}{6}} \RegularizationParam^{-\frac{2}{3}} \NetworkDepth^{\frac{7}{2}} \NetworkWidth^{\frac{1}{3}} \sqrt{\log \NetworkWidth}
        + 2 \left\| \Gradient (\Context_{t,i}; \NNParam_{t-1}) \right\|_{\UCBGramMatrix_{t-1}^{-1}}
    \end{align*}
where the first inequality follows from the triangle inequality, the second inequality holds due to the property $\left\| \Context \right\|_{\UCBGramMatrix_{t-1}^{-1}} \le \frac{1}{\sqrt{\RegularizationParam}} \left\| \Context \right\|_2$, the third inequality follows from Lemma~\ref{aux lemma:lemma B.5 in NeuralUCB} with $\TwoNormCondition = 2\sqrt{t \SelectAction /(\NetworkWidth \RegularizationParam)}$ in Lemma~\ref{extended lemma:lemma 5.2 in NeuralUCB}, and the last inequality holds due to Lemma~\ref{aux lemma:lemma B.6 in NeuralUCB}.
\\
Finally, by taking a union bound about $\ConfidenceParam$, with probability at least $1-5\ConfidenceParam$, we have
    \begin{align*}
        \left| u_{t,i} - \Score_{t,i}^* \right| & 
        \le C_3 t^{\frac{2}{3}} \SelectAction^{\frac{2}{3}} \RegularizationParam^{-\frac{2}{3}} \NetworkWidth^{-\frac{1}{6}} \sqrt{ \log \NetworkWidth}  
        + \frac{\gamma_{t-1}}{\sqrt{\NetworkWidth}} I_4
        \\
        & \le 2 \gamma_{t-1} \left\| \Gradient(\Context_{t,i};\NNParam_{t-1}) / \sqrt{\NetworkWidth} \right\|_{\UCBGramMatrix_{t-1}^{-1}} 
        + C_2 \gamma_{t-1} t^{\frac{1}{6}} \SelectAction^{\frac{1}{6}} \RegularizationParam^{-\frac{2}{3}} \NetworkDepth^{\frac{7}{2}} \NetworkWidth^{-\frac{1}{6}} \sqrt{\log \NetworkWidth}
        \\
        & \phantom{{}={}} + C_3 t^{\frac{2}{3}} \SelectAction^{\frac{2}{3}} \RegularizationParam^{-\frac{2}{3}} \NetworkWidth^{-\frac{1}{6}} \sqrt{ \log \NetworkWidth} \, .
    \end{align*}
\\
In particular, if we define
    \begin{equation*}
        e_t = C_2 \gamma_{t-1} t^{\frac{1}{6}} \SelectAction^{\frac{1}{6}} \RegularizationParam^{-\frac{2}{3}} \NetworkDepth^{\frac{7}{2}} \NetworkWidth^{-\frac{1}{6}} \sqrt{\log \NetworkWidth}
        + C_3 t^{\frac{2}{3}} \SelectAction^{\frac{2}{3}} \RegularizationParam^{-\frac{2}{3}} \NetworkWidth^{-\frac{1}{6}} \sqrt{ \log \NetworkWidth}
    \end{equation*}
and we replace $\delta$ with $\delta/5$, then we have the desired result.
\end{proof}

\subsection{Proof of Corollary~\ref{cor:ucb-optimism}}
\begin{proof}[Proof of Corollary~\ref{cor:ucb-optimism}]
Suppose that Lemma~\ref{lemma_u'_modified UCB} holds. 
Let us denote
    \begin{align*}
        \bar{u}_{t,i} = u_{t,i} + 
        \underbrace{ C_2 \gamma_{t-1} t^{\frac{1}{6}} \SelectAction^{\frac{1}{6}} \RegularizationParam^{-\frac{2}{3}} \NetworkDepth^{\frac{7}{2}} \NetworkWidth^{-\frac{1}{6}} \sqrt{\log \NetworkWidth}
        + C_3 t^{\frac{2}{3}} \SelectAction^{\frac{2}{3}} \RegularizationParam^{-\frac{2}{3}} \NetworkWidth^{-\frac{1}{6}} \sqrt{ \log \NetworkWidth}}_{\UCBErrorTerm_t} \, .
    \end{align*}
Then, we have

    \begin{align*}
        \bar{u}_{t,i} - \Score_{t,i}^* & =  f (\Context_{t,i}; \NNParam_{t-1}) + \gamma_{t-1} \left\| \Gradient (\Context_{t,i}; \NNParam_{t-1}) / \sqrt{\NetworkWidth} \right\|_{\UCBGramMatrix_{t-1}^{-1}} 
        + \UCBErrorTerm_t - \Gradient (\Context_{t,i} ; \NNParam_0)^\top (\NNParam^* - \NNParam_0)
        \\
        & \ge - \underbrace{\left| f (\Context_{t,i};\NNParam_{t-1}) - \Gradient (\Context_{t,i};\NNParam_0)^\top (\NNParam^* - \NNParam_0) \right|}_{I_0}
        + \gamma_{t-1} \left\| \Gradient (\Context_{t,i};\NNParam_{t-1}) / \sqrt{\NetworkWidth} \right\|_{\UCBGramMatrix_{t-1}^{-1}} 
        + \UCBErrorTerm_t
        \\
        & \ge - \underbrace{C_3 t^{\frac{2}{3}} \SelectAction^{\frac{2}{3}} \RegularizationParam^{-\frac{2}{3}} \NetworkWidth^{-\frac{1}{6}} \sqrt{ \log \NetworkWidth} }_{I_2}
        - \underbrace{\frac{\gamma_{t-1}}{\sqrt{\NetworkWidth}} \left\| \Gradient (\Context_{t,i}; \NNParam_0) \right\|_{\UCBGramMatrix_{t-1}^{-1}}}_{I_3}
        + \frac{\gamma_{t-1}}{\sqrt{\NetworkWidth}} \left\| \Gradient (\Context_{t,i};\NNParam_{t-1}) \right\|_{\UCBGramMatrix_{t-1}^{-1}} 
        \\
        & \phantom{{}={}} + \UCBErrorTerm_t + \frac{\gamma_{t-1}}{\sqrt{\NetworkWidth}} \left\| \Gradient (\Context_{t,i};\NNParam_{t-1}) \right\|_{\UCBGramMatrix_{t-1}^{-1}} 
        - \frac{\gamma_{t-1}}{\sqrt{\NetworkWidth}} \left\| \Gradient (\Context_{t,i};\NNParam_{t-1}) \right\|_{\UCBGramMatrix_{t-1}^{-1}} 
        \\
        & = - C_3 t^{\frac{2}{3}} \SelectAction^{\frac{2}{3}} \RegularizationParam^{-\frac{2}{3}} \NetworkWidth^{-\frac{1}{6}} \sqrt{ \log \NetworkWidth} 
        + 2 \frac{\gamma_{t-1}}{\sqrt{\NetworkWidth}} \left\| \Gradient (\Context_{t,i}; \NNParam_{t-1}) \right\|_{\UCBGramMatrix_{t-1}^{-1}}
        + \UCBErrorTerm_t
        \\
        & \phantom{{}={}} - \frac{\gamma_{t-1}}{\sqrt{\NetworkWidth}} \underbrace{\left( \left\| \Gradient (\Context_{t,i}; \NNParam_0) \right\|_{\UCBGramMatrix_{t-1}^{-1}} +\left\| \Gradient (\Context_{t,i};\NNParam_{t-1}) \right\|_{\UCBGramMatrix_{t-1}^{-1}} \right)}_{I_4}
        \\
        & \ge 0 \, ,
    \end{align*}
where the first equation comes from Lemma~\ref{aux lemma:lemma 5.1 in NeuralUCB} and the second inequality follows from~Eq.\eqref{eq:surrogate ucb eq 1}.
\end{proof}

\subsection{Proof of Lemma~\ref{lemma_bound of sum of gradient norm}}
The following lemma is necessary for our proof.
\begin{lemma}[Lemma B.7 in~\citet{zhang2021neural}] \label{aux lemma:Lemma B.7 in NeuralTS}
For any $t \in [\TotalRound]$, suppose that there exists $\bar{C} > 0$ such that the network width $\NetworkWidth$ satisfies
    \begin{equation*}
        \NetworkWidth \ge \bar{C} \TotalRound^6 N^6 L^6 \log( \TotalRound \NetworkDepth \TotalAction / \ConfidenceParam).
    \end{equation*}
Then with probability at least $1-\ConfidenceParam$,
    \begin{equation*}
        \log \det (\mathbf{I} + \RegularizationParam^{-1} \mathbf{K}_t )
        \le \log \det (\mathbf{I} + \RegularizationParam^{-1} \NTKMatrix) + 1 \, ,
    \end{equation*}
where $\mathbf{K}_t = \bar{\mathbf{J}}_t^\top \bar{\mathbf{J}}_t / \NetworkWidth$, $\bar{\mathbf{J}}_t = \left[ \Gradient (\Context_{1, a_{11}}; \NNParam_{0}), \cdots , \Gradient (\Context_{t, a_{tK}};\NNParam_{0}) \right] \in \mathbb{R}^{p \times t\SelectAction}$, and $a_{tk}$ means $k$-th action in the super arm $S_t$ at time $t$, i.e., $S_t:= \{a_{t1}, \ldots, a_{tK}\}$.
\end{lemma}

\begin{proof}
Note that we have
    \begin{align*}
        \det (\UCBGramMatrix_{\TotalRound}) & = \det \left( \UCBGramMatrix_{\TotalRound-1} + \sum_{i \in S_\TotalRound} \Gradient (\Context_{\TotalRound, i};\NNParam_{\TotalRound-1}) \Gradient (\Context_{\TotalRound, i}; \NNParam_{\TotalRound-1})^\top / \NetworkWidth \right)
        \\
        & = \det \left( \UCBGramMatrix_{\TotalRound-1}^{\frac{1}{2}} 
        \left( \mathbf{I} + \sum_{i \in S_\TotalRound} \UCBGramMatrix_{\TotalRound-1}^{-\frac{1}{2}} (\Gradient (\Context_{\TotalRound, i} ; \NNParam_{\TotalRound-1})/\sqrt{\NetworkWidth}) (\Gradient (\Context_{\TotalRound, i}; \NNParam_{\TotalRound-1}) / \sqrt{\NetworkWidth})^\top \UCBGramMatrix_{\TotalRound-1}^{-\frac{1}{2}}
        \right) \UCBGramMatrix_{\TotalRound-1}^{\frac{1}{2}} \right)
        \\
        & = \det ( \UCBGramMatrix_{\TotalRound-1}) \cdot
        \det \left( \mathbf{I} +\sum_{i \in S_\TotalRound} \left( \UCBGramMatrix_{\TotalRound-1}^{-\frac{1}{2}} \Gradient (\Context_{\TotalRound, i}; \NNParam_{\TotalRound-1})/\sqrt{\NetworkWidth} \right) \left( \UCBGramMatrix_{\TotalRound-1}^{-\frac{1}{2}} \Gradient (\Context_{\TotalRound, i}; \NNParam_{\TotalRound-1}) / \sqrt{\NetworkWidth} \right)^\top \right)
        \\
        & = \det ( \UCBGramMatrix_{\TotalRound-1}) \cdot \left(1 + \sum_{i \in S_\TotalRound} \left\| \Gradient ( \Context_{\TotalRound, i} ; \NNParam_{\TotalRound-1}) / \sqrt{\NetworkWidth} \right\|_{\UCBGramMatrix_{\TotalRound-1}^{-1}}^2 \right) 
        \\
        & = \det (\UCBGramMatrix_0) \prod_{t=1}^{\TotalRound} \left(1 + \sum_{i \in S_t} \left\| \Gradient( \Context_{t, i};\NNParam_{t-1}) / \sqrt{\NetworkWidth} \right\|_{\UCBGramMatrix_{t-1}^{-1}}^2 \right) \, .
    \end{align*}
Then, we have
    \begin{equation*} 
        \log \frac{\det(\UCBGramMatrix_{\TotalRound})}{\det (\UCBGramMatrix_0)} = \sum_{t=1}^{\TotalRound} \log \left( 1 + \sum_{i \in S_t} \left\| \Gradient ( \Context_{t, i};\NNParam_{t-1}) / \sqrt{\NetworkWidth} \right\|_{\UCBGramMatrix_{t-1}^{-1}}^2 \right) \, .
    \end{equation*}
On the other hand, for any $t \in [T]$, we have
    \begin{align*}
        \sum_{i \in S_t} \left\| \Gradient ( \Context_{t, i}; \NNParam_{t-1}) / \sqrt{\NetworkWidth} \right\|_{\UCBGramMatrix_{t-1}^{-1}}^2
        & \le \sum_{i \in S_t} \frac{1}{\RegularizationParam} \left\| \Gradient ( \Context_{t, i}; \NNParam_{t-1}) \right\|_2^2 / \NetworkWidth
        \\
        & \le \sum_{i \in S_t} \frac{1}{\RegularizationParam \NetworkWidth} \left( C_2 \sqrt{\NetworkWidth \NetworkDepth} \right)^2 
        \\
        & \le 1 \, ,
    \end{align*}
where the first inequality comes from the property $\left\| \Context \right\|_{\mathbf{A}^{-1}}^2 \le  \left\| \Context \right\|_2^2 / \lambda_{\text{min}}(\mathbf{A})$ for any positive definite matrix $A$, the constant $C_2$ of the second inequality can be derived by Lemma~\ref{aux lemma:lemma B.6 in NeuralUCB}, and the last inequality holds due to the assumption of $\RegularizationParam$. Then using the inequality, $x \le 2 \log(1+x)$ for any $x \in [0,1]$, we have
    \begin{align*}
        \sum_{t=1}^{\TotalRound} \sum_{i \in S_t} \left\| \Gradient ( \Context_{t, i};\NNParam_{t-1}) / \sqrt{\NetworkWidth} \right\|_{\UCBGramMatrix_{t-1}^{-1}}^2 
        & \le 2 \sum_{t=1}^{\TotalRound} \log \left( 1 + \sum_{i \in S_t} \left\| \Gradient ( \Context_{t, i};\NNParam_{t-1}) / \sqrt{\NetworkWidth} \right\|_{\UCBGramMatrix_{t-1}^{-1}}^2 \right)
        \\
        & \le 2 \left| \log \frac{\det \UCBGramMatrix_{\TotalRound}}{\det \RegularizationParam \mathbf{I}} \right|
        \\
        & \le 2 \left| \log \frac{\det \bar{\UCBGramMatrix}_{\TotalRound}}{\det \RegularizationParam \mathbf{I}} \right|
        + C_3 \TotalRound^{\frac{5}{3}} \SelectAction^{\frac{5}{3}} \NetworkDepth^{4} \RegularizationParam^{-\frac{1}{6}} \NetworkWidth^{-\frac{1}{6}} \sqrt{\log \NetworkWidth} \, ,
    \end{align*}
where the last inequality holds due to Lemma~\ref{extended lemma:lemma B.3 in NeuralUCB} for some $C_3>0$.
Furthermore, since we have
    \begin{align}
         \log \frac{\det \bar{\UCBGramMatrix}_{\TotalRound}}{\det \RegularizationParam \mathbf{I}} 
         & = \log \det \left( \bar{\UCBGramMatrix}_{\TotalRound} \left( \RegularizationParam \mathbf{I} \right)^{-1} \right) \nonumber
         \\
         & = \log \det \left( \mathbf{I} + \sum_{t=1}^{\TotalRound} \sum_{i \in S_t} \Gradient (\Context_{t,i};\NNParam_0) \Gradient (\Context_{t,i};\NNParam_0)^\top / (\NetworkWidth \RegularizationParam) \right) \nonumber
         \\
         & = \log \det \left( \mathbf{I} +  \RegularizationParam^{-1} \bar{\mathbf{J}}_{\TotalRound} \bar{\mathbf{J}}_{\TotalRound}^\top / \NetworkWidth \right) \nonumber
         \\
         &
         = \log \det \left( \mathbf{I} + \RegularizationParam^{-1} \bar{\mathbf{J}}_{\TotalRound}^\top \bar{\mathbf{J}}_{\TotalRound} / \NetworkWidth \right) \nonumber
         \\
         & = \log \det \left( \mathbf{I} + \RegularizationParam^{-1} \mathbf{K}_\TotalRound \right) \nonumber 
         \\
         & \le \log \det \left( \mathbf{I} + \RegularizationParam^{-1} \NTKMatrix \right) + 1 \nonumber
         \\
         & = \EffectiveDim \log (1 + \TotalRound \TotalAction / \RegularizationParam ) + 1 \, , \label{eq_logdet Z_bar_T / Z0} 
    \end{align}
where the first, second equation and the first inequality holds naively, the third equality uses the definition of $\bar{\mathbf{J}}_t$, the fourth equality holds since for any matrix $\mathbf{A} \in \mathbb{M}_n(\mathbb{R})$ the nonzero eigenvalues of $\mathbf{I} + \mathbf{A}\mathbf{A}^\top$ and $\mathbf{I} + \mathbf{A}^\top \mathbf{A}$ are same, which means $\det (\mathbf{I} + \mathbf{A}\mathbf{A}^\top) = \det (\mathbf{I} + \mathbf{A}^\top \mathbf{A})$, the first inequality follows from Lemma~\ref{aux lemma:Lemma B.7 in NeuralTS}, and the last equality uses the definition of effective dimension in Definition~\ref{def_effective dimension}.
Finally, by taking a union bound about $\delta$, with probability at least $1 - 2 \delta$, we have
    \begin{equation*}
        \sum_{t=1}^\TotalRound \sum_{i \in S_t} \left\| \Gradient (\Context_{t,i};\NNParam_{t-1}) / \sqrt{\NetworkWidth}\right\|_{\UCBGramMatrix_{t-1}^{-1}}^2
        \le
            2 \EffectiveDim \log (1 + \TotalRound \TotalAction / \RegularizationParam ) + 2 
            + C_3 \TotalRound^{\frac{5}{3}} K^{\frac{5}{3}} \NetworkDepth^{4} \RegularizationParam^{-\frac{1}{6}} \NetworkWidth^{-\frac{1}{6}} \sqrt{\log \NetworkWidth} \, .
    \end{equation*}
By replacing $\delta$ with $\delta/2$, we have the desired result.
\end{proof}

\subsection{Proof of Theorem~\ref{thm_CN-UCB}} \label{subsec:proof of CN-UCB}
\begin{proof}[Proof of Theorem~\ref{thm_CN-UCB}]

We define the following event:
\begin{equation*}
    \begin{split}
        & \Efrak_1 := \left\{ | u_{t,i} - h(\xb_{t,i}) | \le 2 \gamma_{t-1} \| \gb (\xb_{t,i}; \thetab_{t-1}) / \sqrt{m} \|_{\Zb^{-1}_{t-1}} + e_t, \forall i \in [N], 1\le t \le T \right\} \, , 
        \\
        & \Efrak_2 := \left\{         
            \sum_{t=1}^\TotalRound \sum_{i \in S_t} \left\| \Gradient (\Context_{t,i};\NNParam_{t-1}) / \sqrt{\NetworkWidth}\right\|_{\UCBGramMatrix_{t-1}^{-1}}^2
            \le
            2 \EffectiveDim \log (1 + \TotalRound \TotalAction / \RegularizationParam ) + 2 
            + C_3 \TotalRound^{\frac{5}{3}} K^{\frac{5}{3}} \NetworkDepth^{4} \RegularizationParam^{-\frac{1}{6}} \NetworkWidth^{-\frac{1}{6}} \sqrt{\log \NetworkWidth}
            \right\} \, , 
        \\
        & \Efrak := \Efrak_1 \cap \Efrak_2 \, .
    \end{split}    
\end{equation*}
Then, we decompose the cumulative expected regret into two components: when $\Efrak$ occurs and when $\Efrak$ does not happen.
\begin{align*}
    \Regret(T) & = \EE \left[ \sum_{t=1}^T \left( R(S^*_t, \vb_t^*) - R(S_t, \vb^*_t) \right) \ind(\Efrak) \right]
                + \EE \left[ \sum_{t=1}^T \left( R(S^*_t, \vb_t^*) - R(S_t, \vb^*_t) \right) \ind(\Efrak^\mathsf{c}) \right]
                \\
                & \le \EE \left[ \sum_{t=1}^T \underbrace{ \left( R(S^*_t, \vb_t^*) - R(S_t, \vb^*_t) \right) \ind(\Efrak)}_{\Ical_t}\right] + \Ocal(1) \, ,
\end{align*}
where the inequality holds since we have $\Efrak$ holds with probability at least $1 - T^{-1}$ by Lemma~\ref{lemma_u'_modified UCB} and Lemma~\ref{lemma_bound of sum of gradient norm}.
To bound $\Ical_t$, we have
    \begin{align}
        \Ical_t 
        & \le R(S_t^*, \ScoreVector_t^*) - R(S_t, \ScoreVector_t^*)
        \nonumber
        \\
        & \le R(S_t^*, \mathbf{u}_t + \mathbf{\UCBErrorTerm}_t) - R(S_t, \ScoreVector_t^*) \label{eq:instantaneous regret}
        \\
        & \le R(S_t, \mathbf{u}_t + \mathbf{\UCBErrorTerm}_t) - R(S_t, \ScoreVector_t^*) \nonumber
        \\
        & \le \LipConst \sqrt{ \sum_{i \in S_t} \left( u_{t,i} + \UCBErrorTerm_t - \Score_{t,i}^* \right)^2}
        \nonumber
        \\
        & \le \LipConst \sqrt{ \sum_{i \in S_t} \left( 2 \gamma_{t-1} \left\| \Gradient (\Context_{t,i};\NNParam_{t-1}) / \sqrt{\NetworkWidth} \right\|_{\UCBGramMatrix_{t-1}^{-1}} 
        + 2 \UCBErrorTerm_t \right)^2} \nonumber
        \\
        & \le 4 \LipConst \sqrt{ \sum_{i \in S_t} \left( \max \left\{ \gamma_{t-1} \left\| \Gradient (\Context_{t,i};\NNParam_{t-1}) / \sqrt{\NetworkWidth} \right\|_{\UCBGramMatrix_{t-1}^{-1}} 
        , \UCBErrorTerm_t  \right\} \right)^2} \, , \label{eq1_ucb thm proof}
    \end{align}
where $\LipConst$ is a Lipschitz constant, the first inequality holds due to the monotonicity of the reward function, the second inequality comes from the choice of the oracle, i.e., $S_t = \Oracle_{\mathcal{S}}(\mathbf{u}_t + \mathbf{\UCBErrorTerm}_t)$, the third inequality follows from the Lipschitz continuity of the reward function, the fourth inequality comes from Lemma~\ref{lemma_u'_modified UCB} and the last inequality holds due to the property, $a+b \le 2 \max \{ a, b\}$.

On the other hand, if we denote $\mathcal{A}_i := \gamma_{t-1} \left\| \Gradient (\Context_{t,i};\NNParam_{t-1}) / \sqrt{\NetworkWidth} \right\|_{\UCBGramMatrix_{t-1}^{-1}}$, then we have
    \begin{align}
        \sqrt{ \sum_{i \in S_t} \left( \max \left\{ \gamma_{t-1} \left\| \Gradient (\Context_{t,i};\NNParam_{t-1}) / \sqrt{\NetworkWidth} \right\|_{\UCBGramMatrix_{t-1}^{-1}} 
        , \UCBErrorTerm_t  \right\} \right)^2}
        & =\sqrt{ \sum_{\mathcal{A}_i \ge \UCBErrorTerm_t} \mathcal{A}_i^2 + \sum_{\mathcal{A}_i < \UCBErrorTerm_t} \UCBErrorTerm_t^2} \nonumber
        \\
        & \le \sqrt{ \sum_{i \in S_t} \mathcal{A}_i^2 + \sum_{i \in S_t} \UCBErrorTerm_t^2} \nonumber
        \\
        & \le \sqrt{ \sum_{i \in S_t} \mathcal{A}_i^2} + \sqrt{ \sum_{i \in S_t} \UCBErrorTerm_t^2} \nonumber
        \\
        & = \sqrt{ \sum_{i \in S_t} \mathcal{A}_i^2} + \sqrt{ \SelectAction} \UCBErrorTerm_t
        \label{eq2_ucb thm proof} \, .
    \end{align}

By substituting~Eq.\eqref{eq2_ucb thm proof} into Eq.\eqref{eq1_ucb thm proof}, we have
    \begin{align}
        R(S_t^*, \ScoreVector_t^*) - R(S_t, \ScoreVector_t^*) 
        \le 4 \LipConst \left( \sqrt{ \sum_{i \in S_t} \gamma_{t-1}^2 \left\| \Gradient (\Context_{t,i};\NNParam_{t-1}) / \sqrt{\NetworkWidth} \right\|_{\UCBGramMatrix_{t-1}^{-1}}^2} + \sqrt{ \SelectAction} \UCBErrorTerm_t \right) \label{eq3_ucb thm proof}
    \end{align}

Therefore, by summing Eq.\eqref{eq3_ucb thm proof} over all $t \in [\TotalRound]$, we have
    \begin{align}
        \Regret(\TotalRound) 
        & \le 4 \LipConst \sum_{t=1}^{\TotalRound} \left(\sqrt{ \sum_{i \in S_t} \gamma_{t-1}^2 \left\| \Gradient (\Context_{t,i};\NNParam_{t-1}) / \sqrt{\NetworkWidth} \right\|_{\UCBGramMatrix_{t-1}^{-1}}^2}  
        + \sqrt{ \SelectAction} \UCBErrorTerm_t \right) \nonumber
        \\
        & \le 4 \LipConst \gamma_{\TotalRound} \sum_{t=1}^{\TotalRound} \sqrt{ \sum_{i \in S_t}  \left\| \Gradient (\Context_{t,i};\NNParam_{t-1}) / \sqrt{\NetworkWidth} \right\|_{\UCBGramMatrix_{t-1}^{-1}}^2}  
            + 4 \LipConst \sqrt{ \SelectAction} \TotalRound \UCBErrorTerm_\TotalRound \nonumber
        \\
        & \le 4 \LipConst \gamma_{\TotalRound} \sqrt{ \TotalRound \sum_{t=1}^{\TotalRound} \sum_{i \in S_t}  \left\| \Gradient (\Context_{t,i};\NNParam_{t-1}) / \sqrt{\NetworkWidth} \right\|_{\UCBGramMatrix_{t-1}^{-1}}^2}  
            + 4 \LipConst \sqrt{ \SelectAction} \TotalRound \UCBErrorTerm_\TotalRound \nonumber
        \\
        & \le 4 \LipConst \gamma_{\TotalRound} 
        \sqrt{\TotalRound\left( 2 \EffectiveDim \log (1+\TotalRound \TotalAction / \RegularizationParam)+ 2 
        + \bar{C}_1  \TotalRound^{\frac{5}{3}} K^{\frac{5}{3}} \NetworkDepth^4  \RegularizationParam^{-\frac{1}{6}} \NetworkWidth^{-\frac{1}{6}} \sqrt{ \log \NetworkWidth} \right)}
            + 4 \LipConst \sqrt{ \SelectAction} \TotalRound \UCBErrorTerm_\TotalRound \, , \label{eq4_ucb thm proof}
    \end{align}
where the second inequality holds since $\gamma_t \le \gamma_\TotalRound$ and $ \UCBErrorTerm_t \le \UCBErrorTerm_\TotalRound$, the third inequality follows from the Cauchy-Schwarz inequality and the last inequality comes from Lemma~\ref{lemma_bound of sum of gradient norm} with an absolute constant $\bar{C}_1 >0$.

Meanwhile, we bound $\gamma_\TotalRound$ as follows:
    \begin{align}
        \gamma_\TotalRound 
        & = \Gamma_{1,T} \left( \SubGaussian \sqrt{ \log \frac{ \det \UCBGramMatrix_\TotalRound}{\det \RegularizationParam \mathbf{I}} 
            + C_{\Gamma, 2} \TotalRound^{\frac{5}{3}} K^{\frac{5}{3}} \NetworkDepth^4 \RegularizationParam^{-\frac{1}{6}} \NetworkWidth^{-\frac{1}{6}} \sqrt{\log \NetworkWidth} - 2 \log \delta} + \sqrt{\RegularizationParam} \NormParam \right) \nonumber
        \\
        & \phantom{{}={}} 
            + (\RegularizationParam + \bar{C}_2 \TotalRound \SelectAction \NetworkDepth) \left((1-\StepSize \NetworkWidth \RegularizationParam)^{\frac{\GDSteps}{2}} \sqrt{\TotalRound\SelectAction/\RegularizationParam}
            + \Gamma_{3,T} \right) \nonumber
        \\
        & \le \Gamma_{1,T} \left( \SubGaussian \sqrt{ \log \frac{ \det \bar{\Zb}_T}{\det \RegularizationParam \mathbf{I}} 
            + 2 C_{\Gamma, 2} \TotalRound^{\frac{5}{3}} K^{\frac{5}{3}} \NetworkDepth^4 \RegularizationParam^{-\frac{1}{6}} \NetworkWidth^{-\frac{1}{6}} \sqrt{\log \NetworkWidth} - 2 \log \delta} + \sqrt{\RegularizationParam} \NormParam \right) \nonumber
        \\
        & \phantom{{}={}} 
            + (\RegularizationParam + \bar{C}_2 \TotalRound \SelectAction \NetworkDepth) \left((1-\StepSize \NetworkWidth \RegularizationParam)^{\frac{\GDSteps}{2}} \sqrt{\TotalRound\SelectAction/\RegularizationParam}
            + \Gamma_{3,T} \right) \nonumber
        \\
        & \le \Gamma_{1,T} \left( \SubGaussian \sqrt{ \tilde{d} \log(1 + T N / \lambda) + 1 
            + 2 C_{\Gamma, 2} \TotalRound^{\frac{5}{3}} K^{\frac{5}{3}} \NetworkDepth^4 \RegularizationParam^{-\frac{1}{6}} \NetworkWidth^{-\frac{1}{6}} \sqrt{\log \NetworkWidth} - 2 \log \delta} + \sqrt{\RegularizationParam} \NormParam \right) \nonumber
        \\
        & \phantom{{}={}} 
            + (\RegularizationParam + \bar{C}_2 \TotalRound \SelectAction \NetworkDepth) \left((1-\StepSize \NetworkWidth \RegularizationParam)^{\frac{\GDSteps}{2}} \sqrt{\TotalRound\SelectAction/\RegularizationParam}
            + \Gamma_{3,T} \right) \label{eq5_ucb thm proof}
    \end{align}
where the fist inequality holds due to Lemma~\ref{extended lemma:lemma B.3 in NeuralUCB}, the second inequality holds due to Eq.\eqref{eq_logdet Z_bar_T / Z0}. 

Note that by setting 
$\StepSize = C_1 (\TotalRound \SelectAction \NetworkWidth \NetworkDepth + \NetworkWidth \RegularizationParam)^{-1}$
and
$\GDSteps = 2 \log \left( \frac{\sqrt{\RegularizationParam/\TotalRound \SelectAction}}{ \RegularizationParam + \bar{C}_2 \TotalRound \SelectAction \NetworkDepth }\right) \frac{\TotalRound \SelectAction \NetworkDepth}{C_1 \RegularizationParam}$,
we have
    \begin{equation*}
        (\RegularizationParam + \bar{C}_2 \TotalRound \SelectAction \NetworkDepth) (1-\StepSize \NetworkWidth \RegularizationParam)^{\frac{\GDSteps}{2}} \sqrt{\TotalRound\SelectAction/\RegularizationParam} \le 1 \, .
    \end{equation*}

By choosing sufficiently large $m$ such that
    \begin{align*}
        & \Gamma_{1,T} = \sqrt{1 + C_{\Gamma,1} T^{\frac{7}{6}} K^{\frac{7}{6}} \NetworkDepth^4 \RegularizationParam^{-\frac{7}{6}} \NetworkWidth^{-\frac{1}{6}} \sqrt{\log \NetworkWidth}} \le 2
        \\
        & \Gamma_{2,T} = C_{\Gamma, 2} \TotalRound^{\frac{5}{3}} K^{\frac{5}{3}} \NetworkDepth^4 \RegularizationParam^{-\frac{1}{6}} \NetworkWidth^{-\frac{1}{6}} \sqrt{\log \NetworkWidth} \le 1 \, ,
        \\
        & C_1  \TotalRound^{\frac{5}{3}} \SelectAction^{\frac{3}{2}} \NetworkDepth^4  \RegularizationParam^{-\frac{1}{6}} \NetworkWidth^{-\frac{1}{6}} \sqrt{ \log \NetworkWidth} \le 1 \, ,
        \\
        & (\lambda + \bar{C}_2 T K L ) \Gamma_{3,T}
            = (\lambda + \bar{C}_2 T K L ) C_{\Gamma,3} \TotalRound^{\frac{7}{6}} \SelectAction^{\frac{7}{6}} \NetworkDepth^{\frac{7}{2}} \RegularizationParam^{-\frac{7}{6}} \NetworkWidth^{-\frac{1}{6}} \sqrt{\log \NetworkWidth} (1+\sqrt{ \TotalRound \SelectAction / \RegularizationParam}) \le 1 \, ,
        \\
        & \TotalRound \UCBErrorTerm_\TotalRound \le \gamma_\TotalRound + 1
        \le 2 \SubGaussian \sqrt{\EffectiveDim \log (1 + \TotalRound \TotalAction / \RegularizationParam) + 3 - 2 \log \delta} + 2 \sqrt{\RegularizationParam} \NormParam + 3 \, ,
    \end{align*}
and combining all the results, $\Regret(\TotalRound)$ can be bounded by
    \begin{align}
        \Regret(\TotalRound)
        & \le 4 \LipConst \sqrt{\TotalRound \left( 2 \EffectiveDim \log (1+\TotalRound \TotalAction / \RegularizationParam)+ 3 \right)}
        \bigg[ 2 \SubGaussian \sqrt{\EffectiveDim \log (1 + \TotalRound \TotalAction / \RegularizationParam) + 3 - 2 \log \delta} + 2 \sqrt{\RegularizationParam} \NormParam + 2 \bigg] \nonumber
        \\
        & + 4 \LipConst \sqrt{\SelectAction} \bigg( 2 \SubGaussian \sqrt{\EffectiveDim \log (1 + \TotalRound \TotalAction / \RegularizationParam) + 3 - 2 \log \delta} + 2 \sqrt{\RegularizationParam} \NormParam + 3 \bigg) \nonumber \, .
    \end{align}
\end{proof}

\section{Regret Bound for $\CNTS$} \label{appx:sec_regret for CN-TS}

\subsection{Proof Lemma~\ref{lemma_Optimistic sampling}}
\begin{proof}[proof of Lemma~\ref{lemma_Optimistic sampling}]
For given $\mathcal{F}_t$, since $\tilde{\Score}_{t,i}^{(j)} \sim \mathcal{N}(f(\Context_{t,i};\NNParam_{t-1}), \ExplorationVariance^2 \sigma_{t,i}^2)$, we have
    \begin{align*}
        & \mathbb{P} \left( \max_j \tilde{\Score}_{t,i}^{(j)} + \TSErrorTerm > h(\Context_{t,i}) \mid
        \mathcal{F}_t, \EventEstimating_t \right)
        = 1 - \mathbb{P} \left( \tilde{\Score}_{t,i}^{(j)} + \TSErrorTerm \le h(\Context_{t,i}), \forall j \in [\TotalSampleNum] \mid \mathcal{F}_t, \EventEstimating_t \right)
        \\
        & = 1 - \mathbb{P} \left( \frac{\tilde{\Score}_{t,i}^{(j)} - f(\Context_{t,i};\NNParam_{t-1}) + \TSErrorTerm}{\ExplorationVariance \sigma_{t,i}} 
        \le \frac{h(\Context_{t,i}) - f(\Context_{t,i};\NNParam_{t-1})}{\ExplorationVariance \sigma_{t,i}}, \forall j \in [\TotalSampleNum] \mid \mathcal{F}_t, \EventEstimating_t \right)
        \\
        & \ge 1 - \mathbb{P} \left( \frac{\tilde{\Score}_{t,i}^{(j)} - f(\Context_{t,i};\NNParam_{t-1}) + \TSErrorTerm}{\ExplorationVariance \sigma_{t,i}} 
        \le \frac{ \left| h(\Context_{t,i}) - f(\Context_{t,i};\NNParam_{t-1}) \right| }{\ExplorationVariance \sigma_{t,i}}, \forall j \in [\TotalSampleNum] \mid \mathcal{F}_t, \EventEstimating_t \right)
        \\
        & = 1 - \mathbb{P} \left( \frac{\tilde{\Score}_{t,i}^{(j)} - f(\Context_{t,i};\NNParam_{t-1}) }{\ExplorationVariance \sigma_{t,i}} 
        \le \frac{ \left| h(\Context_{t,i}) - f(\Context_{t,i};\NNParam_{t-1}) \right| - \TSErrorTerm}{\ExplorationVariance \sigma_{t,i}}, \forall j \in [\TotalSampleNum] \mid \mathcal{F}_t, \EventEstimating_t \right)
        \\
        & = 1 - \mathbb{P} \left( Z_j \le \frac{ \left| h(\Context_{t,i}) - f(\Context_{t,i};\NNParam_{t-1}) \right| - \TSErrorTerm}{\ExplorationVariance \sigma_{t,i}}, \forall j \in [\TotalSampleNum] \mid \mathcal{F}_t, \EventEstimating_t \right) \, ,
    \end{align*}
where the first inequality is due to $a \le |a|$, for the last equality we denote $Z_j$ as a standard normal random variable. Note that under the event $\EventEstimating_t$, we have $ \left| f(\Context_{t,i};\NNParam_{t-1}) - h(\Context_{t,i}) \right| \le \ExplorationVariance \sigma_{t,i} + \TSErrorTerm$ for all $i \in [\TotalAction]$. Hence, under the event $\EventEstimating_t$,
    \begin{equation*}
        \frac{ \left| h(\Context_{t,i}) - f(\Context_{t,i};\NNParam_{t-1}) \right| - \TSErrorTerm}{\ExplorationVariance \sigma_{t,i}}
        \le \frac{ \ExplorationVariance \sigma_{t,i} + \TSErrorTerm - \TSErrorTerm}{\ExplorationVariance \sigma_{t,i}} = 1 \, .
    \end{equation*}
    
Then, it follows that
    \begin{align*}
        \mathbb{P} \left( \max_j \tilde{\Score}_{t,i}^{(j)} + \TSErrorTerm > h(\Context_{t,i}) \mid
        \mathcal{F}_t, \EventEstimating_t \right)
        & \ge 1 - \left[ \mathbb{P} \left( Z \le 1 \right) \right]^\TotalSampleNum \, .
    \end{align*}
    
Using the anti-concentration inequality in Lemma~\ref{lemma_anti-concentraion}, we have $\mathbb{P}(Z \le 1) \le 1 - \OptimisticProbability$ where $\OptimisticProbability := 1/(4 e \sqrt{\pi})$.
Then finally we have
    \begin{align*}
        \mathbb{P} \bigg( R(S_t, \tilde{\ScoreVector}_t + \boldsymbol{\TSErrorTerm}) \ge R(S_t^*, \ScoreVector_t^*) \mid \mathcal{F}_t,  \EventEstimating_t \bigg)
        & \ge \mathbb{P} \bigg( R(S_t^*, \tilde{\ScoreVector}_t + \boldsymbol{\TSErrorTerm}) \ge R(S_t^*, \ScoreVector_t^*) \mid \mathcal{F}_t,  \EventEstimating_t \bigg)
        \\
        & \ge \mathbb{P} \bigg( \tilde{\Score}_{t,i} + \TSErrorTerm \ge h(\Context_{t,i}), \forall i \in S_t^* \mid \mathcal{F}_t,  \EventEstimating_t \bigg)
        \\ 
        & = \prod_{i \in S_t} \mathbb{P} \bigg( \tilde{\Score}_{t,i} + \TSErrorTerm \ge h(\Context_{t,i}) \mid \mathcal{F}_t,  \EventEstimating_t \bigg)
        \\
        & \ge \left( 1 - \left[ \mathbb{P} ( Z \le 1) \right]^\TotalSampleNum \right)^\SelectAction
        \\
        & \ge \left[ 1 - (1 - \OptimisticProbability)^\TotalSampleNum \right]^\SelectAction
        \\
        & \ge 1 - \SelectAction \left( 1 - \OptimisticProbability \right)^\TotalSampleNum
        \\
        & \ge 1- \left(1 - \OptimisticProbability \right)
        \\
        & = \OptimisticProbability \, ,
    \end{align*}
where the first inequality holds due to the choice of the oracle, the second inequality comes from the monotonicity of the reward function, the third inequality uses the Bernoulli's inequality, and the last inequality comes from the choice of $\TotalSampleNum = \lceil 1 - \frac{\log \SelectAction}{\log(1-\OptimisticProbability)} \rceil$, which means $(1-\OptimisticProbability)^\TotalSampleNum \le \frac{1}{\SelectAction} (1-\OptimisticProbability)$.
\end{proof}

\subsection{Proof of Theorem~\ref{thm_CN-TS}} \label{subsec:proof of CN-TS}
\begin{proof}[Proof of Theorem~\ref{thm_CN-TS}] 
First of all, we decompose the expected cumulative regret as follows:
    \begin{equation*}
        \Regret(\TotalRound) = \underbrace{ \sum_{t=1}^{\TotalRound} \mathbb{E} \left[ R(S_t^*, \ScoreVector_t^*) - R(S_t, \tilde{\ScoreVector}_t + \boldsymbol{\TSErrorTerm}) \right]}_{\Regret_1(\TotalRound)}
        + \underbrace{ \sum_{t=1}^{\TotalRound} \mathbb{E} \left[ R(S_t, \tilde{\ScoreVector}_t + \boldsymbol{\TSErrorTerm}) - R(S_t, \ScoreVector_t^*) \right]}_{\Regret_2(\TotalRound)} \, .
    \end{equation*}
    
From now on, we derive the bounds for $\Regret_1(\TotalRound)$ and $\Regret_2(\TotalRound)$ respectively.
\\
\\
\textbf{Bounding $\Regret_2(\TotalRound)$}
\\
\\
First we decompose $\Regret_2(\TotalRound)$:
    \begin{equation*}
        \Regret_2(\TotalRound) = \sum_{t=1}^{\TotalRound} \mathbb{E} \left[ \underbrace{R(S_t, \tilde{\ScoreVector}_t + \boldsymbol{\TSErrorTerm}) - R(S_t, \hat{\ScoreVector}_t + \boldsymbol{\TSErrorTerm})}_{I_2} \right]
        + \sum_{t=1}^{\TotalRound} \mathbb{E} \left[ \underbrace{R(S_t, \hat{\ScoreVector}_t + \boldsymbol{\TSErrorTerm} ) - R(S_t, \ScoreVector_t^*)}_{I_1} \right] \, .
    \end{equation*}
For $I_1$, we have
    \begin{align*}
        \left| R(S_t, \hat{\ScoreVector}_t + \boldsymbol{\TSErrorTerm} ) - R(S_t, \ScoreVector_t^*) \right|
        & \le C_0^{(1)} \sqrt{ \sum_{i \in S_t} \left( f(\Context_{t,i};\NNParam_{t-1}) + \TSErrorTerm - h(\Context_{t,i}) \right)^2}
        \\
        & \le C_0^{(1)} \sqrt{ \sum_{i \in S_t} \left( \ExplorationVariance \sigma_{t,i} + 2 \TSErrorTerm \right)^2}
        \\
        & \le C_0^{(1)} \sqrt{  \sum_{i \in S_t} \left( 2 \max \{ \ExplorationVariance \sigma_{t,i}, 2 \TSErrorTerm \} \right)^2}
        \\
        & \le 2C_0^{(1)} \sqrt{ \sum_{i \in S_t} ( \ExplorationVariance \sigma_{t,i})^2  + \sum_{i \in S_t} 4 \TSErrorTerm^2}
        \\
        & \le 2C_0^{(1)} \left[ \sqrt{\sum_{i \in S_t} (\ExplorationVariance \sigma_{t,i})^2} + \sqrt{ \sum_{i \in S_t} 4 \TSErrorTerm^2} \right]
        \\
        & = 2C_0^{(1)} \left[ \ExplorationVariance \sqrt{\sum_{i \in S_t} \sigma_{t,i}^2} + 2 \TSErrorTerm \sqrt{ \SelectAction}  \right] \, ,
    \end{align*}
where the first inequality holds due to the Lipschitz continuity for a constant $C_0^{(1)} >0$, the second inequality holds due to the event $\EventEstimating_t$ holds with high probability, the third inequality follows from the property that $a+b \le 2 \max \{a, b\}$, and the last inequality uses the fact that $\sqrt{a+b} \le \sqrt{a}+\sqrt{b}$ for any $a, b \ge 0$.

On the other hand, for $I_2$ we have
    \begin{align*}
        \left| R(S_t, \tilde{\ScoreVector}_t + \boldsymbol{\TSErrorTerm}) - R(S_t, \hat{\ScoreVector}_t + \boldsymbol{\TSErrorTerm}) \right|
        & \le C_0^{(2)} \sqrt{ \sum_{i \in S_t} \left( \tilde{\Score}_{t,i} - f(\Context_{t,i};\NNParam_{t-1}) \right)^2 }
        \\
        & \le C_0^{(2)}
        \sqrt{ \sum_{i \in S_t} \SigmaEventConstant_t^2 \ExplorationVariance^2 \sigma_{t,i}^2 }
        \\
        & = C_0^{(2)} \SigmaEventConstant_t \ExplorationVariance \sqrt{ \sum_{i \in S_t} \sigma_{t,i}^2} \, ,
    \end{align*}
where the first inequality holds for some Lipschitz continuity constant $C_0^{(2)} > 0$, the second inequality holds due to the event $\EventSampling_t$ holds with high probability.

By combining the bounds of $I_1$ and $I_2$, we derive the bound for $\Regret_2(\TotalRound)$ as follows:
    \begin{align}
        \Regret_2(\TotalRound) 
        & \le 2C_0 \sum_{t=1}^{\TotalRound} \mathbb{E} \left[ \ExplorationVariance \sqrt{ \sum_{i \in S_t} \sigma_{t,i}^2 } + 2 \TSErrorTerm \sqrt{\SelectAction} \right]
        + C_0 \ExplorationVariance \SigmaEventConstant_\TotalRound \sum_{t=1}^{\TotalRound} \mathbb{E} \left[ \sqrt{ \sum_{i \in S_t} \sigma_{t,i}^2} \right] \nonumber
        \\
        & = C_0 \ExplorationVariance (\SigmaEventConstant_\TotalRound + 2) \mathbb{E} \left[ \sum_{t=1}^{\TotalRound} \sqrt{ \sum_{i \in S_t} \sigma_{t,i}^2} \right]
        + 2 C_0 \TotalRound \sqrt{\SelectAction} \TSErrorTerm \nonumber
        \\ 
        & \le C_0 \ExplorationVariance  (\SigmaEventConstant_\TotalRound + 2)  \mathbb{E} \left[ \sqrt{ \TotalRound \sum_{t=1}^{\TotalRound} \sum_{i \in S_t} \sigma_{t,i}^2} \right]
        + 2 C_0 \TotalRound \sqrt{\SelectAction} \TSErrorTerm \nonumber
        \\
        & = C_0 \ExplorationVariance  (\SigmaEventConstant_\TotalRound + 2) \mathbb{E} \left[ \sqrt{ \TotalRound \RegularizationParam \sum_{t=1}^{\TotalRound} \sum_{i \in S_t} \left\| \Gradient (\Context_{t,i} ; \NNParam_{t-1}) / \sqrt{ \NetworkWidth} \right\|_{\TSGramMatrix_t^{-1}}^2 } \right]
        +  2 C_0 \TotalRound \sqrt{\SelectAction} \TSErrorTerm \nonumber 
        \\
        & \le C_0 \ExplorationVariance (\SigmaEventConstant_\TotalRound + 2) \sqrt{ \TotalRound \RegularizationParam \left( 2 \EffectiveDim \log(1+ \TotalRound \TotalAction / \RegularizationParam)
        + 2 + C_1  \TotalRound^{\frac{5}{3}} \SelectAction^{\frac{3}{2}} \NetworkDepth^4  \RegularizationParam\!^{-\frac{1}{6}} \NetworkWidth\!^{-\frac{1}{6}}\!\sqrt{ \log \NetworkWidth} \right)} \nonumber
        \\
        & \phantom{{}={}}
        +  2 C_0 \TotalRound \sqrt{\SelectAction} \TSErrorTerm \label{eq_bound of R2} \, ,
    \end{align}
where $C_1>0$ is a constant, the first inequality uses $\SigmaEventConstant_t \le \SigmaEventConstant_\TotalRound$ and $C_0 = \max \{ C_0^{(1)}, C_0^{(2)} \}$, the second inequality follows from the Cauchy-Schwarz inequality, and the last inequality holds due to Lemma~\ref{lemma_bound of sum of gradient norm}.
\\
\\
\textbf{Bounding $\Regret_1(\TotalRound)$}
\\
\\
Note that a sufficient condition for ensuring the success of $\CNTS$ is to show that the probability of sampling being optimistic is high enough. Lemma~\ref{lemma_Optimistic sampling} gives a lower bound of the probability that the reward induced by sampled scores is larger than the reward induced by the expected scores up to the approximation error. For our analysis, first we define $\SetOfConcentratedSamples$ the set of concentrated samples for which the reward induced by sampled scores concentrate appropriately to the reward induced by the estimated scores. Also, we define the set of optimistic samples $\SetOfOptimisticSamples$ which coinciding with $\SetOfConcentratedSamples$.
    \begin{align*}
        & \SetOfConcentratedSamples := \left\{ \{ \tilde{\Score}_{t,i}^{(j)} \mid i \in [\TotalAction] \}_{j=1}^{\TotalSampleNum} =: \dot{\ScoreVector}_t^{1:\TotalSampleNum} \mid R(S_t, \tilde{\ScoreVector}_t + \boldsymbol{\TSErrorTerm} ) - R(S_t, \hat{\ScoreVector}_t + \boldsymbol{\TSErrorTerm}) \le C_0 \sqrt{\sum_{i \in S_t} (\SigmaEventConstant_t \ExplorationVariance \sigma_{t,i})^2} \right\} \, ,
        \\
        & \SetOfOptimisticSamples := \left\{ \{ \tilde{\Score}_{t,i}^{(j)} \mid i \in [\TotalAction] \}_{j=1}^{\TotalSampleNum} =: \ddot{\ScoreVector}_t^{1:\TotalSampleNum} \mid R(S_t, \tilde{\ScoreVector}_t + \boldsymbol{\TSErrorTerm}) >  R(S_t^*, \ScoreVector_t^*) \right\} \cap \SetOfConcentratedSamples \, .
    \end{align*}
Also, note that the event $\mathcal{E}_t := \EventSampling_t \cap \EventEstimating_t$, which means 
    \begin{equation*}
        \mathcal{E}_t = \{ \left| f(\Context_{t,i};\NNParam_{t-1}) - h(\Context_{t,i}) \right| \le \ExplorationVariance \sigma_{t,i} + \TSErrorTerm , \forall i \in [\TotalAction] \} 
        \cap \{ \left| \tilde{\Score}_{t,i} - f(\Context_{t,i};\NNParam_{t-1}) \right| \le \SigmaEventConstant_t \ExplorationVariance \sigma_{t,i} , \forall i \in [\TotalAction] \} \, .
    \end{equation*}
For our notations, we denote $\dot{S}_t$ as the super arm induced by the sampled score $\dot{\ScoreVector}_t^{1:\TotalSampleNum} \in \SetOfConcentratedSamples$ and  $\boldsymbol{\TSErrorTerm}$. Also we represent $R(S, \dot{\ScoreVector}_t^{1:\TotalSampleNum} + \boldsymbol{\TSErrorTerm})$ the reward under the sampled score $\dot{\ScoreVector}_t^{1:\TotalSampleNum}$ and $\boldsymbol{\TSErrorTerm}$. Also, we define $\ddot{S}_t$ as the super arm induced by $\ddot{\ScoreVector}_t \in \SetOfOptimisticSamples$ and $\boldsymbol{\TSErrorTerm}$. Similarly we can define $R(S, \ddot{\ScoreVector}_t + \boldsymbol{\TSErrorTerm})$.

Recall that $S_t = \argmax R(S, \tilde{\ScoreVector}_t + \boldsymbol{\TSErrorTerm})$.
Then, for any $\dot{\ScoreVector}_t^{1:\TotalSampleNum} \in \SetOfConcentratedSamples$, we have
    \begin{equation*}
        \bigg( R(S_t^*, \ScoreVector_t^*) - R(S_t, \tilde{\ScoreVector}_t + \boldsymbol{\TSErrorTerm}) \bigg) \ind \left( \mathcal{E}_t \right)
        \le \left( R(S_t^*, \ScoreVector_t^*) - \inf_{\dot{\ScoreVector}_t^{1:\TotalSampleNum} \in \SetOfConcentratedSamples} \max_{S} R(S, \dot{\ScoreVector}_t^{1:\TotalSampleNum} + \boldsymbol{\TSErrorTerm}) \right) \ind  \left( \mathcal{E}_t \right) \, .
    \end{equation*}
Note that we can decompose 
    \begin{equation*}
        R(S_t^*, \ScoreVector_t^*) - R(S_t, \tilde{\ScoreVector}_t + \boldsymbol{\TSErrorTerm}) = \bigg( R(S_t^*, \ScoreVector_t^*) - R(S_t, \tilde{\ScoreVector}_t + \boldsymbol{\TSErrorTerm}) \bigg) \ind (\mathcal{E}_t) 
        + \bigg( R(S_t^*, \ScoreVector_t^*) - R(S_t, \tilde{\ScoreVector}_t + \boldsymbol{\TSErrorTerm}) \bigg) \ind (\mathcal{E}_t^{\mathsf{c}}) \, .
    \end{equation*}
    
Since the event $\mathcal{E}_t$ holds with high probability, we can bound the summation of the second term in the right hand side as follows:
    \begin{equation*}
        \sum_{t=1}^{\TotalRound} \mathbb{E} \left[ R(S_t^*, \ScoreVector_t^*) - R(S_t, \tilde{\ScoreVector}_t + \boldsymbol{\TSErrorTerm}) \right] 
        = \sum_{t=1}^{\TotalRound} \underbrace{\mathbb{E} \left[ \bigg( R(S_t^*, \ScoreVector_t^*) - R(S_t, \tilde{\ScoreVector}_t + \boldsymbol{\TSErrorTerm}) \bigg) \ind (\mathcal{E}_t) \right]}_{I_3} + \mathcal{O}(1) \, .
    \end{equation*}
Therefore, we need to bound the summation of $I_3$. Note that we have
    \begin{align*}
        & \mathbb{E} \left[ \bigg( R(S_t^*, \ScoreVector_t^*) - R(S_t, \tilde{\ScoreVector}_t + \boldsymbol{\TSErrorTerm}) \bigg) \ind (\mathcal{E}_t) \mid \mathcal{F}_t \right]
        \\
        & \le \mathbb{E} \left[ \left( R(S_t^*, \ScoreVector_t^*) - \inf_{\dot{\ScoreVector}_t^{1:\TotalSampleNum} \in \SetOfConcentratedSamples} \max_{S} R(S, \dot{\ScoreVector}_t^{1:\TotalSampleNum} + \boldsymbol{\TSErrorTerm}) \right) \ind  \left( \mathcal{E}_t \right) \mid \mathcal{F}_t \right]
        \\
        & = \mathbb{E} \left[ \left( R(S_t^*, \ScoreVector_t^*) - \inf_{\dot{\ScoreVector}_t^{1:\TotalSampleNum} \in \SetOfConcentratedSamples} \max_{S} R(S, \dot{\ScoreVector}_t^{1:\TotalSampleNum} + \boldsymbol{\TSErrorTerm}) \right) \ind  \left( \mathcal{E}_t \right) \mid \mathcal{F}_t, \ddot{\ScoreVector}_t^{1:\TotalSampleNum} \in \SetOfOptimisticSamples \right]
        \\
        & \le \mathbb{E} \left[ \left( R(\ddot{S}_t, \tilde{\ScoreVector}_t + \boldsymbol{\TSErrorTerm}) - \inf_{\dot{\ScoreVector}_t^{1:\TotalSampleNum} \in \SetOfConcentratedSamples} \max_{S} R(S, \dot{\ScoreVector}_t^{1:\TotalSampleNum} + \boldsymbol{\TSErrorTerm}) \right) \ind  \left( \mathcal{E}_t \right) \mid \mathcal{F}_t, \ddot{\ScoreVector}_t^{1:\TotalSampleNum} \in \SetOfOptimisticSamples \right]
        \\
        & \le \mathbb{E} \left[ \left( R(\ddot{S}_t, \tilde{\ScoreVector}_t + \boldsymbol{\TSErrorTerm}) -  \inf_{\dot{\ScoreVector}_t^{1:\TotalSampleNum} \in \SetOfConcentratedSamples} R(\ddot{S}_t, \dot{\ScoreVector}_t^{1:\TotalSampleNum} + \boldsymbol{\TSErrorTerm}) \right) \ind  \left( \mathcal{E}_t \right) \mid \mathcal{F}_t, \ddot{\ScoreVector}_t^{1:\TotalSampleNum} \in \SetOfOptimisticSamples \right]
        \\
        & = \mathbb{E} \left[ \sup_{\dot{\ScoreVector}_t^{1:\TotalSampleNum} \in \SetOfConcentratedSamples} \left( R(\ddot{S}_t, \tilde{\ScoreVector}_t + \boldsymbol{\TSErrorTerm}) -  R(\ddot{S}_t, \dot{\ScoreVector}_t^{1:\TotalSampleNum} + \boldsymbol{\TSErrorTerm}) \right) \ind  \left( \mathcal{E}_t \right) \mid \mathcal{F}_t, \ddot{\ScoreVector}_t^{1:\TotalSampleNum} \in \SetOfOptimisticSamples \right]
        \\
        & = \mathbb{E} \left[ \left( R(\ddot{S}_t, \tilde{\ScoreVector}_t + \boldsymbol{\TSErrorTerm}) -  R(\ddot{S}_t, \ddot{\ScoreVector}_t^{1:\TotalSampleNum} + \boldsymbol{\TSErrorTerm}) \right) \ind  \left( \mathcal{E}_t \right) \mid \mathcal{F}_t, \ddot{\ScoreVector}_t^{1:\TotalSampleNum} \in \SetOfOptimisticSamples \right]
        \\
        & = \mathbb{E} \left[ \left( R(\ddot{S}_t, \tilde{\ScoreVector}_t + \boldsymbol{\TSErrorTerm}) - R(\ddot{S}_t, \hat{\ScoreVector}_t + \boldsymbol{\TSErrorTerm}) + R(\ddot{S}_t, \hat{\ScoreVector}_t + \boldsymbol{\TSErrorTerm}) - R(\ddot{S}_t, \ddot{\ScoreVector}_t^{1:\TotalSampleNum} + \boldsymbol{\TSErrorTerm}) \right) \ind  \left( \mathcal{E}_t \right) \mid \mathcal{F}_t, \ddot{\ScoreVector}_t^{1:\TotalSampleNum} \in \SetOfOptimisticSamples \right]
        \\
        & \le \mathbb{E} \left[ 2 C_0 \sqrt{ \sum_{i \in \ddot{S}_t} ( \SigmaEventConstant_t \ExplorationVariance \sigma_{t,i} )^2} \ind  \left( \mathcal{E}_t \right) \mid \mathcal{F}_t, \ddot{\ScoreVector}_t^{1:\TotalSampleNum} \in \SetOfOptimisticSamples \right]
        \\
        & = 2 C_0 \SigmaEventConstant_t \ExplorationVariance \mathbb{E} \left[ \sqrt{ \sum_{i \in \ddot{S}_t} \sigma_{t,i}^2} \ind  \left( \mathcal{E}_t \right) \mid \mathcal{F}_t, \ddot{\ScoreVector}_t^{1:\TotalSampleNum} \in \SetOfOptimisticSamples \right]
        \\
        & \le 2 C_0 \SigmaEventConstant_t \ExplorationVariance \mathbb{E} \left[ \sqrt{ \sum_{i \in \ddot{S}_t} \sigma_{t,i}^2} \mid \mathcal{F}_t, \ddot{\ScoreVector}_t^{1:\TotalSampleNum} \in \SetOfOptimisticSamples \right] \cdot \mathbb{P}(\mathcal{E}_t) \, ,
    \end{align*}
where $C_0 >0$ is a Lipschitz constant. On the other hand, from Lemma~\ref{lemma_Optimistic sampling}, we have
    \begin{equation*}
        \mathbb{P} \bigg( R(S_t, \tilde{\ScoreVector}_t + \boldsymbol{ \TSErrorTerm}) > R(S_t^*, \ScoreVector_t^*) \mid \mathcal{F}_t, \mathcal{E}_t \bigg) \ge 1/(4e\sqrt{\pi}) := \OptimisticProbability \, ,
    \end{equation*}
which means that
    \begin{align*}
        \mathbb{P} (\ddot{\ScoreVector}_t^{1:\TotalSampleNum} \in \SetOfOptimisticSamples \mid \mathcal{F}_t, \mathcal{E}_t)
        & = \mathbb{P} \left( R(\ddot{S}_t, \tilde{\ScoreVector}_t + \boldsymbol{\TSErrorTerm}) > R(S_t^*, \ScoreVector_t^*) \; \text{and} \; \ddot{\ScoreVector}_t^{1:\TotalSampleNum} \in \SetOfConcentratedSamples \mid \mathcal{F}_t, \mathcal{E}_t \right)
        \\
        & \ge \mathbb{P} \left( R(\ddot{S}_t, \tilde{\ScoreVector}_t + \boldsymbol{\TSErrorTerm}) > R(S_t^*, \ScoreVector_t^*) \mid \mathcal{F}_t, \mathcal{E}_t \right)
        - \mathbb{P} \left( \ddot{\ScoreVector}_t^{1:\TotalSampleNum} \notin \SetOfConcentratedSamples \mid \mathcal{F}_t, \mathcal{E}_t \right)
        \\
        & \ge \OptimisticProbability - \mathcal{O}(t^{-1})
        \\
        & \ge \OptimisticProbability/2.
    \end{align*}
Then, we can write 
    \begin{align*}
        \mathbb{E} \left[ \sqrt{ \sum_{i \in S'_t} \sigma_{t,i}^2} \mid \mathcal{F}_t, \mathcal{E}_t \right]
        & \ge \mathbb{E} \left[ \sqrt{ \sum_{i \in \ddot{S}_t} \sigma_{t,i}^2} \mid \mathcal{F}_t, \mathcal{E}_t , \ddot{\ScoreVector}_t^{1:\TotalSampleNum} \in \SetOfOptimisticSamples \right]
        \cdot \mathbb{P} \left( \ddot{\ScoreVector}_t^{1:\TotalSampleNum} \in \SetOfOptimisticSamples \mid \mathcal{F}_t, \mathcal{E}_t \right)
        \\
        & \ge \mathbb{E} \left[ \sqrt{ \sum_{i \in \ddot{S}_t} \sigma_{t,i}^2} \mid \mathcal{F}_t, \mathcal{E}_t , \ddot{\ScoreVector}_t^{1:\TotalSampleNum} \in \SetOfOptimisticSamples \right] \cdot \OptimisticProbability/2 \, ,
    \end{align*}
where $S'_t$ is a super arm induced by any sampled scores. By combining the results, we have
    \begin{align*}
        \mathbb{E} \left[ \bigg( R(S_t^*, \ScoreVector_t^*) - R(S_t, \tilde{\ScoreVector}_t + \boldsymbol{\TSErrorTerm}) \bigg) \ind (\mathcal{E}_t) \mid \mathcal{F}_t \right]
        & \le 2 C_0 \SigmaEventConstant_t \ExplorationVariance \mathbb{E} \left[ \sqrt{ \sum_{i \in \ddot{S}_t} \sigma_{t,i}^2} \mid \mathcal{F}_t, \mathcal{E}_t, \ddot{\ScoreVector}_t^{1:\TotalSampleNum} \in \SetOfOptimisticSamples \right] \cdot \mathbb{P}(\mathcal{E}_t) 
        \\
        & \le \frac{4 C_0 \SigmaEventConstant_t \ExplorationVariance}{\OptimisticProbability} \mathbb{E} \left[ \sqrt{ \sum_{i \in S'_t} \sigma_{t,i}^2} \mid \mathcal{F}_t, \mathcal{E}_t \right] \cdot \mathbb{P}(\mathcal{E}_t)
        \\
        & \le \frac{4 C_0 \SigmaEventConstant_t \ExplorationVariance}{\OptimisticProbability} \mathbb{E} \left[ \sqrt{ \sum_{i \in S'_t} \sigma_{t,i}^2} \mid \mathcal{F}_t \right] \, .
    \end{align*}
Summing over all $t \in [\TotalRound]$ and the failure event into consideration, we have
    \begin{align}
        \sum_{t=1}^{\TotalRound} \mathbb{E} \left[ \bigg( R(S_t^*, \ScoreVector_t^*) - R(S_t, \tilde{\ScoreVector}_t + \boldsymbol{\TSErrorTerm}) \bigg) \ind (\mathcal{E}_t) \mid \Filtration_{t} \right]
        & \le \frac{4 C_0 \SigmaEventConstant_\TotalRound \ExplorationVariance}{\OptimisticProbability} \sum_{t=1}^{\TotalRound}  \mathbb{E} \left[ \sqrt{ \sum_{i \in S'_t} \sigma_{t,i}^2} \mid \mathcal{F}_t \right] \label{eq_star} \, .
    \end{align}
Note that the summation on the RHS contains an expectation, so we cannot directly apply Lemma~\ref{lemma_bound of sum of gradient norm}. Instead, since we can write
    \begin{align*}
        \sum_{t=1}^{\TotalRound} \mathbb{E} \left[ \sqrt{ \sum_{i \in S'_t} \sigma_{t,i}^2} \mid \mathcal{F}_t \right]  
        = \sum_{t=1}^{\TotalRound} \sqrt{\sum_{i \in S''_t} \sigma_{t,i}^2 } 
        + \sum_{t=1}^{\TotalRound} \left( \mathbb{E} \left[ \sqrt{ \sum_{i \in S'_t} \sigma_{t,i}^2} \mid \mathcal{F}_t \right]
        - \sqrt{\sum_{i \in S''_t} \sigma_{t,i}^2 } \right) \, ,
    \end{align*}
where $S''_t$ is any super arm induced by arbitrary sampled scores. By using Lemma~\ref{lemma_bound of sum of gradient norm} we have
    \begin{align} \label{eq_sum_sigma^2}
        \sum_{t=1}^{\TotalRound} \sqrt{\sum_{i \in S''_t} \sigma_{t,i}^2 } 
        & \le \sqrt{\TotalRound \sum_{t=1}^{\TotalRound} \sum_{i \in S''_t} \sigma_{t,i}^2 } \nonumber
        \\
        & \le \sqrt{ \TotalRound \RegularizationParam \left( 2 \EffectiveDim \log(1+ \TotalRound \TotalAction / \RegularizationParam) + 2 + C_1 \TotalRound^{\frac{5}{3}} \SelectAction^{\frac{3}{2}} \NetworkDepth^4  \RegularizationParam\!^{-\frac{1}{6}} \NetworkWidth\!^{-\frac{1}{6}}\!\sqrt{ \log \NetworkWidth} \right)} \, ,
    \end{align}
where $C_1 >0$ is a constant.

On the other hand, let $Y_t = \sum_{k=1}^{t} \left( \mathbb{E} \left[ \sqrt{ \sum_{i \in S'_k} \sigma_{k,i}^2} \mid \mathcal{F}_k \right]  -  \sqrt{\sum_{i \in S''_k} \sigma_{k,i}^2 } \right)$. Since we have
    \begin{align*}
        & Y_t - Y_{t-1} = \mathbb{E} \left[ \sqrt{ \sum_{i \in S'_t} \sigma_{t,i}^2} \mid \mathcal{F}_t \right]  -  \sqrt{\sum_{i \in S''_t} \sigma_{t,i}^2 } \, ,
    \end{align*}
which implies,
    \begin{align*}
        \mathbb{E} \left[ Y_t - Y_{t-1} \mid \mathcal{F}_t \right] = \mathbb{E} \left[ \mathbb{E} \left[ \sqrt{ \sum_{i \in S'_t} \sigma_{t,i}^2} \mid \mathcal{F}_t \right] \mid \mathcal{F}_t \right] - \mathbb{E} \left[ \sqrt{ \sum_{i \in S''_t} \sigma_{t,i}^2} \mid \mathcal{F}_t \right] = 0 \, ,
    \end{align*}
then $Y_t$ is a martingale for all $1 \le t \le \TotalRound$.

Note that we can bound $\left| Y_t - Y_{t-1}\right|$ as follows:
    \begin{align*}
        \left| Y_t - Y_{t-1}\right| 
        & = \left| \mathbb{E} \left[ \sqrt{\sum_{i \in S'_t} \sigma_{t,i}^2} \mid \mathcal{F}_t \right] - \sqrt{ \sum_{i \in S''_t} \sigma_{t,i}^2} \right|
        \\
        & \le \mathbb{E} \left[ \sqrt{\sum_{i \in S'_t} (C_2 \sqrt{\NetworkDepth})^2 }\mid \mathcal{F}_t \right] 
        + \sqrt{ \sum_{i \in S''_t} (C_2 \sqrt{\NetworkDepth})^2}
        \\
        & = 2 C_2 \sqrt{\NetworkDepth \SelectAction} \, ,
    \end{align*}
where the inequality holds due to Lemma~\ref{aux lemma:lemma B.6 in NeuralUCB} for some positive constant $C_2$. Then, applying the Azuma-Hoeffding inequality (Lemma~\ref{lemma_Azuma}), which means,
    \begin{equation}\label{eq_expected_sum_sigma^2}
        \sum_{t=1}^{\TotalRound} \left( \mathbb{E} \left[ \sqrt{ \sum_{i \in S'_t} \sigma_{t,i}^2} \mid \mathcal{F}_t \right]  -  \sqrt{\sum_{i \in S''_t} \sigma_{t,i}^2 } \right) 
        \le C_2 \sqrt{8 \TotalRound \NetworkDepth \SelectAction \log\TotalRound} \, ,
    \end{equation}
with probability $1-\TotalRound^{-1}$. Combining Eq.\eqref{eq_sum_sigma^2} and Eq.\eqref{eq_expected_sum_sigma^2}, we have
    \begin{align} 
        \mathbb{E} \left[ \sqrt{ \sum_{i \in S'_t} \sigma_{t,i}^2} \mid \mathcal{F}_t \right] 
        & \le \sqrt{ \TotalRound \RegularizationParam \left( 2 \EffectiveDim \log(1+ \TotalRound \TotalAction / \RegularizationParam) + 2 + C_1 \TotalRound^{\frac{5}{3}} \SelectAction^{\frac{3}{2}} \NetworkDepth^4  \RegularizationParam\!^{-\frac{1}{6}} \NetworkWidth\!^{-\frac{1}{6}}\!\sqrt{ \log \NetworkWidth} \right)} \nonumber
        \\
        & \phantom{{}={}}
        + C_2 \sqrt{8 \TotalRound \NetworkDepth \SelectAction \log\TotalRound} \label{eq_star2}
    \end{align}
    
By substituting Eq.\eqref{eq_star2} for Eq.\eqref{eq_star}, we have the bound for $\Regret_1(\TotalRound)$ as follows:
    \begin{align}
        \Regret_1(\TotalRound) 
        & \le \frac{4 C_0 \SigmaEventConstant_t \ExplorationVariance}{\OptimisticProbability} \bigg[ \sqrt{ \TotalRound \RegularizationParam \left( 2 \EffectiveDim \log(1+ \TotalRound \TotalAction / \RegularizationParam) + 2 + C_1 \TotalRound^{\frac{5}{3}} \SelectAction^{\frac{3}{2}} \NetworkDepth^4  \RegularizationParam\!^{-\frac{1}{6}} \NetworkWidth\!^{-\frac{1}{6}}\!\sqrt{ \log \NetworkWidth} \right)} \nonumber
        \\
        & \phantom{{}={}}
        + C_2 \sqrt{8 \TotalRound \NetworkDepth \SelectAction \log\TotalRound} \bigg]
        + \mathcal{O}(1) \label{eq_bound of R1}
    \end{align}
Finally, combining Eq.\eqref{eq_bound of R1} and Eq.\eqref{eq_bound of R2} we have
    \begin{align}
        \Regret(\TotalRound)
        & \le \frac{4 C_0 \SigmaEventConstant_\TotalRound \ExplorationVariance}{\OptimisticProbability} \bigg[ \sqrt{ \TotalRound \RegularizationParam \left( 2 \EffectiveDim \log(1+ \TotalRound \TotalAction / \RegularizationParam) + 2 + C_1 \TotalRound^{\frac{5}{3}} \SelectAction^{\frac{3}{2}} \NetworkDepth^4  \RegularizationParam\!^{-\frac{1}{6}} \NetworkWidth\!^{-\frac{1}{6}}\!\sqrt{ \log \NetworkWidth} \right)} \nonumber
        \\
        & \phantom{{}={}}
        + C_2 \sqrt{8 \TotalRound \NetworkDepth \SelectAction \log\TotalRound} \bigg]
        +  2 C_0 \TotalRound \sqrt{\SelectAction} \TSErrorTerm
        + \mathcal{O}(1) \nonumber
        \\
        & \phantom{{}={}}
        + C_0 \ExplorationVariance (\SigmaEventConstant_\TotalRound + 2) \sqrt{ \TotalRound \RegularizationParam \left( 2 \EffectiveDim \log(1+ \TotalRound \TotalAction / \RegularizationParam)
        + 2 + C_1  \TotalRound^{\frac{5}{3}} \SelectAction^{\frac{3}{2}} \NetworkDepth^4  \RegularizationParam\!^{-\frac{1}{6}} \NetworkWidth\!^{-\frac{1}{6}}\!\sqrt{ \log \NetworkWidth} \right)}
    \end{align}
Then choosing $\NetworkWidth$ such that
    \begin{align*}
        & C_1 \TotalRound^{\frac{5}{3}} \SelectAction^{\frac{3}{2}} \NetworkDepth^4  \RegularizationParam\!^{-\frac{1}{6}} \NetworkWidth\!^{-\frac{1}{6}}\!\sqrt{ \log \NetworkWidth} \le 1 \, ,
        \\
        & C_{\TSErrorTerm, 1} \TotalRound^{\frac{5}{3}} \SelectAction^{\frac{2}{3}}  \NetworkDepth^{3} \RegularizationParam^{-\frac{2}{3}} \NetworkWidth^{-\frac{1}{6}} \sqrt{\log \NetworkWidth} \le \frac{1}{4} \, ,
        \\
        & C_{\TSErrorTerm, 3} \TotalRound^{\frac{13}{6}} \SelectAction^{\frac{7}{6}} \NetworkDepth^{4} \RegularizationParam^{-\frac{7}{6}} \NetworkWidth^{-\frac{1}{6}} \sqrt{\log\NetworkWidth}(1+\sqrt{\TotalRound \SelectAction/\RegularizationParam}) \le \frac{1}{4} \, ,
        \\
        & C_{\TSErrorTerm, 4} \TotalRound^{\frac{13}{6}} \SelectAction^{\frac{7}{6}} \RegularizationParam^{-\frac{2}{3}} \NetworkDepth^{\frac{9}{2}} \NetworkWidth^{-\frac{1}{6}} \sqrt{\log \NetworkWidth}
        \left( \NormParam + \SubGaussian \sqrt{\log \det (\mathbf{I} + \NTKMatrix/\RegularizationParam) + 2 - 2\log\ConfidenceParam} \right)
        \le \frac{1}{4} \, .
    \end{align*}
Also by setting $\GDSteps = 2 \log \left( \frac{1}{4 \TotalRound C_{\TSErrorTerm,2}} \frac{\sqrt{\RegularizationParam}}{\TotalRound \SelectAction \NetworkDepth}\right) \frac{\TotalRound \SelectAction \NetworkDepth}{\bar{C}_1}$, we have
    \begin{align*}
        \TotalRound  C_{\TSErrorTerm, 2} (1-\StepSize \NetworkWidth \RegularizationParam)^{\GDSteps/2} \sqrt{\TotalRound \SelectAction \NetworkDepth / \RegularizationParam} \le \frac{1}{4} \,
    \end{align*}
which follows, $\TotalRound \TSErrorTerm \le 1$.
Hence, $\Regret(\TotalRound)$ can be bounded by
    \begin{align*}
        \Regret(\TotalRound)
        & \le \frac{4 C_0 \SigmaEventConstant_\TotalRound \ExplorationVariance}{\OptimisticProbability} \bigg[ \sqrt{ \TotalRound \RegularizationParam \left( 2 \EffectiveDim \log(1+ \TotalRound \TotalAction / \RegularizationParam) + 3 \right)}
        + C_2 \sqrt{8 \TotalRound \NetworkDepth \SelectAction \log\TotalRound} \bigg]
        +  2 C_0  \sqrt{\SelectAction}
        + \mathcal{O}(1) \nonumber
        \\
        & \phantom{{}={}}
        + C_0 \ExplorationVariance (\SigmaEventConstant_\TotalRound + 2) \sqrt{ \TotalRound \RegularizationParam \left( 2 \EffectiveDim \log(1+ \TotalRound \TotalAction / \RegularizationParam)+ 3 \right)} \, .
    \end{align*}
\end{proof}

\section{Auxiliary Lemmas}

\begin{lemma} \label{lemma_anti-concentraion}
\cite{abramowitz1964handbook}
For a Gaussian distributed random variable $Z$ with mean $\mu$ and variance $\sigma^2$, for any $z \ge 1$, 
    \begin{equation*}
        \frac{1}{2\sqrt{\pi}z} \exp({-z^2/2}) \le \mathbb{P}(\left| Z - \mu \right| > z \sigma) \le \frac{1}{\sqrt{\pi}z} \exp({-z^2/2}) \, .
    \end{equation*}
\end{lemma}

\begin{lemma}[Azuma-Hoeffding inequality] \label{lemma_Azuma}
If a super-martingale $(Y_t, t \ge 0)$ corresponding to filtration $\mathcal{F}_t$, satisfies $\left| Y_t - Y_{t-1} \right| < \SigmaEventConstant_t$ for some constant $\SigmaEventConstant_t$, for all $t=1,\ldots,T$, then for any $a \ge 0$,
    \begin{equation*}
        \mathbb{P} ( Y_t - Y_{t-1} \ge a ) \le 2\exp \left({-\frac{a^2}{2 \sum_{t=1}^{T} \SigmaEventConstant_t^2}}\right) \, .
    \end{equation*}
\end{lemma}

\section{Extensions from Neural Bandits for Single Action}

In this section, we describe how the auxiliary lemmas used in the neural bandit works~\citep{zhou2020neural, zhang2021neural} for single action can be extended to the combinatorial action settings.
The main distinction is that in single action settings, the amount of data to be trained at time $t$ is $t$, whereas in combinatorial action settings, it is $tK$. Therefore, by properly accounting for this difference, we can obtain the following results.

\begin{definition}
For simplicity, we restate some definitions used in this section. 
    \begin{align*}
        & \bar{\UCBGramMatrix}_t = \RegularizationParam \mathbf{I} + \sum_{k=1}^{t} \sum_{i \in S_k} \Gradient(\Context_{k, i};\NNParam_0) \Gradient(\Context_{k, i};\NNParam_0)^\top / \NetworkWidth \, ,
        \\
        & \UCBGramMatrix_t (\text{or } \TSGramMatrix_t) = \RegularizationParam \mathbf{I} + \sum_{k=1}^{t} \sum_{i \in S_k} \Gradient(\Context_{k, i};\NNParam_{k-1}) \Gradient(\Context_{k, i};\NNParam_{k-1})^\top / \NetworkWidth \, ,
        \\
        & \bar{\sigma}_{t,i}^2 = \RegularizationParam \Gradient(\Context_{t, i};\NNParam_0)^\top \bar{\UCBGramMatrix}_{t-1} \Gradient(\Context_{t, i};\NNParam_0) / \NetworkWidth \, ,
        \\
        & \sigma_{t,i}^2 = \RegularizationParam \Gradient(\Context_{t, i};\NNParam_{t-1})^\top \TSGramMatrix_{t-1} \Gradient(\Context_{t, i};\NNParam_{t-1}) / \NetworkWidth \, ,
        \\
        & \bar{\GradientMatrix}_t = \left[ \Gradient(\Context_{1, a_{11}};\NNParam_0), \ldots, \gb(\xb_{1, a_{1K}}; \thetab_0), \ldots \Gradient(\Context_{t, a_{tK}} ;\NNParam_0) \right] \in \mathbb{R}^{p \times t \SelectAction} \, ,
        \\
        & \GradientMatrix_t = \left[ \Gradient(\Context_{1, a_{11}};\NNParam_{t-1}), \ldots, \gb(\xb_{1, a_{1K}}; \thetab_{t-1}), \ldots \Gradient(\Context_{t, a_{tK}};\NNParam_{t-1}) \right] \in \mathbb{R}^{p \times t \SelectAction} \, ,
        \\
        & \mathbf{y}_t = \left[ \Score_{1, a_{11}}, \ldots, v_{1, a_{1K}}, \ldots , \Score_{t, a_{tK}} \right]^\top \in \mathbb{R}^{t \SelectAction} \, ,
    \end{align*}
where $a_{tk}$ is the $k$-th action in the super arm $S_t$ at time $t$, i.e., $S_t:= \{a_{t1}, \cdots, a_{tK} \}$.
\end{definition}

\begin{lemma}[Lemma 5.2 in~\citet{zhou2020neural}] \label{extended lemma:lemma 5.2 in NeuralUCB}
Suppose that there exist some positive constants $\bar{C}_1, \bar{C}_2 > 0$ such that for any $\delta \in (0,1)$, $\eta \le \bar{C}_1 (T K m L + m \lambda)^{-1}$ and
    \begin{align*}
        & m \ge \bar{C}_2 K^{-\frac{1}{2}} L^{-\frac{3}{2}} \lambda^{\frac{1}{2}} \left( \log(TNL^2/\delta) \right)^{\frac{3}{2}} \, ,
        \\
        & m (\log m)^{-3} \ge \bar{C}_2 \left(
            T K L^{12} \lambda^{-1}
            + T^4 K^4 L^{18} \lambda^{-10} (\lambda +T K L)^6
            + T^7 K^7 L^{21} \lambda^{-7} (1 + \sqrt{TK/\lambda})^6 \right) \, .
    \end{align*}
Then, with probability at least $1 - \delta$, we have
    \begin{align*}
        & \| \thetab_t - \thetab_0 \|_2 \le 2 \sqrt{tK/(m\lambda)} \, ,
        \\
        & \| \thetab^* - \thetab_t \|_{\Zb_t} \le \gamma_t / \sqrt{m} \, .
    \end{align*}
\end{lemma}

\begin{lemma}[Lemma B.2 in~\citet{zhou2020neural}] \label{Lemma B.2 in NeuralUCB}
There exist some constants $\{ \bar{C}_i \}_{i=1}^{4} > 0$ such that for any $\delta \in (0,1)$, if for any $t \in [T]$, $\eta, m$ satisfy that
    \begin{align*}
        & 2\sqrt{t K /(m \lambda)} \ge \bar{C}_1 \NetworkWidth^{-\frac{3}{2}} \NetworkDepth^{-\frac{3}{2}} \left( \log(\TotalRound \TotalAction \NetworkDepth^2 / \ConfidenceParam) \right)^{\frac{3}{2}} \, ,
        \\
        & 2\sqrt{t \SelectAction /(\NetworkWidth \RegularizationParam)} \le \bar{C}_2 \min \left\{ \NetworkDepth^{-6} \left( \log\NetworkWidth \right)^{-\frac{3}{2}},
            \left( \NetworkWidth \RegularizationParam^2 \StepSize^2 \NetworkDepth^{-6} (\log \NetworkWidth)^{-1} \right)^{\frac{3}{8}} \right\} \, ,
        \\
        & \StepSize \le \bar{C}_3 (\NetworkWidth \RegularizationParam + t \SelectAction \NetworkWidth \NetworkDepth )^{-1} \, ,
        \\
        & \NetworkWidth^{\frac{1}{6}} \ge \bar{C}_4 t^{\frac{7}{6}} K^{\frac{7}{6}} \NetworkDepth^{\frac{7}{2}} \lambda^{-\frac{7}{6}} \sqrt{\log \NetworkWidth} ( 1+ \sqrt{t \SelectAction / \RegularizationParam}) \, ,
    \end{align*}
then, with probability at least $1 - \delta$, we have $\| \NNParam_t - \NNParam_0 \| \le 2 \sqrt{t \SelectAction /(\NetworkWidth \RegularizationParam)}$ and
\begin{equation*}
    \| \NNParam_t - \NNParam_0 - \bar{\UCBGramMatrix}_t^{-1} \bar{\GradientMatrix}_t \mathbf{y}_t / m \|_2
    \le
    (1 - \StepSize \NetworkWidth \RegularizationParam)^{\frac{\GDSteps}{2}} \sqrt{t\SelectAction/(\NetworkWidth \RegularizationParam)}
        + \bar{C}_5 t^{\frac{7}{6}} \SelectAction^{\frac{7}{6}} \NetworkDepth^{\frac{7}{2}} \RegularizationParam^{-\frac{7}{6}} \NetworkWidth^{-\frac{2}{3}} \sqrt{\log \NetworkWidth} ( 1+ \sqrt{t \SelectAction/\RegularizationParam}) \, ,
\end{equation*}
where $\bar{C}_5 > 0 $ is an absolute constant.
\end{lemma}

\begin{lemma}[Lemma B.3 in~\citet{zhou2020neural}] \label{extended lemma:lemma B.3 in NeuralUCB}
There exist some constants $\{ \bar{C}_i \}_{i=1}^5 > 0$ such that for any $\delta \in (0,1)$, if for any $t \in [T]$, $m$ satisfies that
    \begin{equation*}
        \bar{C}_1 m^{-\frac{3}{2}} L^{-\frac{3}{2}} \left( log(T N L^2/\delta) \right)^{\frac{3}{2}} 
        \le 2 \sqrt{tK/(m \lambda)} \le
        \bar{C}_2 L^{-6} (\log m )^{-\frac{3}{2}} \, ,
    \end{equation*}
then, with probability at least $1 - \delta$, for any $t \in [T]$ we have
    \begin{align*}
        & \| \Zb_t \|_2 \le \lambda + \bar{C}_3 t K L \, ,
        \\
        & \| \bar{\Zb}_t - \Zb_t \|_F \le \bar{C}_4 t^{\frac{7}{6}} K^{\frac{7}{6}} L^4 \lambda^{-\frac{1}{6}} m^{-\frac{1}{6}}\sqrt{\log m} \, ,
        \\
        & \left| \log \frac{\det \bar{\Zb}_t}{\det \lambda \Ib} - \log \frac{\det \Zb_t}{\det \lambda \Ib} \right|
            \le \bar{C}_5 t^{\frac{5}{3}} K^{\frac{5}{3}} L^4 \lambda^{-\frac{1}{6}} m^{-\frac{1}{6}} \sqrt{\log m} \, ,
    \end{align*}
where $\bar{C}_3, \bar{C}_4, \bar{C}_5 >0$ are some absolute constants, and $\bar{\Zb}_t = \lambda \Ib + \sum_{k=1}^{t-1} \sum_{i \in S_k} \gb(\xb_{k,i}; \thetab_0) \gb(\xb_{k,i}; \thetab_0)^\top$.

\end{lemma}

\begin{lemma}[Lemma C.2 in~\citet{zhou2020neural}] \label{aux lemma:lemma C.2 in NeuralUCB}
For any $\delta \in (0,1)$, $ \bar{C}_1, \bar{C}_2 > 0$, suppose that $\tau$ satisfies
    \begin{equation*}
        \bar{C}_1 m^{-\frac{3}{2}} L^{-\frac{3}{2}} \left( \log(T N L^2 / \delta) \right)^{\frac{3}{2}}
        \le \tau \le
        \bar{C}_2 L^{-6} (\log m)^{-\frac{3}{2}} \, ,
    \end{equation*}
Then, with probability at least $1 - \delta$, if for any $j \in [J]$, $\| \thetab^{(j)} - \thetab^{(0)} \|_2 \le \tau$, we have the following results for any $j, s \in [J]$,
    \begin{align*}
        & \| \Jb^{(j)} \|_F \le \bar{C}_3 \sqrt{t K m L} \, ,
        \\
        & \| \Jb^{(j)} - \Jb^{(0)} \|_F \le \bar{C}_4 \tau^{\frac{1}{3}} L^{\frac{7}{2}} \sqrt{t K m \log m}  \, ,
        \\
        & \| \fb^{(s)} - \fb^{(j)} - ( \Jb^{(j)})^\top ( \thetab^{(s)} - \thetab^{(j)} ) \|_2 \le \bar{C}_5 \tau^{\frac{4}{3}} L^3 \sqrt{t K m \log m} \, ,
        \\
        & \| \yb \|_2 \le \sqrt{tK} \, ,
    \end{align*}
where $\bar{C}_3, \bar{C}_4, \bar{C}_5 >0$ are some absolute constants.
\end{lemma}

\begin{lemma}[Lemma C.3 in~\citet{zhou2020neural}] \label{aux lemma:lemma C.3 in NeuralUCB}
For any $\delta \in (0,1)$ and $ \{ \bar{C}_i \}_{i=1}^4 > 0$, suppose that $\tau, \eta$ satisfy
    \begin{align*}
        & \bar{C}_1 m^{-\frac{3}{2}} L^{-\frac{3}{2}} \left( \log(T N L^2/\delta) \right)^{\frac{3}{2}} 
            \le 
            \bar{C}_2 L^{-6} (\log m)^{-\frac{3}{2}} \, ,
        \\
        & \eta \le \bar{C}_3 (m \lambda + tKmL)^{-1} \, ,
        \\
        & \tau^{\frac{8}{3}} \le \bar{C}_4 m \lambda^2 \eta^2 L^{-6} (\log m )^{-1} \, .
    \end{align*}
Then, with probability at least $1 - \delta$, if for any $j \in [J]$, $\| \thetab^{(j)} - \thetab^{(0)} \|_2 \le \tau$, we have that for any $j \in [J]$, $\| \fb^{(j)} - \yb \|_2 \le 2 \sqrt{tK}$.
\end{lemma}

\begin{lemma}[Lemma C.4 in~\citet{zhou2020neural}] \label{aux lemma:lemma C.4 in NeuralUCB}
For any $\delta \in (0,1)$ and $ \{ \bar{C}_i \}_{i=1}^3 > 0$, suppose that $\tau, \eta$ satisfy
    \begin{align*}
        & \bar{C}_1 m^{-\frac{3}{2}} L^{-\frac{3}{2}} \left( \log(T N L^2/\delta) \right)^{\frac{3}{2}} 
            \le 
            \bar{C}_2 L^{-6} (\log m)^{-\frac{3}{2}} \, ,
        \\
        & \eta \le \bar{C}_3 (m \lambda + tKmL)^{-1} \, .
    \end{align*}
Then, with probability at least $1 - \delta$, we have for any $j \in [J]$,
    \begin{align*}
        & \| \tilde{\thetab}^{(j)} - \thetab^{(0)} \|_2 \le \sqrt{tK/(m \lambda)} \, ,
        \\
        & \| \tilde{\thetab}^{(j)} - \thetab^{(0)} - \bar{\Zb}^{-1} \bar{\Jb} \yb / m \|_2 \le (1 - \eta m \lambda)^{\frac{j}{2}} \sqrt{tK/(m \lambda)} \, .
    \end{align*}
\end{lemma}

\section{Extension of Regret Analysis to $\alpha$-approximation Oracle} \label{appendix_Alpha Oracle}
In this section, we extend our regret analysis to the case when the agent only has access to an $\alpha$-approximation oracle, $\mathbb{O}_{\mathcal{S}}^{\alpha}$.
First, we replace $S_t$ with $S_t^{\alpha} = \mathbb{O}_{\mathcal{S}}^{\alpha} (\mathbf{u}_t + \mathbf{e}_t)$ for $\CNUCB$ (Algorithm~ \ref{alg:CN-UCB}) and $S_t^{\alpha} = \mathbb{O}_{\mathcal{S}}^{\alpha} (\mathbf{\tilde{v}}_t + \boldsymbol{\epsilon})$ for $\CNTS$ (Algorithm \ref{alg:CN-TS}).
The total regret $\Regret(\TotalRound)$ is replaced with an $\alpha$-regret defined as:
$$\mathcal{R}^{\alpha}(T) = \sum_{t=1}^{T} \mathbb{E} \left[ \alpha R(S_t^*, \mathbf{v}_t^*) - R(S_t^\alpha, \mathbf{v}_t^*) \right]$$.

For $\CNUCB$, note that $\alpha R(S_t,\mathbf{u}_t + \mathbf{e}_t) \le R(S_t^\alpha,\mathbf{u}_t + \mathbf{e}_t)$.
Also, $\alpha R(S_t^*, \mathbf{v}_t^*) \le \alpha R(S_t^*,\mathbf{u}_t + \mathbf{e}_t) \le \alpha R(S_t,\mathbf{u}_t + \mathbf{e}_t) \le R(S_t^\alpha,\mathbf{u}_t + \mathbf{e}_t) $ .  
We can derive that the $\alpha$-regret bound of $\CNUCB$ is $\OTilde(\tilde{d}\sqrt{T})$ or $\OTilde(\sqrt{ \EffectiveDim \TotalRound \SelectAction})$, whichever is higher, by substituting the following notations in Appendix~\ref{subsec:proof of CN-UCB}:
\begin{gather*}
    R(S_t^*, \mathbf{v}_t^*) \xrightarrow{} \alpha R(S_t^*, \mathbf{v}_t^*) \\ R(S_t^*,\mathbf{u}_t + \mathbf{e}_t) \xrightarrow{} \alpha R(S_t^*,\mathbf{u}_t + \mathbf{e}_t) \\ 
    S_t \xrightarrow{} S_t^\alpha \, .
\end{gather*}

For $\CNTS$, note that $\alpha R(S_t,\mathbf{\tilde{v}}_t + \boldsymbol{\epsilon}) \le R(S_t^\alpha,\mathbf{\tilde{v}}_t + \boldsymbol{\epsilon})$.
We split $\alpha$-regret as follows:
\begin{align*}
    \mathcal{R}^{\alpha}(T) & = \mathcal{R}_1^{\alpha}(T) + \mathcal{R}_2^{\alpha}(T) \\
    & = \sum_{t=1}^{T} \mathbb{E} \left[ \alpha R(S_t^*, \mathbf{v}_t^*) - \alpha R(S_t^\alpha, \mathbf{\tilde{v}}_t + \boldsymbol{\epsilon}) \right] \\
    & + \sum_{t=1}^{T} \mathbb{E} \left[ \alpha R(S_t^\alpha, \mathbf{\tilde{v}}_t + \boldsymbol{\epsilon}) - R(S_t^\alpha, \mathbf{v}_t^*) \right] \, .
\end{align*}

By replacing $S_t$ with $S_t^\alpha$ in Appendix~\ref{subsec:proof of CN-TS}, we can get the $\alpha$-regret bound of $\mathcal{R}_2^{\alpha}(T)$. 
For $\mathcal{R}_1^{\alpha}(T)$, since 
$\alpha R(S_t^*, \mathbf{v}_t^*) - \alpha R(S_t^\alpha, \mathbf{\tilde{v}}_t + \boldsymbol{\epsilon})
\le \alpha R(S_t^*, \mathbf{v}_t^*) - \alpha R(S_t, \mathbf{\tilde{v}}_t + \boldsymbol{\epsilon}) 
\le R(S_t^*, \mathbf{v}_t^*) - R(S_t, \mathbf{\tilde{v}}_t + \boldsymbol{\epsilon})$, 
we know that $\mathcal{R}_1^{\alpha}(T) \le \mathcal{R}_1(T)$.  
By combining the results, we can conclude that the $\alpha$-regret bound of $\CNTS$ is $\OTilde(\EffectiveDim\sqrt{\TotalRound  \SelectAction})$.

\section{When Time Horizon $\TotalRound$ Is Unknown}\label{sec:unknown_T}

For Theorems~\ref{thm_CN-UCB} and \ref{thm_CN-TS}, 
we assumed that $\TotalRound$ is known for the sake of clear exposition for our proposed algorithms and their regret analysis.
However, the knowledge of $\TotalRound$ is not essential both for the algorithms and their analysis.
With slight modifications, our proposed algorithms can be applied to the settings where $\TotalRound$ is unknown.
In this section, we propose the variants of $\CNUCB$ and $\CNTS$: $\CNUCBD$ and $\CNTSD$, and show that their regret upper bounds are of the same order of regret as those of $\CNUCB$ and $\CNTS$ up to logarithmic factors.

\subsection{Algorithms}

$\CNUCBD$ and $\CNTSD$ utilize a doubling technique \cite{besson2018doubling} in which the network size stays fixed during each epoch but is updated after the end of each epoch whose length $\EpochPeriod$ doubles the length of a previous epoch.
This way, even when $\TotalRound$ is unknown, the networks size can be set adaptively over epochs.

The algorithms first initialize the variables related to $\EpochPeriod$, especially the hidden layer width $\NetworkWidth_{\EpochPeriod}$ and the number of parameters of the neural network $p(\EpochPeriod)$.
For each round, after playing super arm $S_t$ and observing the scores $\{ \Score_{t, i} \}_{i \in S_t}$,
$\CNUCBD$ and $\CNTSD$ call the $\Update$ algorithm.
Until $\EpochPeriod$, $\Update$ algorithm updates $\NNParam_{t}$ and $\UCBGramMatrix_{t}$ or $\TSGramMatrix_{t}$ as if $\EpochPeriod$ is the time horizon.
If $t$ reaches $\EpochPeriod$, $\Update$ algorithm \textit{doubles} the value of $\EpochPeriod$.
After reinitializing the variables related to the \textit{doubled} $\EpochPeriod$,
which includes reconstructing the neural network to have a larger hidden layer width $\NetworkWidth_{\EpochPeriod}$,
the algorithm updates all of the $\NNParam_{t'}$ and $\UCBGramMatrix_{t'}$ or $\TSGramMatrix_{t'}$ for $t'= 0, \cdots, t$.
    $\Update$ algorithm returns $\NNParam_{t}$ and $\UCBGramMatrix_{t}$ or $\TSGramMatrix_{t}$ to $\CNUCBD$ or $\CNTSD$.
This process continues until $t$ reaches $\TotalRound$.

Note that the computation complexity of each round of $\CNUCB$ and $\CNUCBD$ heavily depends on how quickly they can compute the inverse of the gram matrix $\UCBGramMatrix$.
Since $\UCBGramMatrix \in \mathbb{M}_p(\mathbb{R})$, and $p$ depends on $\NetworkWidth$, the computation speed of each round in $\CNUCB$ is relatively slow as $\NetworkWidth$ is a large constant.  On the other hand, $\CNUCBD$ can show faster computation speed, especially at the beginning rounds, where $\NetworkWidth$ is kept relatively small.
The same argument can be applied to $\CNTS$ and $\CNTSD$.

$\CNUCBD$ is summarized in Algorithm~\ref{alg:CN-UCB-D}.
$\CNTSD$ is summarized in Algorithm~\ref{alg:CN-TS-D}.
The $\Update$ algorithm is summarized in Algorithm~\ref{alg:Update}.

\begin{algorithm*}[ht]
   \caption{$\CNUCBD$}
   \label{alg:CN-UCB-D}
\begin{algorithmic}
    \STATE {\bfseries Input:} Epoch period $\EpochPeriod$, network depth $\NetworkDepth$.
    \STATE {\bfseries Initialization: } Initialize \{network width $\NetworkWidth_\EpochPeriod$,
        regularization parameter $\RegularizationParam_\EpochPeriod$,
        norm parameter $\NormParam_\EpochPeriod$,
        step size $\StepSize_\EpochPeriod$,
        number of gradient descent steps $\GDSteps_\EpochPeriod$\} with respect to $\EpochPeriod$, \,
        set number of parameters of neural network $p(\EpochPeriod) = \NetworkWidth_\EpochPeriod d + \NetworkWidth_\EpochPeriod^2 (\NetworkDepth-2) + \NetworkWidth_\EpochPeriod$, \,
        $\UCBGramMatrix_{0} = \RegularizationParam_{\EpochPeriod} \mathbf{I}_{p(\EpochPeriod)}$, \,
        randomly initialize $\NNParam_{0}$ as described in Section \ref{subsec:CN-UCB}
    
    \WHILE{$t \neq \TotalRound$}
    \STATE Observe $\{ \mathbf{x}_{t,i} \}_{i \in [\TotalAction]}$
    \STATE Compute $\hat{\Score}_{t,i} = f(\mathbf{x}_{t,i}; \NNParam_{t-1})$ and 
    $u_{t,i} = \hat{\Score}_{t,i} + \gamma_{t-1} \left\| \Gradient(\mathbf{x}_{t,i}; \NNParam_{t-1}) / \sqrt{\NetworkWidth_\EpochPeriod} \right\|_{\UCBGramMatrix_{t-1}^{-1}}$ for $i \in [\TotalAction]$
    \STATE Let $S_t = \Oracle_{\mathcal{S}} (\mathbf{u}_t + \mathbf{\UCBErrorTerm}_t)$ 
    \STATE Play super arm $S_t$ and observe $\{ \Score_{t, i} \}_{i \in S_t}$
    \STATE \Update($t, \EpochPeriod$)
    \STATE Compute $\gamma_t$ and $e_{t+1}$ described in lemma~\ref{lemma_u'_modified UCB} (replace
    \{$\RegularizationParam$,
    $\NetworkWidth$,
    $\mathbf{I}$,
    $\NormParam$,
    $\StepSize$,
    $\GDSteps$\} with
    \{$\RegularizationParam_\EpochPeriod$,
    $\NetworkWidth_\EpochPeriod$,
    $\mathbf{I}_{p(\EpochPeriod)}$,
    $\NormParam_\EpochPeriod$,
    $\StepSize_\EpochPeriod$,
    $\GDSteps_\EpochPeriod$\})
    \ENDWHILE
\end{algorithmic}
\end{algorithm*}

\begin{algorithm*}[ht]
  \caption{$\CNTSD$}
  \label{alg:CN-TS-D}
\begin{algorithmic}
  \STATE {\bfseries Input:} Epoch period $\EpochPeriod$,
  network depth $\NetworkDepth$,
  sample size $\TotalSampleNum$
  \STATE {\bfseries Initialization: } Initialize \{network width $\NetworkWidth_{\EpochPeriod}$,
  regularization parameter $\RegularizationParam_{\EpochPeriod}$,
  exploration variance $\ExplorationVariance_{\EpochPeriod}$, 
  step size $\StepSize_{\EpochPeriod}$, 
  number of gradient descent steps $\GDSteps_{\EpochPeriod}$\} with respect to $\EpochPeriod$, \,
  set number of parameters of neural network $p(\EpochPeriod) = \NetworkWidth_{\EpochPeriod} d + \NetworkWidth_{\EpochPeriod}^{2} (L-2) + \NetworkWidth_{\EpochPeriod}$, \,
  $\TSGramMatrix_{0} = \RegularizationParam_{\EpochPeriod} \mathbf{I}_{p(\EpochPeriod)}$, \,
  randomly initialize $\NNParam_{0}$  as described in Section \ref{subsec:CN-UCB} 
  \WHILE{$t \neq \TotalRound$}
  \STATE Observe $\{ \mathbf{x}_{t,i} \}_{i \in [\TotalAction]}$.
  \STATE Compute $\sigma_{t, i}^2 = \RegularizationParam_{\EpochPeriod} \Gradient(\Context_{t,i}; \NNParam_{t-1})^\top \TSGramMatrix_{t-1}^{-1} \Gradient(\Context_{t,i}; \NNParam_{t-1}) 
  / \NetworkWidth_\EpochPeriod$ for each $i \in [\TotalAction]$
  \\
  \STATE Sample $\{ \tilde{\Score}_{t,i}^{(j)} \}_{j=1}^{\TotalSampleNum}$ independently from $\mathcal{N}( f(\Context_{t,i}; \NNParam_{t-1}), \ExplorationVariance_{\EpochPeriod}^{2} \sigma_{t,i}^2)$ for each $i \in [\TotalAction]$
  \STATE Compute $\tilde{\Score}_{t, i} = \max_{j} \tilde{\Score}_{t,i}^{(j)}$ for each $i \in [\TotalAction]$
  \STATE Let $S_t = \Oracle_{\mathcal{S}} (\tilde{\ScoreVector}_t + \boldsymbol{\TSErrorTerm})$ 
  \STATE Play super arm $S_t$ and observe $\{ \Score_{t, i} \}_{i \in S_t}$
  \STATE \Update($t, \EpochPeriod$)
  \ENDWHILE
\end{algorithmic}
\end{algorithm*}

\begin{algorithm}[ht]
  \caption{\Update($t, \EpochPeriod$)}
  \label{alg:Update}
\begin{algorithmic}
    \STATE {\bfseries Input:} Epoch period $\EpochPeriod$, round $t$
    \IF{$t < \EpochPeriod$ }
    \STATE Update $\UCBGramMatrix_{t} = \UCBGramMatrix_{t-1} + \sum_{i \in S_t} \Gradient(\Context_{t,i}; \NNParam_{t-1})  \Gradient(\Context_{t,i}; \NNParam_{t-1})^\top / \NetworkWidth_{\EpochPeriod}$
    \STATE Update $\NNParam_{t}$ to minimize the loss Eq.\eqref{eq_L2-loss} using gradient descent with $\StepSize_{\EpochPeriod}$ for $\GDSteps_{\EpochPeriod}$ times
    \ELSE
    
    \STATE $\EpochPeriod \xleftarrow{} 2\EpochPeriod$
    
    \STATE Reinitialize \{$\NetworkWidth_{\EpochPeriod}$,
    $\RegularizationParam_{\EpochPeriod}$,
    $\StepSize_{\EpochPeriod}$, 
    $\GDSteps_{\EpochPeriod}$\} with respect to $\EpochPeriod$,\, 
    set $p(\EpochPeriod) = \NetworkWidth_{\EpochPeriod} d + \NetworkWidth_{\EpochPeriod}^{2} (L-2) + \NetworkWidth_{\EpochPeriod}$, \,
    randomly reinitialize $\NNParam_{0}$ as described in Section \ref{subsec:CN-UCB}.
    
    For $\CNUCBD$,
    reinitialize $\NormParam_{\EpochPeriod}$ with respect to $\EpochPeriod$, \,
    $\UCBGramMatrix_{0} = \RegularizationParam_{\EpochPeriod} \mathbf{I}_{p(\EpochPeriod)}$
    
    For $\CNTSD$,
    reinitialize $\ExplorationVariance_{\EpochPeriod}$ with respect to $\EpochPeriod$, \,
    $\TSGramMatrix_{0} = \RegularizationParam_{\EpochPeriod} \mathbf{I}_{p(\EpochPeriod)}$
    
    \FOR{$t'=1, \cdots, t$}

    \STATE For $\CNUCBD$, $ \UCBGramMatrix_{t'} = \UCBGramMatrix_{t'-1} + \sum_{i \in S_{t'}} \Gradient (\Context_{t',i}; \NNParam_{t'-1}) \Gradient(\Context_{t',i};\NNParam_{t'-1})^\top / \NetworkWidth_{\EpochPeriod}$    
    \STATE For $\CNTSD$, $ \TSGramMatrix_{t'} = \TSGramMatrix_{t'-1} + \sum_{i \in S_{t'}} \Gradient (\Context_{t',i}; \NNParam_{t'-1}) \Gradient(\Context_{t',i};\NNParam_{t'-1})^\top / \NetworkWidth_{\EpochPeriod}$    
    \STATE Update $\NNParam_{t'}$ to minimize the loss (4)  using gradient descent with $\StepSize_{\EpochPeriod}$ for $\GDSteps_{\EpochPeriod}$ times
    
    \ENDFOR
    \ENDIF
    \STATE {\bfseries Return: } $\NNParam_{t}$, $\UCBGramMatrix_{t}$ or $\TSGramMatrix_{t}$
\end{algorithmic}
\end{algorithm}

\subsection{Regret Analysis}
The regret upper bounds of $\CNUCBD$ and $\CNUCB$ (or $\CNTSD$ and $\CNTS$) have the same rate up to logarithmic factors.
We provide the sketch of proof.

By modifying Definitions~\ref{def_NTK matrix} and \ref{def_effective dimension} with respect to $\EpochPeriod$, the effective dimension $\EffectiveDim_{\EpochPeriod}$ can be written as $\EffectiveDim_{\EpochPeriod} = \log \det ( \mathbf{I} + \mathbf{H_{\EpochPeriod}}/\RegularizationParam_{\EpochPeriod} ) / \log (1+\EpochPeriod \TotalAction / \RegularizationParam_{\EpochPeriod} )$.  
Denote the epoch periods as $\EpochPeriod_{n} = 2^{n}\EpochPeriod_{0}$, where $n \in \mathbb{Z}_{\ge 0}$ and $\EpochPeriod_{0}$ is the initial epoch period.

If $\TotalRound < \EpochPeriod_{0}$, $\CNUCBD$ and $\CNTSD$ are equivalent to $\CNUCB$ and $\CNTS$ respectively. 
In this case, there is no change in the regret upper bounds.
Meanwhile, if $\TotalRound \ge \EpochPeriod_{0}$, there exists $\hat{n} \in \mathbb{Z}_{+}$ such that $\EpochPeriod_{\hat{n}-1} \le T < \EpochPeriod_{\hat{n}}$.  Denote the instantaneous regret as $\InstantaneousRegret_t$. Define $\sum_{t=a}^{b} \InstantaneousRegret_t := 0$ if $a > b$.
Then the regret can be written as
\begin{align*}
        \Regret(\TotalRound) 
        & = \sum_{t=1}^{\EpochPeriod_{0}} \InstantaneousRegret_t 
        + \sum_{t=\EpochPeriod_{0} + 1}^{\EpochPeriod_{1}} \InstantaneousRegret_t 
        + \cdots
        + \sum_{\EpochPeriod_{\hat{n}-2} + 1}^{\EpochPeriod_{\hat{n}-1}} \InstantaneousRegret_t
        + \sum_{t=\EpochPeriod_{\hat{n}-1} + 1}^{\TotalRound} \InstantaneousRegret_t \, .
\end{align*}

Let $\EffectiveDim := \max \{\EffectiveDim_{\EpochPeriod_0}, \ldots, \EffectiveDim_{\EpochPeriod_{\hat{n}}} \}$. 
For $\CNUCBD$, each sum has an upper bound $\OTilde (\max\{\EffectiveDim_{\EpochPeriod_{n}}, \sqrt{\EffectiveDim_{\EpochPeriod_{n}} \SelectAction} \} \sqrt{\EpochPeriod_{n}})$. 
Thus, the regret is bounded by $\OTilde (\max\{\EffectiveDim, \sqrt{\EffectiveDim \SelectAction}\} \sqrt{ 2\TotalRound})$\,.  
Similarly, for $\CNTSD$, each sum has upper bound $\OTilde (\EffectiveDim_{\EpochPeriod_{n}} \sqrt{\EpochPeriod_{n}\SelectAction})$ and the regret has upper bound $\OTilde (\EffectiveDim \sqrt{2\TotalRound \SelectAction})$\,. 

\section{Specific Examples of Combinaotrial Feedback Models} \label{appendix_combinaotiral feedbacks}

As mentioned in Remark~\ref{remark_extension of reward}, algorithms having a reward function satisfying Assumptions~\ref{assum:monotone} and ~\ref{assum:lips} encompasses various combinatorial feedback models, suggesting that these assumptions are not restrictive.
In this section, we provide specific examples.

\subsection{Semi-bandit Model}
In the semi-bandit setting, after choosing a superarm, the agent observes all of the scores (or feedback) associated with the superarm and receives a reward as a function of the scores.
The main text of this paper describes how our algorithms cover semi-bandit feedback models.
Recall that in semi-bandit setting, if the feature vectors are independent then the score of each arm is independent. 
Meanwhile, in ranking models (or click models), chosen arms may have a position within the superarm, and the scores of arms may depend on its own attractiveness as well as its position.

\subsection{Document-based Model}
The document-based model is a click model that assumes the scores of an arm are identical to its attractiveness. 
The attractiveness of an arm is determined by the context of arm.
Formally, for each arm $i \in [\TotalAction]$, let $\alpha(\Context_{t,i}) \in [0, 1]$ be the attractiveness of arm $i$ at time $t$.
Then the document-based model assumes that the score function of $\Context_{t,i}$ in the $k$-th position is defined as
\begin{equation} \label{eq:h in document-based model}
    h(\Context_{t,i}, k) = \alpha(\Context_{t,i}) \ind(k \le \SelectAction) \, .
\end{equation}
Note that $h$ in Eq.\eqref{eq:h in document-based model} is bounded in $[0,1]$. 
Since a neural network is a universal approximator, we can utilize neural networks to estimate the score of arm $i$ in position $k$ as follows:
\begin{equation*}
    \hat{h}(\Context_{t,i}, k) = f(\Context_{t,i}, k ; \NNParam_{t-1}) \, .
\end{equation*}
Note that for any $k \in [\SelectAction]$, the score of an arm only depends on the attractiveness of the arm. 
Hence, our algorithms can be directly applicable to the document-based model without any modification.

\subsection{Position-based Model}
In the document-based model, the score of an arm is invariant to the position within the super arm. 
However, in the position-based model, the score of a chosen arm varies depending on its position. 
Let $\chi: [\SelectAction] \rightarrow [0,1]$ be a function that measures the quality of a position within the super arm. 
The position-based model assumes that the score function of a chosen arm associated to $\Context_{t,i}$ and located in the $k$-th position is defined as
\begin{equation}
    h(\Context_{t,i}, k) = \alpha(\Context_{t,i}) \chi(k) \,.
\end{equation}
Note that the score of an arm can change as its position moves within the superarm.
We can slightly modify our suggested algorithms to reflect this.
First, we introduce a modified neural network $\dot{f}(\Context_{t,i}, k; \NNParam_{t-1})$ that estimates the score of each arm at every available position.
By this, the action space of each round increases from $\TotalAction$ to $\TotalAction\SelectAction$.
The regret bound only changes as much as the action space increases.
Denote the gradient of $\dot{f}(\Context_{t,i}, k; \NNParam_{t-1})$ as $\dot{\Gradient}(\Context_{t,i}, k; \NNParam_{t-1})$.

Furthermore, we replace the oracle to $\dot{\Oracle}_{\mathcal{S}} \left( \{u_{t,i}(k) + \UCBErrorTerm_t \}_{i \in [\TotalAction], k \in [\SelectAction]} \right)$ that considers the position of the arms.  
The oracle $\dot{\Oracle}_{\mathcal{S}}$ chooses only one arm for one position.
Also, an arm that has been chosen for a certain position cannot be chosen for another position.
As an optimization problem having the above constraints can be solved with linear programming, $\dot{\Oracle}_{\mathcal{S}} \left( \{u_{t,i}(k) + \UCBErrorTerm_t \}_{i \in [\TotalAction], k \in [\SelectAction]} \right)$ can compute exact optimization within polynomial time.
Modified algorithm for a position-based model is described in Algorithm~\ref{alg:CNB for position-based model}.

\begin{algorithm*}[ht]
   \caption{Combinatorial neural bandits for for position-based model}
   \label{alg:CNB for position-based model}
\begin{algorithmic}
    \STATE Initialize as Algorithm~\ref{alg:CN-UCB}
    \FOR{$t = 1, ..., T$} 
    \STATE Observe $\{ \mathbf{x}_{t,i} \}_{i \in [\TotalAction]}$
    \IF {Exploration == UCB}
        \STATE Compute $u_{t,i}(k) = \dot{f}(\Context_{t,i}, k; \NNParam_{t-1}) + \gamma_{t-1} \| \dot{\Gradient}(\Context_{t,i}, k; \NNParam_{t-1}) / \sqrt{\NetworkWidth}) \|_{\UCBGramMatrix_{t-1}^{-1}}$ for $i \in [\TotalAction], k \in [\SelectAction]$
        \STATE Let $S_t = \dot{\Oracle}_{\mathcal{S}} \left( \{u_{t,i}(k) + \UCBErrorTerm_t \}_{i \in [\TotalAction], k \in [\SelectAction]} \right)$
    \ELSIF {Exploration == TS}
        \STATE Compute $ \sigma_{t,i}^2(k) = \RegularizationParam \dot{\Gradient}(\Context_{t,i}, k ; \NNParam_{t-1})^\top \TSGramMatrix_{t-1}^{-1} \dot{\Gradient}(\Context_{t,i}, k; \NNParam_{t-1}) / \NetworkWidth$ for $i \in [\TotalAction], k \in [\SelectAction]$
        \STATE Sample $\{ \tilde{\Score}_{t,i}^{(j)}(k) \}_{j=1}^{\TotalSampleNum}$ independently from $\mathcal{N}( \dot{f}(\Context_{t,i}, k; \NNParam_{t-1}), \ExplorationVariance^2 \sigma_{t,i}^2(k))$ for $i \in [\TotalAction], k \in [\SelectAction]$
        \STATE Compute $\tilde{\Score}_{t, i}(k)= \max_{j} \tilde{\Score}_{t,i}^{(j)}(k)$ for $i \in [\TotalAction], k \in [\SelectAction]$    
        \STATE Let $S_t = \dot{\Oracle}_{\mathcal{S}} \left( \{\tilde{\Score}_{t, i}(k) + \TSErrorTerm \}_{i \in [\TotalAction], k \in [\SelectAction]} \right)$
    \ENDIF
    \STATE Play super arm $S_t$ and observe $\{ \Score_{t, i}(k_i) \}_{i \in S_t}$
    \STATE (UCB) Update $\UCBGramMatrix_{t} = \UCBGramMatrix_{t-1} + \sum_{i \in S_t} \dot{\Gradient}(\Context_{t,i}, k_i; \NNParam_{t-1})  \dot{\Gradient}(\Context_{t,i}, k_i; \NNParam_{t-1})^\top / \NetworkWidth$
    \STATE (TS) Update $\TSGramMatrix_{t} = \TSGramMatrix_{t-1} + \sum_{i \in S_t} \dot{\Gradient}(\Context_{t,i}, k_i ; \NNParam_{t-1}) \dot{\Gradient} (\Context_{t,i}, k_i; \NNParam_{t-1})^\top /\NetworkWidth$
    \STATE Update $\NNParam_{t}$ to minimize the loss in Eq.\eqref{eq_L2-loss} using gradient descent with $\StepSize$ for $\GDSteps$ times
  \ENDFOR
\end{algorithmic}
\end{algorithm*}

\begin{algorithm*}[ht]
   \caption{Combinatorial neural bandits for for cascade feedback model}
   \label{alg:CNB for cascade model}
\begin{algorithmic}
    \STATE Initialize as Algorithm~\ref{alg:CN-UCB}, $\{ \psi_k \in [0,1] \}_{k \in [\SelectAction]} $: position discount factors
    \FOR{$t = 1, ..., T$} 
    \STATE Observe $\{ \mathbf{x}_{t,i} \}_{i \in [\TotalAction]}$
    \IF {Exploration == UCB}
        \STATE Compute $u_{t,i} = f(\Context_{t,i}; \NNParam_{t-1}) + \gamma_{t-1} \| \Gradient(\Context_{t,i}, k; \NNParam_{t-1}) / \sqrt{\NetworkWidth}) \|_{\UCBGramMatrix_{t-1}^{-1}}$ for $i \in [\TotalAction]$
        \STATE Let $S_t = \Oracle_{\mathcal{S}} \left( \{u_{t,i} + \UCBErrorTerm_t \}_{i \in [\TotalAction]} \right)$
    \ELSIF {Exploration == TS}
        \STATE Compute $ \sigma_{t,i}^2 = \RegularizationParam \Gradient(\Context_{t,i} ; \NNParam_{t-1})^\top \TSGramMatrix_{t-1}^{-1} \Gradient(\Context_{t,i}; \NNParam_{t-1}) / \NetworkWidth$ for $i \in [\TotalAction]$
        \STATE Sample $\{ \tilde{\Score}_{t,i}^{(j)}\}_{j=1}^{\TotalSampleNum}$ independently from $\mathcal{N}( f(\Context_{t,i}; \NNParam_{t-1}), \ExplorationVariance^2 \sigma_{t,i}^2)$ for $i \in [\TotalAction]$
        \STATE Compute $\tilde{\Score}_{t, i}= \max_{j} \tilde{\Score}_{t,i}^{(j)}$ for $i \in [\TotalAction]$    
        \STATE Let $S_t = \Oracle_{\mathcal{S}} \left( \{\tilde{\Score}_{t, i} + \TSErrorTerm \}_{i \in [\TotalAction]} \right)$
    \ENDIF
    \STATE Play super arm $S_t$ and observe $\mathfrak{F}_t, \{ \psi_k v_{t,k} \}_{k \in [\mathfrak{F}_t]}$
    \STATE (UCB) Update $\UCBGramMatrix_{t} = \UCBGramMatrix_{t-1} + \sum_{k \in [\mathfrak{F}_t]} \Gradient(\Context_{t,k}; \NNParam_{t-1})  \Gradient(\Context_{t,i}; \NNParam_{t-1})^\top / \NetworkWidth$
    \STATE (TS) Update $\TSGramMatrix_{t} = \TSGramMatrix_{t-1} + \sum_{k \in [\mathfrak{F}_t]} \Gradient(\Context_{t,k}; \NNParam_{t-1}) \Gradient (\Context_{t,k}; \NNParam_{t-1})^\top /\NetworkWidth$
    \STATE Update $\NNParam_{t}$ to minimize the loss in Eq.\eqref{eq_L2-loss} using gradient descent with $\StepSize$ for $\GDSteps$ times
  \ENDFOR
\end{algorithmic}
\end{algorithm*}

\subsection{Cascade Model}
In the cascade model, the agent suggests arms to a user one-by-one in order of the positions of the arms within the superarm.
The user scans the arms one-by-one until she selects an arm that she likes, which ends the suggestion procedure. Note that the suggestion procedure potentially may end before the agent shows all the arms in the superarm to the user. Also, the user may not select any arm after she scans all the arms in the superarm.
Hence, unlike the previously mentioned models, where the agent receives all of the scores of the chosen arms, 
in the cascade model, the agent only receives the scores of the arms observed by the user. 

Let us assume that the score the agent receives when the user selects an arm in the $1$-st position is $1$. 
In case the same arm is in the $k$-th position, the score the agent receives when the user selects the same arm must be less than $1$.
To reflect this feature, we consider a position discount factor $\psi_k \in [0,1], k \le \SelectAction$ that is multiplied to the attractiveness of the arm.
The observed score of an arm is determined by its attractiveness and the position discount factor that is multiplied to it.
The mechanism estimating the attractiveness using a neural network is same as the one for the semi-bandits.
The only difference is that the agent only receives the discounted scores of the arms observed by the user.

Suppose that the user selects $\mathfrak{F}_t$-th arm.
Then the agent observes the discounted scores for the first $\mathfrak{F}_t$ arms in $S_t$. 
Update is based on the discounted scores, $\psi_k \Score_{t,k}, k \le \mathfrak{F}_t$. 
An adjusted Algorithm for the cascade model is described in Algorithm~\ref{alg:CNB for cascade model}.
In addition, in case we have no information of the position discount factor, we can deal with the cascade model same as the position-based model.

\section{Additional Related Work}
As mentioned in Section~\ref{sec:intro}, the proposed methods are the first neural network-based combinatorial bandit algorithms with regret guarantees. As for the previous combinatorial TS algorithms, \citet{wen2015efficient} proposed a TS algorithm for a contextual combinatorial bandits with semi-bandit feedback and a linear score function. However, the regret bound for the algorithm is only analyzed in the Bayesian setting (hence establishing the Bayesian regret) which is a weaker notion of regret and much easier to control in combinatorial action settings. To our knowledge, \citet{oh2019thompson} was the first work to establish the worst-case regret bound for a variant of contextual combinatorial bandits, multinomial logit (MNL) contextual bandits, utilizing the optimistic sampling procedure similar to $\CNTS$. Yet, our proposed algorithm differs from \citet{oh2019thompson} in that we sample directly from the score space rather than the parameter space which avoids the computational complexity of sampling a high-dimensional network parameters. 
More importantly, \citet{oh2019thompson} exploit the structure of the MNL choice feedback model to derive the regret bound whereas we address a more general semi-bandit feedback without any assumptions on the structure of the feedback model. 

\section{Additional Experiments}
\begin{figure*}[t!]
    \begin{center}
    \centerline{\includegraphics[scale=0.4]{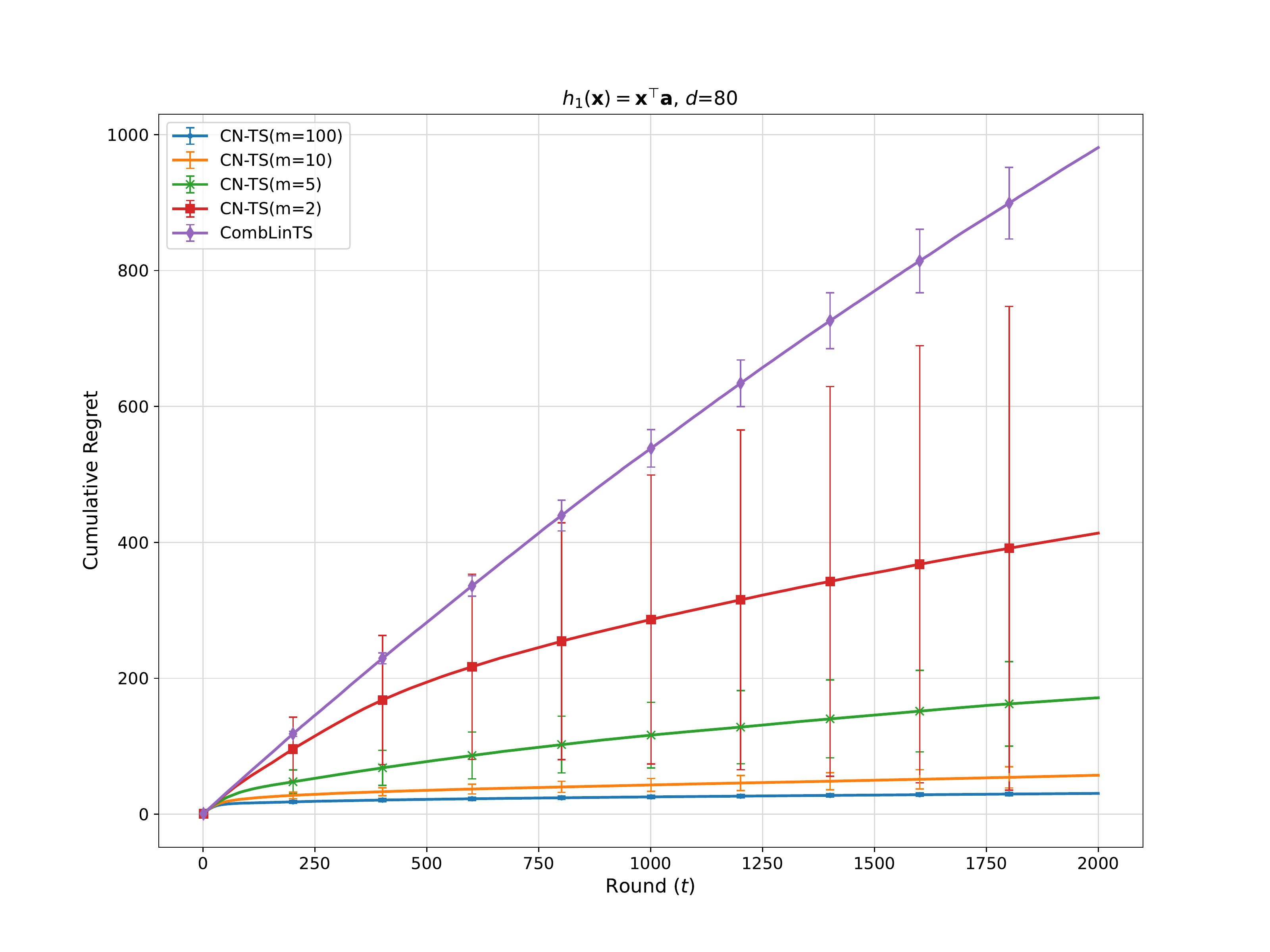}}
    \vskip -0.2in 
    \caption{Cumulative regret of $\CNTS$ and \texttt{CombLinTS} with respect to the network width ($m$).}
    \label{Fig3}
    \end{center}
\vskip -0.4in 
\end{figure*}

In Experiment 1, the linear combinatorial bandit algorithms perform worse than our proposed algorithms, even for the linear score function.
One of the possible reasons for this is that the neural network based algorithms use much larger number of parameters than the linear model based algorithms, overparametrized for the problem setting.
Overparametrized neural networks have been shown to have superior generalization performances. See~\citet{allenzhu2019convergence, allenzhu2019learning}.
Note that the regret performance is about the generalization to the unseen data rather than it is
about the fit to the existing data. 
In this aspect, overparameterized neural network can show superior
performance over the linear model.
This is supported by Figure~\ref{Fig3}. In Figure~\ref{Fig3}, we demonstrate the empirical performances of $\CNTS$ ans $\CombLinTS$ as the network width $m$ decreases. 
We can see that by decreasing $m$, the results of the neural network models and 
linear models become more similar, i.e., the gap between the regrets reduce.

\end{document}